\documentclass{bmvc2k}
\usepackage{multirow}
\usepackage{tablefootnote}
\usepackage{subcaption}
\usepackage{placeins}

\title{Attention is All We Need:\\Nailing Down Object-centric Attention for Egocentric Activity Recognition}

\addauthor{Swathikiran Sudhakaran$^{1,2}$}{sudhakaran@fbk.eu}{1}
\addauthor{Oswald Lanz$^1$}{lanz@fbk.eu}{2}

\addinstitution{
$^1$\parbox[t]{\linewidth}{Fondazione Bruno Kessler\\Trento, Italy}
}
\addinstitution{
$^2$\parbox[t]{\linewidth}{University of Trento\\Trento, Italy}
}

\runninghead{Sudhakaran, Lanz}{Object-centric Attention for Activity Recognition}

\def\etal{\emph{et al}\bmvaOneDot}

\usepackage{soul}
\newcommand{\rev}[2]{#2}

\begin{document}

\maketitle

\begin{abstract}

In this paper we propose an end-to-end trainable deep neural network model for egocentric activity recognition. Our model is built on the observation that egocentric activities are highly characterized by the objects and their locations in the video. Based on this, we develop a spatial attention mechanism that enables the network to attend to regions containing objects that are correlated with the activity under consideration. We learn highly specialized \rev{spatial}{} attention maps for each frame using class-specific activations from a CNN pre-trained for generic image recognition, and use them for spatio-temporal encoding of the video with a convolutional LSTM. \rev{We validate our method on standard egocentric activity benchmarks, our model surpasses state-of-the-art recognition accuracy by a large margin, with an average improvement of 19\% compared to existing approaches.}{Our model is trained in a weakly supervised setting using raw video-level activity-class labels. Nonetheless, on standard egocentric activity benchmarks our model surpasses by up to +6\% points recognition accuracy the currently best performing method that leverages hand segmentation and object location strong supervision for training.}
We \rev{}{visually analyze attention maps generated by the network, revealing that the network successfully identifies the relevant objects present in the video frames which may explain the strong 
recognition performance. We} also discuss an extensive ablation analysis regarding the design choices.

\end{abstract}

\section{Introduction}
\label{sec:intro}

Automated analysis of videos for content understanding is one of the most challenging and well researched areas in computer vision and multimedia and possesses a vast array of applications ranging from surveillance, behavior understanding, video indexing and retrieval, human-machine interaction, etc. The majority of researchers working on video understanding problem concentrates on action recognition from distant or third person views while egocentric activity analysis has been investigated \rev{only}{more} recently. One of the major challenges in egocentric video analysis is the presence of ego-motion due to the frequent body movement of the camera wearer. This ego-motion may or may not be \rev{a}{} representative of the action performed by the observer and as a result, existing techniques proposed for general action recognition become less suitable for egocentric video analysis. Another reason hampering the development of this area is the lack of large scale datasets to enable the training of deep neural networks. However, with the ubiquitous availability and popularity of first-person cameras in recent times, providing more attention to this research area is of paramount importance. 

In this work, we take on the problem of fine-grained recognition of egocentric activities, which is more challenging than egocentric action recognition. Action recognition involves \rev{in}{} identifying a generalized motion pattern of hands such as take, put, stir, pour, etc. whereas activity recognition concerns more fine-grained \rev{problems}{composite patterns} such as take bread, take water, put sugar in coffee, put bread in plate, etc. For developing a system capable of recognizing activities, it is pertinent to identify both the hand motion patterns as well as the objects on to which \rev{the motion}{a manipulation} is being applied to. Majority of state-of-the-art techniques use hand segmentation \cite{ma2016deeper} \cite{li2015delving} \cite{singh2016first}, gaze location \cite{li2015delving} or object bounding boxes \cite{ma2016deeper} for identifying the location of the relevant objects in the scene that can assist in identifying the activity. These approaches require complex pre-processing which includes human intervention for generating hand masks or gaze locations of the video frames. Even though there exist wearable devices which can estimate the gaze direction \rev{}{to dispose of the excessive cost of manual annotation}, these may cause discomfort to the user or result in inaccuracies \rev{}{during distraction or short interruption of the activity or} if the user is wearing glasses. \rev{Considering the aforementioned problems, we propose a novel end-to-end trainable deep neural network for recognizing first-person activities which uses a spatial attention mechanism for detecting the location of the relevant object pattern in the scene that is a representative of the activity class.}{

Considering the aforementioned problems that prevent from leveraging massive collections of natural activity videos, it is \rev{imperative}{essential} to develop techniques capable of identifying the relevant objects without being trained with full supervision. Towards this end, we 
present a CNN-RNN architecture that is trained in a weak supervision setting to predict the raw video-level activity-class label associated with the clip. Our CNN backbone is pretrained for generic image recognition and augmented on top with an attention mechanism that uses class activation maps for spatially selective feature extraction. The memory tensor of a convolutional LSTM then tracks the discriminative frame-based features distilled from the video for activity classification. Our design choices are grounded to fine grained activity recognition because: (i) Frame-based activation maps are not bound to reflect image recognition classes, they develop their own representation classes implicitly while training the video-level classification; (ii) Convolutional LSTM maintains the spatial structure of the input sequence all the way up to the final video descriptor used by the activity classification layer, thus facilitating the spatio-temporal encoding of objects and their locations into the descriptor as they develop into the activity over time. Our contributions can be summarized as follows:}
\begin{itemize}
\item We develop \rev{and}{an} end-to-end trainable deep neural network for egocentric activity recognition which \rev{uses}{elicits} spatial attention on relevant objects present in the scene for highly specialized spatio-temporal encoding;
\item\rev{}{We provide insights into the inner workings of our method by visualizing the attention maps that are generated at the CNN-RNN interface of our architecture, they are highly specialized and developed during training without hand segmentation and object location strong supervision and label semantics;
}
\item We perform experimental validation on four benchmark datasets with up to 106 activity classes
. We also perform an extensive ablation study. \rev{}{We release an implementation in pytorch at \href{https://github.com/swathikirans/ego-rnn}{\tt github.com/swathikirans/ego-rnn}}.
\end{itemize}


\section{Related works}
\label{sec:rel_works}

The majority of the methods developed for addressing egocentric activity recognition problem make use of semantic cues such as objects being handled, gaze direction, hand pose, ego-motion, etc. Fathi \etal \cite{fathi2011understanding} developed a hierarchical model based on object-hand configurations that makes use of information about object class, pose, hand optical flow, hand pose, etc. for egocentric activity recognition. In this, the model predicts actions based on the hand-object interactions which in turn is used to predict the activity. Fathi and Rehg \cite{fathi2013modeling} proposes to consider activity as a change of state of the objects and the materials present in the scene. A spatio-temporal pyramid approach in which the bin boundaries are sampled near the objects being handled by the user is proposed by McCandless and Grauman in \cite{mccandless2013object}. Matsuo \etal \cite{matsuo2014attention} proposes a method to generate an attention map of the scene based on visual saliency and ego-motion information. The attention map is then used to classify the detected objects into salient and non-salient objects followed by a temporal pyramidal approach based feature extraction for activity recognition. Li \etal \cite{li2015delving} proposes to encode egocentric cues along with dense trajectory features for activity recognition. They show that egocentric cues such as hand pose, hand motion, head movement, gaze direction, etc. can be used as efficient features for classifying egocentric activities. 

Several deep learning based methods have been proposed for action recognition from third person videos in recent years. Majority of the approaches follow a two-stream network \cite{simonyan2014two} \cite{TSN2016ECCV} \cite{carreira2017quo} consisting of an appearance stream encoding RGB images and a temporal stream encoding stacked optical flow. A number of methods adopting this two-stream architecture along with encoding of egocentric cues have been recently proposed. Singh \etal \cite{singh2016first} adds an additional input stream, called ego-stream, to the two-stream network that takes in a stacked input of hand mask, saliency map and head motion for egocentric action recognition. A two-stream network consisting of an appearance stream trained for hand segmentation and object localization and a temporal stream for action recognition is proposed by Ma \etal \cite{ma2016deeper}. The two individual networks are then jointly trained for recognizing fine-grained activities. Tang \etal \cite{tang2017action} proposes to add an additional stream for encoding depth maps for improving activity recognition performance. Ryoo \etal \cite{ryoo2015pooled} proposes to use a convolutional neural network (CNN) for extracting frame level features which are then applied to a set of temporal filters that perform multiple pooling operations such as max pooling, sum pooling, histogram of gradients pooling, for generating effective feature descriptors.

Existing methods use a single RGB frame for extracting appearance information and neglect how the spatio-temporal changes evolve as the activity progresses. Xingjian \etal \cite{xingjian2015convolutional} proposes a variant of recurrent neural network called convolutional long short-term memory (convLSTM), which is a variant of long short-term memory (LSTM) \cite{hochreiter1997long} with convolutional gates instead of fully-connected gates, for predicting precipitation nowcasting from radar image sequences. The presence of convolutional gates allow the propagation of a memory tensor that enables the preservation of spatial structure of the input tensor. Deep networks that use convLSTM for encoding spatio-temporal changes have been proposed for addressing problems such as violence detection \cite{sudhakaran2017learning}, anomaly detection \cite{medel2016anomaly} and interaction recognition \cite{Sudhakaran_2017_ICCV}. In the proposed method, we choose to use convLSTM for encoding the spatio-temporal features present in the input video. As has been discovered by previous approaches, identifying the objects being handled by the user is useful in improving the performance of an activity recognition system. In the proposed approach, we use the idea of class activation map (CAM) proposed by Zhou \etal \cite{zhou2016learning} for identifying the location of the object, that can be used for discriminating one action from another, in the scene. We build on CAM as a spatial attention mechanism for weighting the features before encoding them.

\section{Proposed method}
\label{sec:prop}
This section explains the proposed method for egocentric activity recognition, consisting of deep feature encoding with spatial attention. We will also briefly discuss class activation maps (CAMs), upon which the proposed activity recognition system is developed.

\subsection{Class activation maps}
\label{subsec:cam}

\begin{figure}[t]
		\centering      
        \begin{subfigure}[b]{0.23\textwidth}
			\includegraphics[scale=0.17]{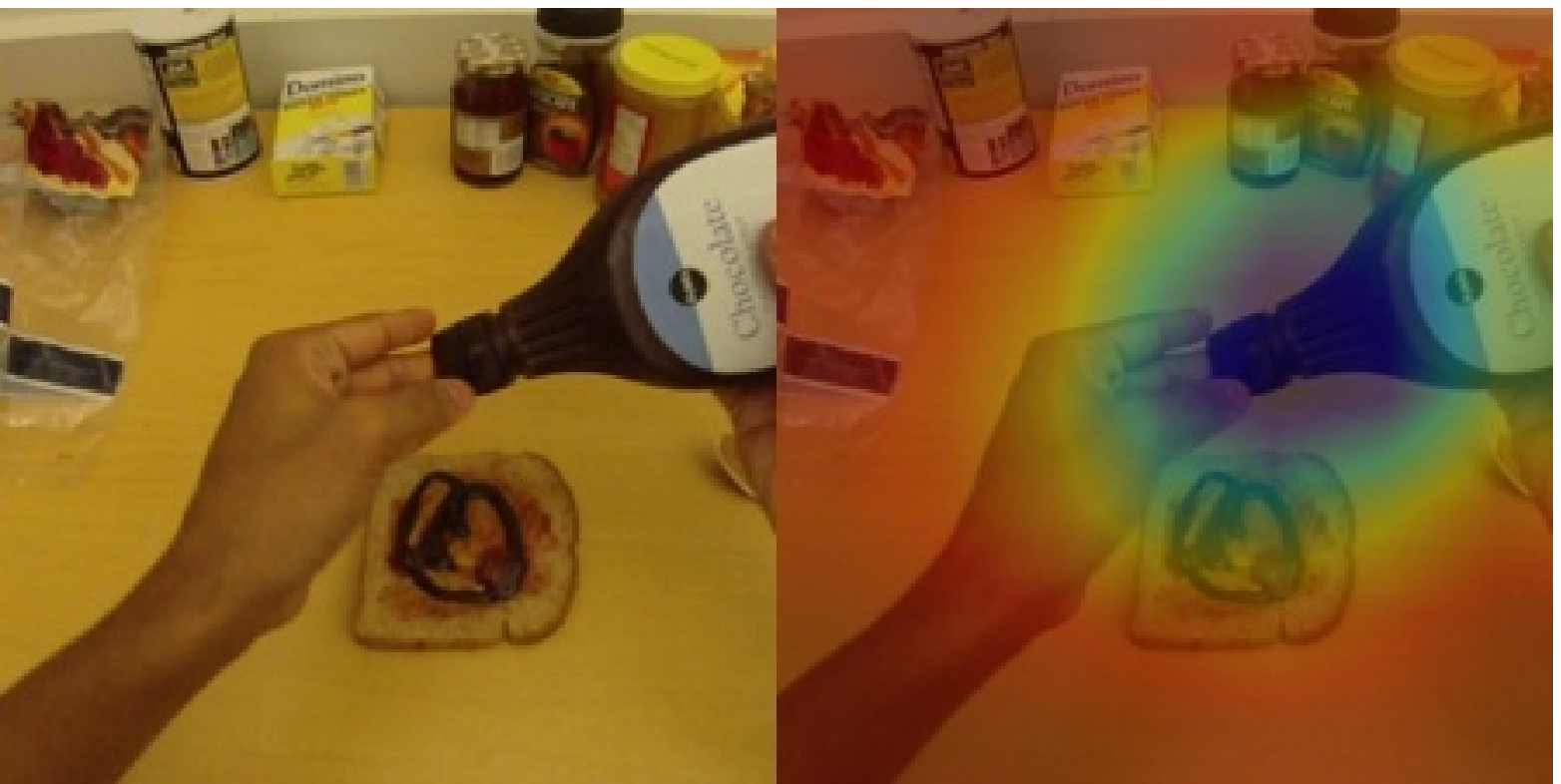}
            \caption{Close chocolate}
            \label{fig_ex1}
		\end{subfigure}
		\begin{subfigure}[b]{0.23\textwidth}
			\includegraphics[scale=0.17]{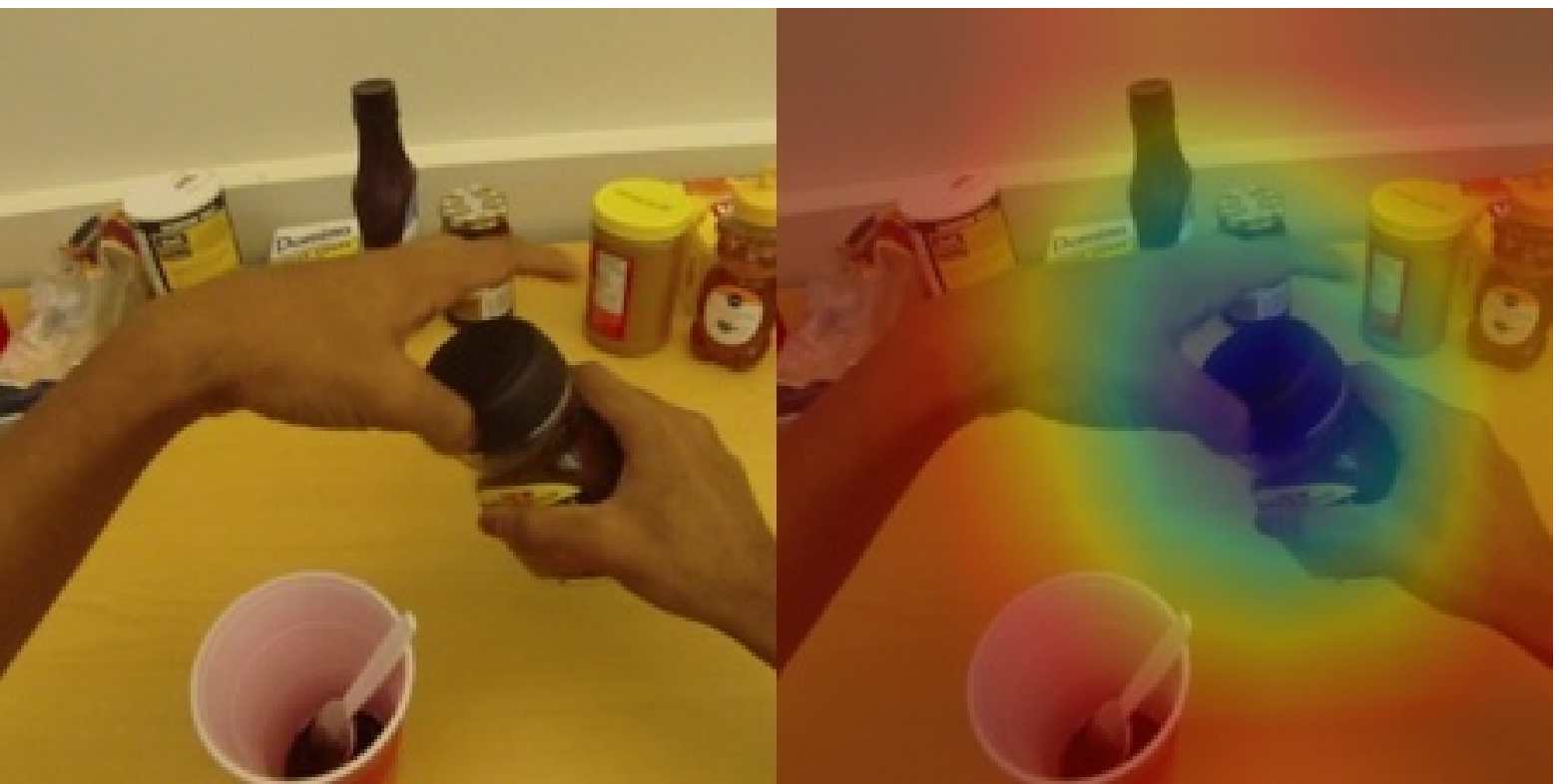}
			\caption{Close coffee}
            \label{fig_ex2}
		\end{subfigure}
        \begin{subfigure}[b]{0.23\textwidth}
			\includegraphics[scale=0.17]{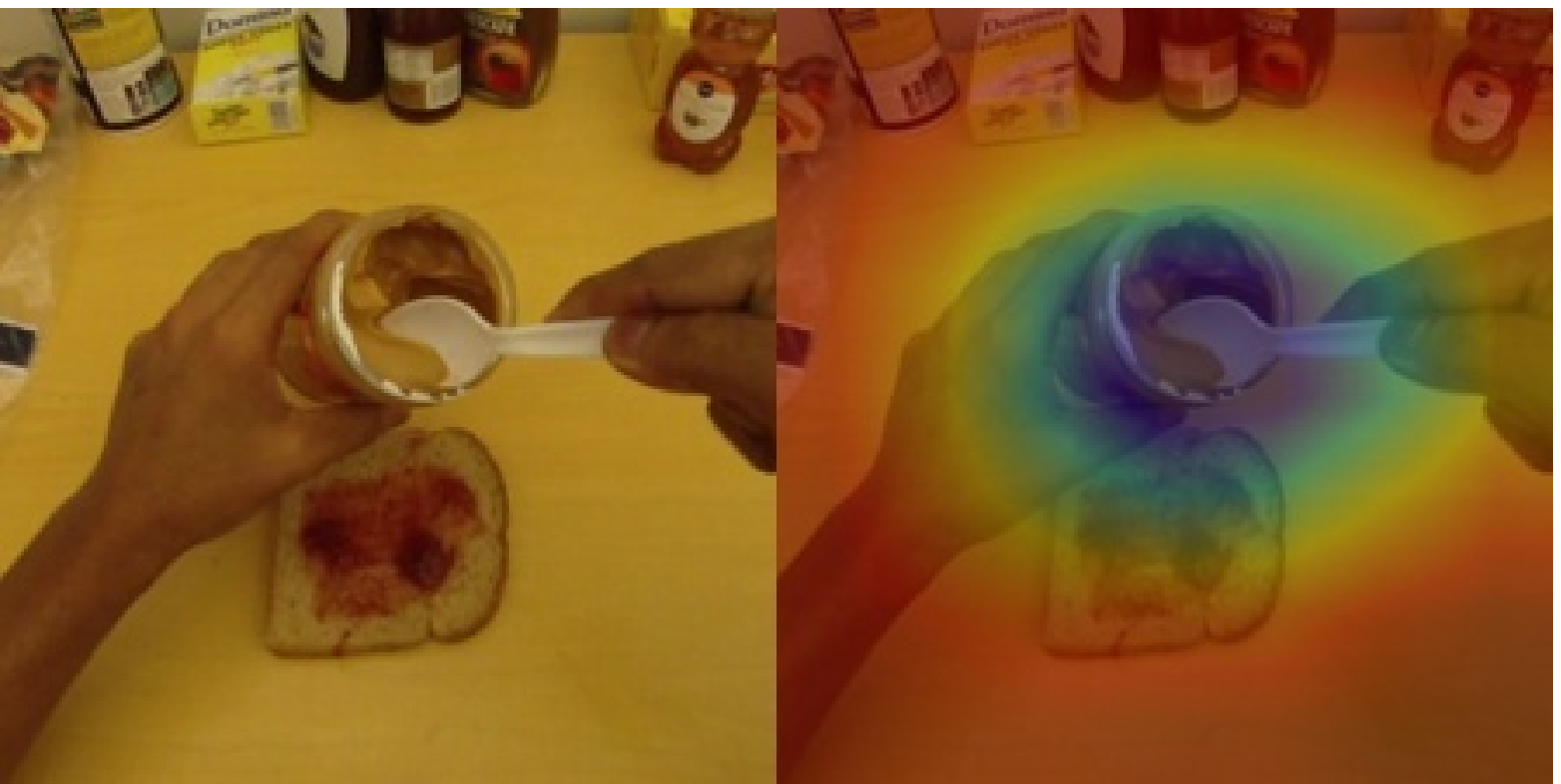}
            \caption{Scoop peanut}
			\label{fig_ex3}
		\end{subfigure}
        \begin{subfigure}[b]{0.23\textwidth}
			\includegraphics[scale=0.17]{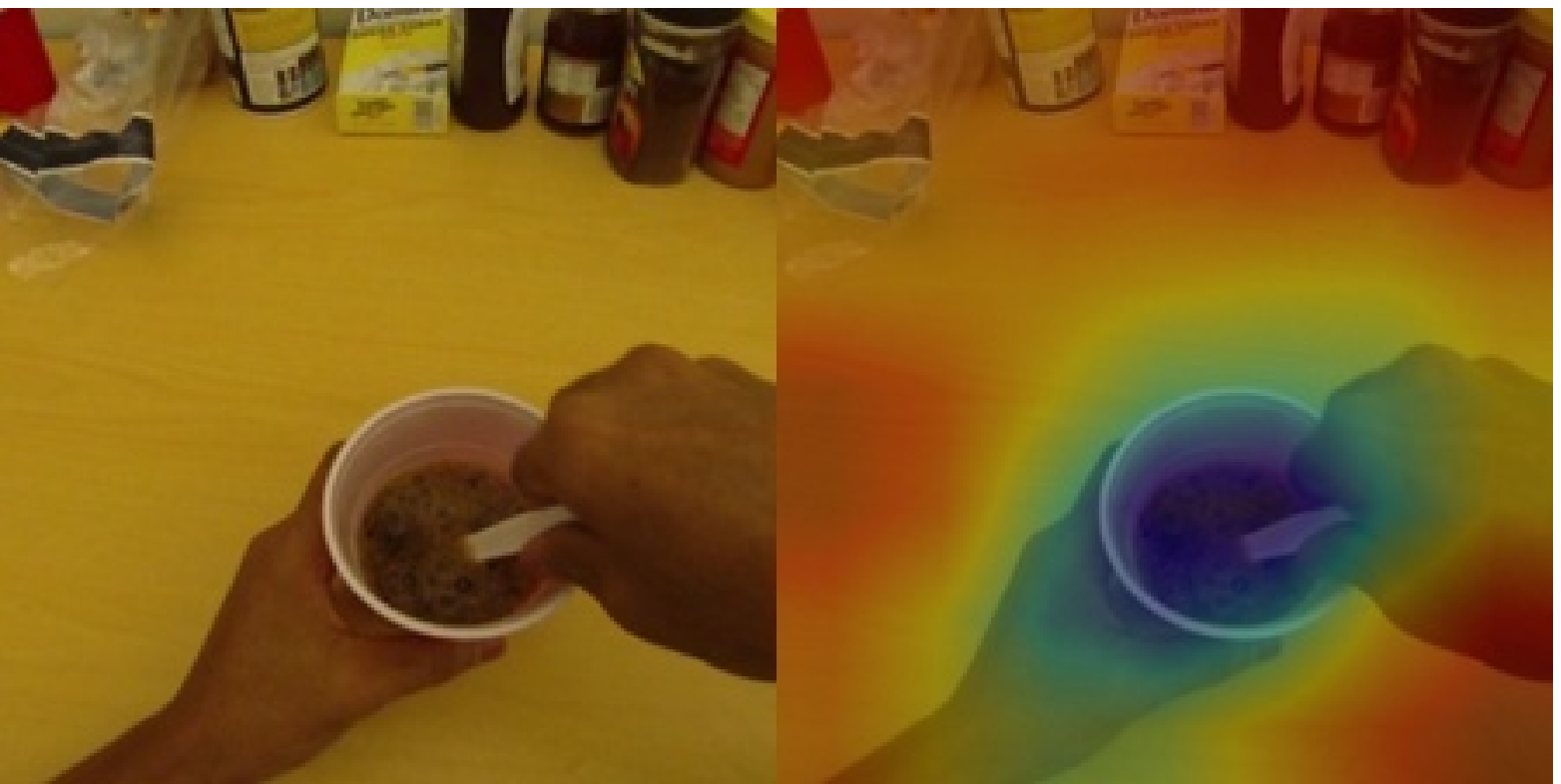}
            \caption{Stir spoon}
			\label{fig_ex4}
		\end{subfigure}\\
        \vskip .5mm
        \begin{subfigure}[b]{0.23\textwidth}
			\includegraphics[scale=0.17]{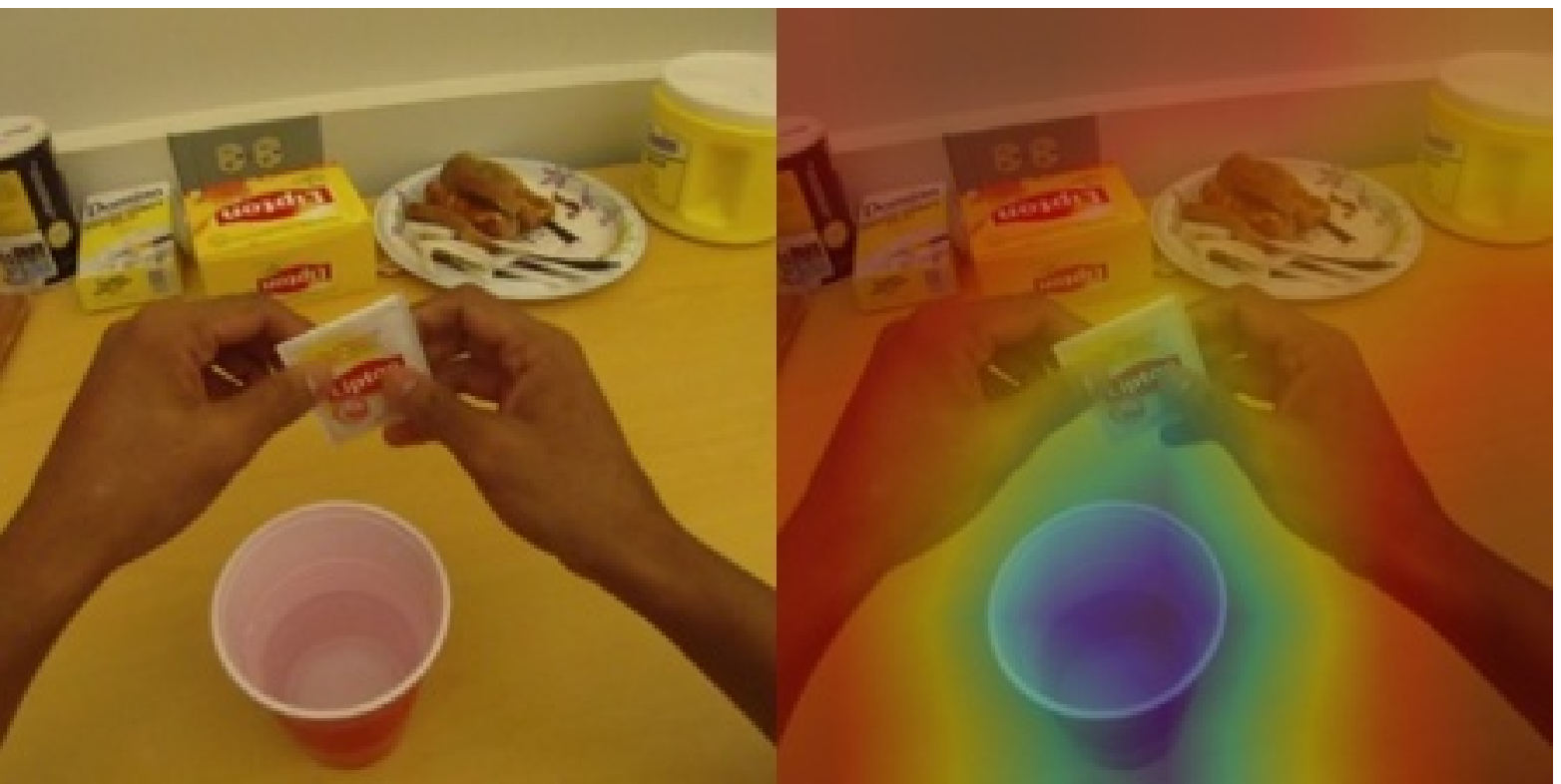}
            \caption{Open tea}
			\label{fig_ex5}
		\end{subfigure}
        \begin{subfigure}[b]{0.23\textwidth}
			\includegraphics[scale=0.17]{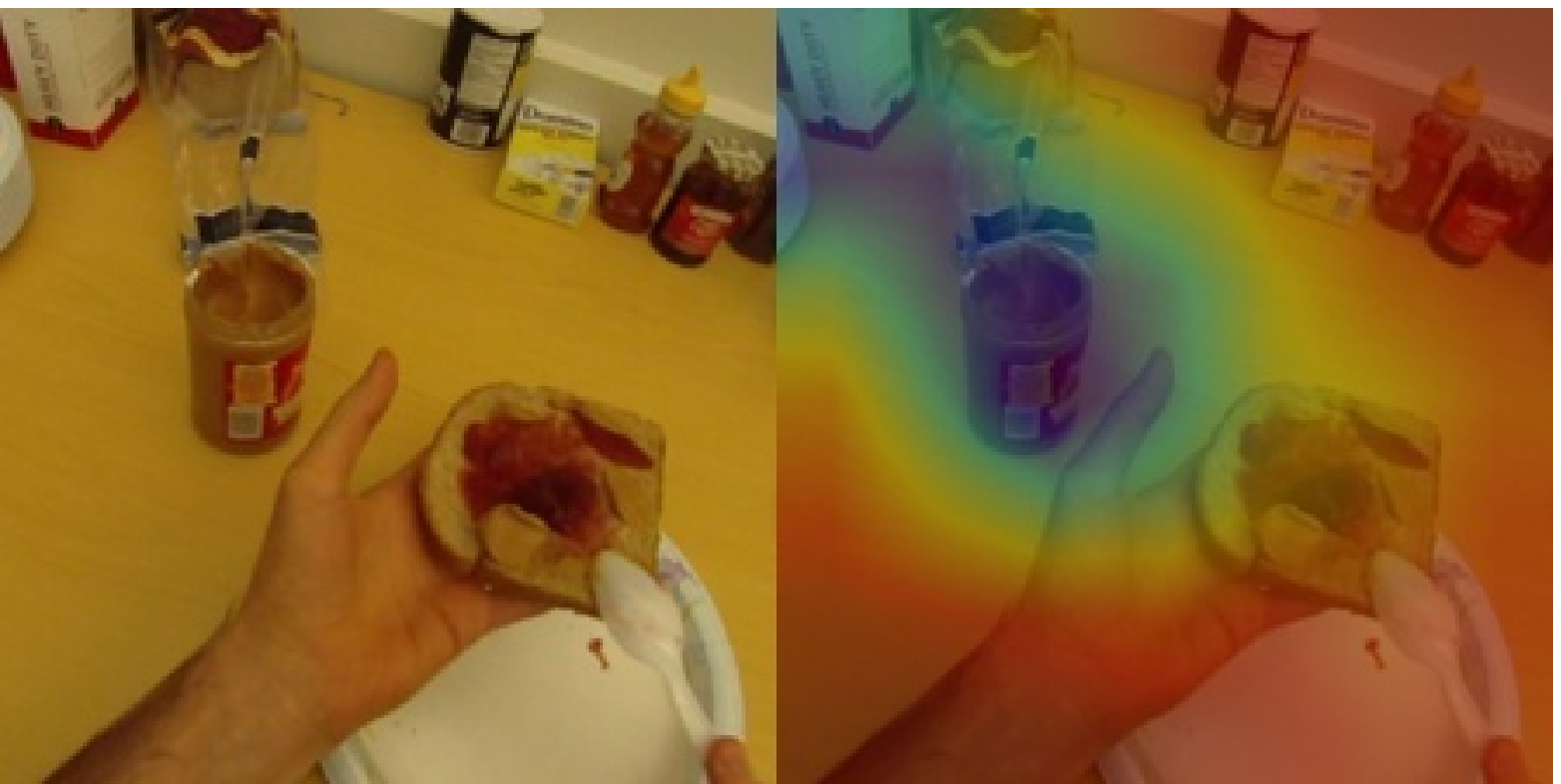}
            \caption{Spread peanut}
			\label{fig_ex6}
		\end{subfigure}
        \begin{subfigure}[b]{0.23\textwidth}
			\includegraphics[scale=0.17]{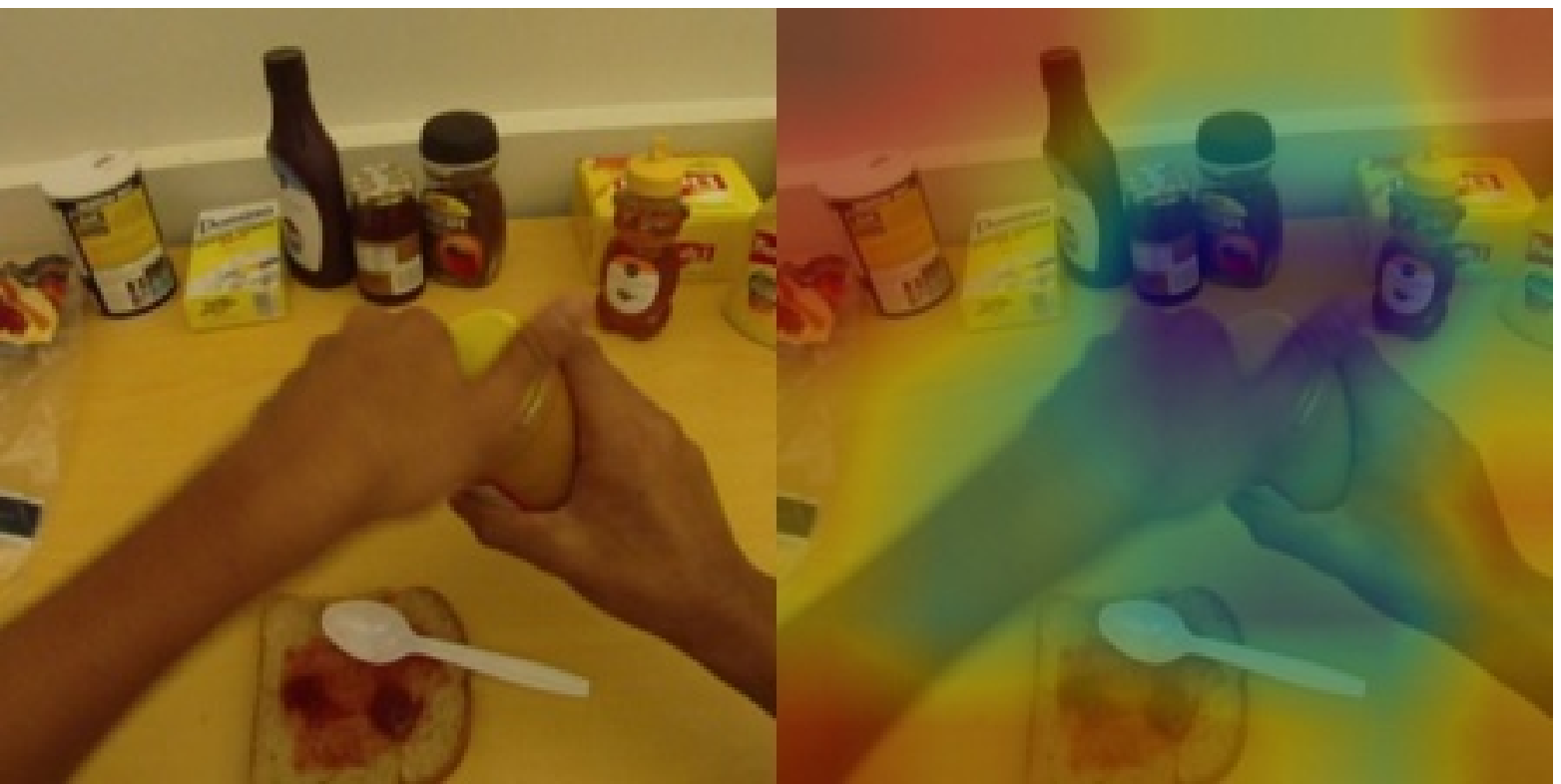}
            \caption{Open peanut}
			\label{fig_ex7}
		\end{subfigure}
        \begin{subfigure}[b]{0.23\textwidth}
			\includegraphics[scale=0.17]{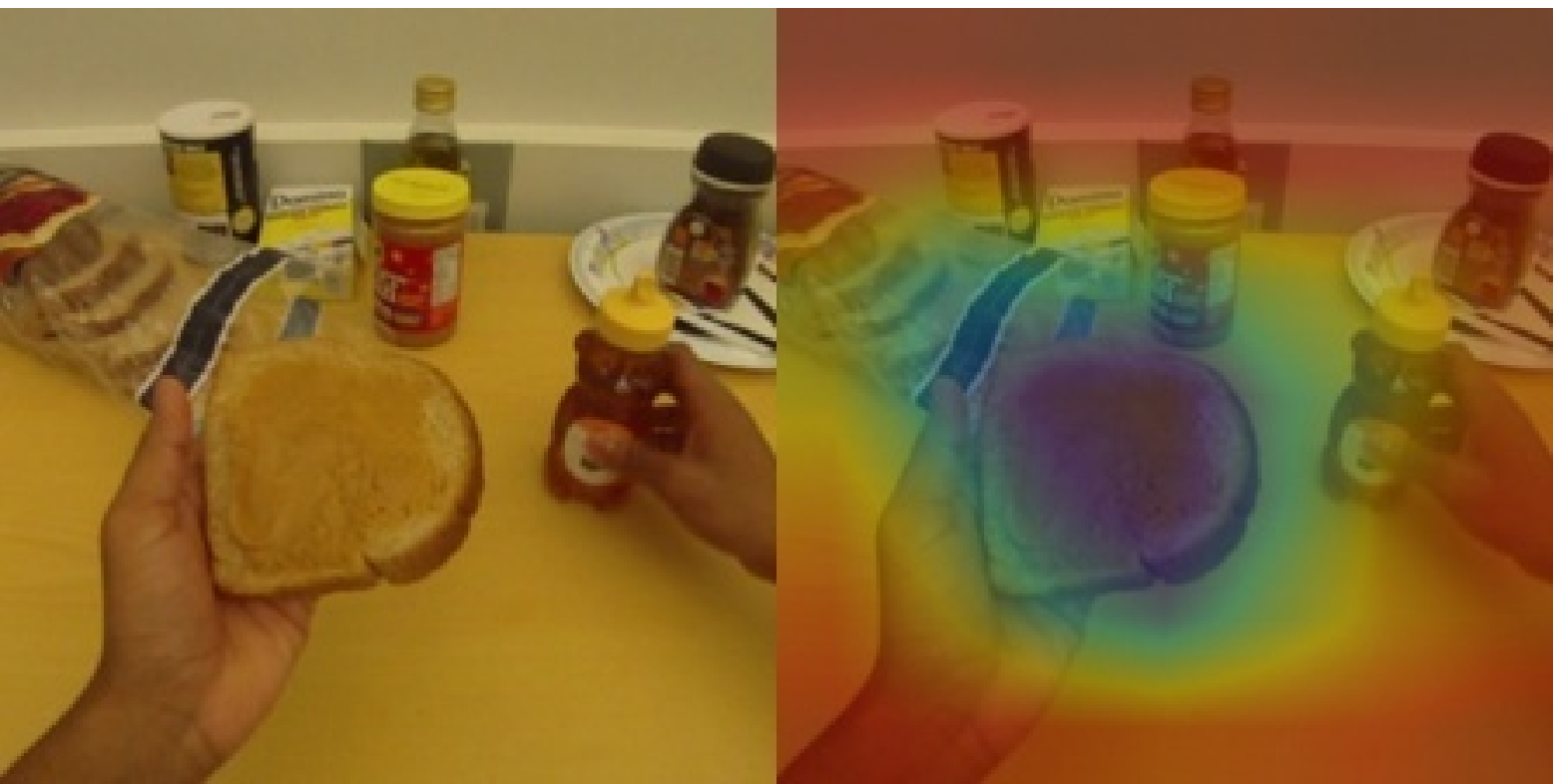}
            \caption{Take honey}
			\label{fig_ex8}
		\end{subfigure}
		\vspace*{2mm}
        \caption{Class activation maps obtained for frames from GTEA 61 dataset using ResNet-34 trained on imagenet.}
		\label{fig:fig_ex}
	\end{figure}
\label{subsec:spat_att}

Zhou \etal \cite{zhou2016learning} proposes to take advantage of the average pooling layer present in modern deep CNN networks such as ResNet \cite{he2016deep}, squeezeNet \cite{iandola2016squeezenet}, denseNet \cite{huang2017densely} for generating class specific saliency maps. Let $f_l(i)$ be the activation of a unit $l$ in the final convolutional layer at spatial location $i$ and $w_l^c$ be the weight corresponding to class $c$ for unit $l$. Then the CAM for class $c$, $M_c(i)$, can be represented as
	\begin{equation}
	M_c(i) = \sum_{l}w_l^cf_l(i)
	\label{eq:CAM}
	\end{equation}
	Using the CAM, $M_c$, thus generated, we can identify the image regions that have been used by the CNN for identifying the class under consideration, which is class $c$. If we use the winning class, i.e., the class category with the highest probability, then the CAM generated gives us a saliency map of the image. Since egocentric activities are representative of the objects being handled by the observer, we can use CAMs for making the network to focus on the regions consisting of the objects.

Figure~\ref{eq:CAM} shows some examples of CAMs generated using ResNet-34 on some of the frames from GTEA 61 dataset. The CAM generated is of the same spatial dimension as the features obtained from the final layer of the network ($7\times 7$ for ResNet-34). The CAM obtained is then resized to the input image dimensions for visualization purpose. From the figures \ref{fig_ex1}, \ref{fig_ex2}, \ref{fig_ex3}, \ref{fig_ex4}, we can see that the network, which is pre-trained on imagenet dataset, is capable of identifying the discriminant regions in the image which can be used for recognizing the activity class. However, in Figure~\ref{fig_ex5}, \ref{fig_ex6}, \ref{fig_ex7}, \ref{fig_ex8}, the regions obtained from the winning class of the CNN cannot be considered as a representative of the activity. In section \ref{subsec:spat_att}, we will explain how we integrate and specialize CAM as spatial attention to identify the appropriate regions that can discriminate the activity under consideration from others.

\subsection{Spatial attention using class activation maps}

	 \begin{figure}[t]
		\centering
		\includegraphics[width=\textwidth]{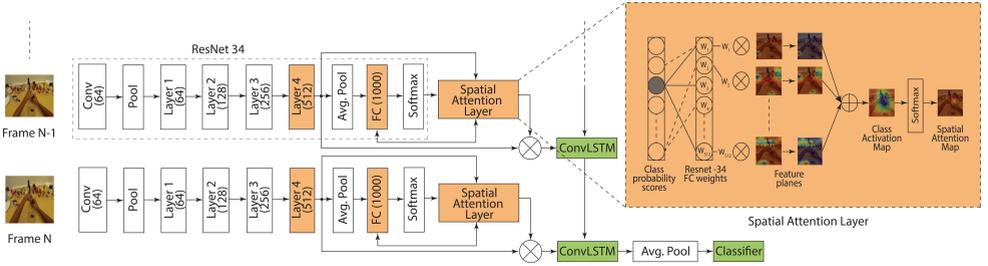}
		\vspace*{2mm}
		\caption{Block diagram of the proposed egocentric action recognition schema. We use ResNet-34 for frame-based feature extraction and spatial attention map generation, and convLSTM for temporal aggregation. Colored blocks indicate trained network layers, we use a two-stage strategy: stage 1 green, stage 2 green+orange.
		}
		\label{fig:block_dia}
	\end{figure}

The block diagram of the proposed idea for encoding the RGB frames is shown in Figure~\ref{fig:block_dia}. In this, we use a ResNet-34 network pre-trained on imagenet as the backbone architecture. For each RGB input frame, we will first compute the CAM using the winning class as given by equation \ref{eq:CAM}. The CAM thus obtained is then converted to a probability map by applying the softmax operation along the spatial dimensions. This spatial attention map is then multiplied with the output of the final convolutional layer of ResNet-34, that is:
\begin{equation}
f_{SA}(i) = f(i) \odot \frac{e^{M_c(i)}}{\sum_{i'}e^{M_c(i')}}
\label{eq:spat_attn}
\end{equation}
where, $f(i)$ represents the output feature from the final convolutional layer of ResNet-34 at a spatial location $i$, $M_c(i)$ is the CAM obtained using the winning class $c$, $f_{SA}(i)$ is the image feature after spatial attention is applied and $\odot$ represents the Hadamard product.

Once we get image features with spatial attention ($f_{SA}$), the next step is to perform temporal encoding of frame level features. We use a convolutional long short-term memory (ConvLSTM) module for performing this operation, which has been shown to be successful in spatio-temporal encoding of features \cite{medel2016anomaly} \cite{sudhakaran2017learning} \cite{Sudhakaran_2017_ICCV}. The convLSTM works similar to that of the traditional LSTM with the exceptions of having 3D tensors as input and hidden state instead of vectors, and the presence of convolutional layers in the gates instead of fully-connected layers. By using convLSTM for temporal encoding, it becomes possible to encode the changes in both temporal and spatial dimensions simultaneously. This is important since the network needs to track changes in both these dimensions if it is to learn how an activity evolves. The working of the convLSTM module is represented by the following equations:
\begin{eqnarray}
i_t = \sigma(w_x^i* f_{SA} + w_h^i*h_{t-1} + b^i)\\
f_t = \sigma(w_x^f* f_{SA} + w_h^f*h_{t-1} + b^f)\\
\tilde{c}_t = \tanh(w_x^{\tilde{c}}*f_{SA} + w_h^{\tilde{c}}*h_{t-1} + b^{\tilde{c}})\\
c_t = \tilde{c}_t\odot f_{SA} + c_{t-1}\odot f_t\\
o_t = \sigma(w_x^o * f_{SA} + w_h^o*h_{t-1} + b^o)\\
h_t = o_t\odot \tanh(c_t)
\label{eq:convLSTM}
\end{eqnarray}
\rev{}{where $\sigma$ is the sigmoid function, $i_t$, $f_t$, $o_t$, $c_t$ and $h_t$ represent the input state, forget state, output state, memory state and hidden state, respectively, of the convLSTM. The trainable weights and biases of the convLSTM are represented using $w$ and $b$.} The spatial stream network works in the following way. During each time step, an input RGB frame will be applied to the network. The spatial attention will be computed and then image features with spatial attention is applied to the convLSTM module for temporal encoding. Once all the input frames are applied to the network, the memory of the convLSTM module, $c_t$ is selected as the feature describing the entire video frames. Spatial average pooling operation is then applied on this $c_t$ to obtain the video feature descriptor which is then fed to the classifier layer consisting of a fully-connected layer for generating the class-category scores. We follow a two-stage training strategy in which only the classifier and the convLSTM layers are trained in the first stage followed by a second stage, where the convolutional layers of the final layer and the fully-connected (FC) layer of the ResNet-34 network are also trained in addition to the convLSTM and classifier layers. This is represented using the colored blocks in Figure~\ref{fig:block_dia}. By training the convolutional layers and FC layer of the ResNet-34 network, the network's capability of localizing the relevant features for discriminating the activity will be greatly improved.

\section{Experimental Results}
\label{sec:res}

The proposed egocentric activity recognition method is evaluated on four datasets, namely, GTEA 61, GTEA 71, GTEA Gaze+ and EGTEA Gaze+ datasets. We follow the same evaluation settings used by previous approaches. For GTEA 61 dataset, we report the results obtained on the fixed split (subject S2 for evaluation) as well as leave-one-subject-out cross-validation setting. For GTEA 71 and GTEA Gaze+ datasets, the leave-one-subject-out cross-validation results are reported. In the case of EGTEA Gaze+ dataset, we report the results on each of the three provided train-test splits as well as the average across the three splits.

\subsection{Implementation details}
\label{subsec:impl_det}

As mentioned previously, ResNet-34 pre-trained on imagenet dataset is used as the base architecture for extracting frame level features and generating the object specific spatial attention map. We use a convLSTM module with 512 hidden units for temporal encoding. The network is trained in two stages as explained in section \ref{subsec:spat_att}. In the first stage, the network is trained for 200 epochs with an initial learning rate of $10^{-3}$ and the learning rate is decayed by a factor of 0.1 after 25, 75 and 150 epochs. We also apply dropout at a rate of 0.7 at the fully-connected classifier layer. In the second stage, we train the network for 150 epochs with a learning rate of $10^{-4}$ and is decayed after 25 and 75 epochs by a factor of 0.1. We use ADAM optimization algorithm with a batch size of 32 during training. We select 25 frames from each video, uniformly sampled in time, as the inputs to the network.

As is commonly done with deep learning based action recognition techniques \cite{simonyan2014two} \cite{TSN2016ECCV} \cite{ma2016deeper} \cite{singh2016first} \cite{tang2017action} \cite{carreira2017quo}, we also use a temporal network that uses stacked optical flow images as input, in order to better encode the motion changes occuring in the input video. For this, we train a ResNet-34 pre-trained on imagenet. We follow the cross-modality initialization approach proposed by Wang \etal in \cite{TSN2016ECCV} for initializing the input layer. In this approach, the weights of the first convolutional layer is initialized by computing the mean of the RGB pre-trained network's weights and replicating them by the number of channels of the target temporal network. The temporal network is trained for 750 epochs with a batch size of 32 using stochastic gradient descent algorithm with a momentum of 0.9. The learning rate is fixed as $10^{-2}$ initially and is reduced by a factor of 0.5 after 150, 300 and 500 epochs. We use five stacked optical flow images as the input to the network and during evaluation, we average the scores of 5 such stacks, uniformly sampled in time, to obtain the final classification scores. We use TV-L1 algorithm \cite{zach2007duality} for optical flow computation.

Once the spatial and temporal networks are trained, we follow two approaches for the fusion of these two. In the first approach, as proposed in \cite{simonyan2014two}, we simply average the scores obtained from each of the two networks to obtain the final classification score. In the second approach, we concatenate the outputs of the two networks and add a new fully-connected layer on top to get the class-category scores. \rev{We then train the newly added fully-connected layer along with the fully-connected layers of the individual networks for 50 epochs with a learning rate of $10^{-3}$ which is reduced by a factor of 0.5 after 5, 20 and 50 epochs}{We then fine-tune all the layers down to Layer4 of both networks for 250 epochs with a learning rate of 10$^{-2}$, which is reduced by a factor of 0.1 after each epoch}. \rev{ADAM}{SGD} algorithm is used as the optimization algorithm.

Since the amount of training samples are limited and to avoid overfitting, we use corner cropping, scale jittering and random horizontal flipping approaches as proposed in \cite{TSN2016ECCV} during training stages. During evaluation, the center crop of the frames are used to get the classification scores.

\begin{table}[t]
	\begin{center}
    \begin{tabular}{|c|c|}
		\hline
		  Configuration & \parbox{0.9in}{\centering Accuracy (\%)}  \\
         \hline
         LRCN & 34.48 \\
         \hline
         ConvLSTM & 51.72 \\
         \hline
           ConvLSTM-attention & 63.79 \\
          \hline\hline
          temporal-optical flow & 44.83 \\
          \hline
          temporal-warp flow & 48.28 \\
          \hline\hline
          two-stream (average) & 67.24 \\
       	  \hline
          two-stream (joint train) & \rev{97.41}{77.59}\\
          \hline
	\end{tabular}
    \end{center}
    \caption{Comparison of different configurations on the fixed split of GTEA 61 dataset.}
	\label{tab:conf_comp}
\end{table}

\subsection{Ablation studies}
\label{subsec:abl_stud}

\begin{figure}[t]
		\centering
        \begin{subfigure}[b]{0.32\textwidth}
        \includegraphics[scale=0.23]{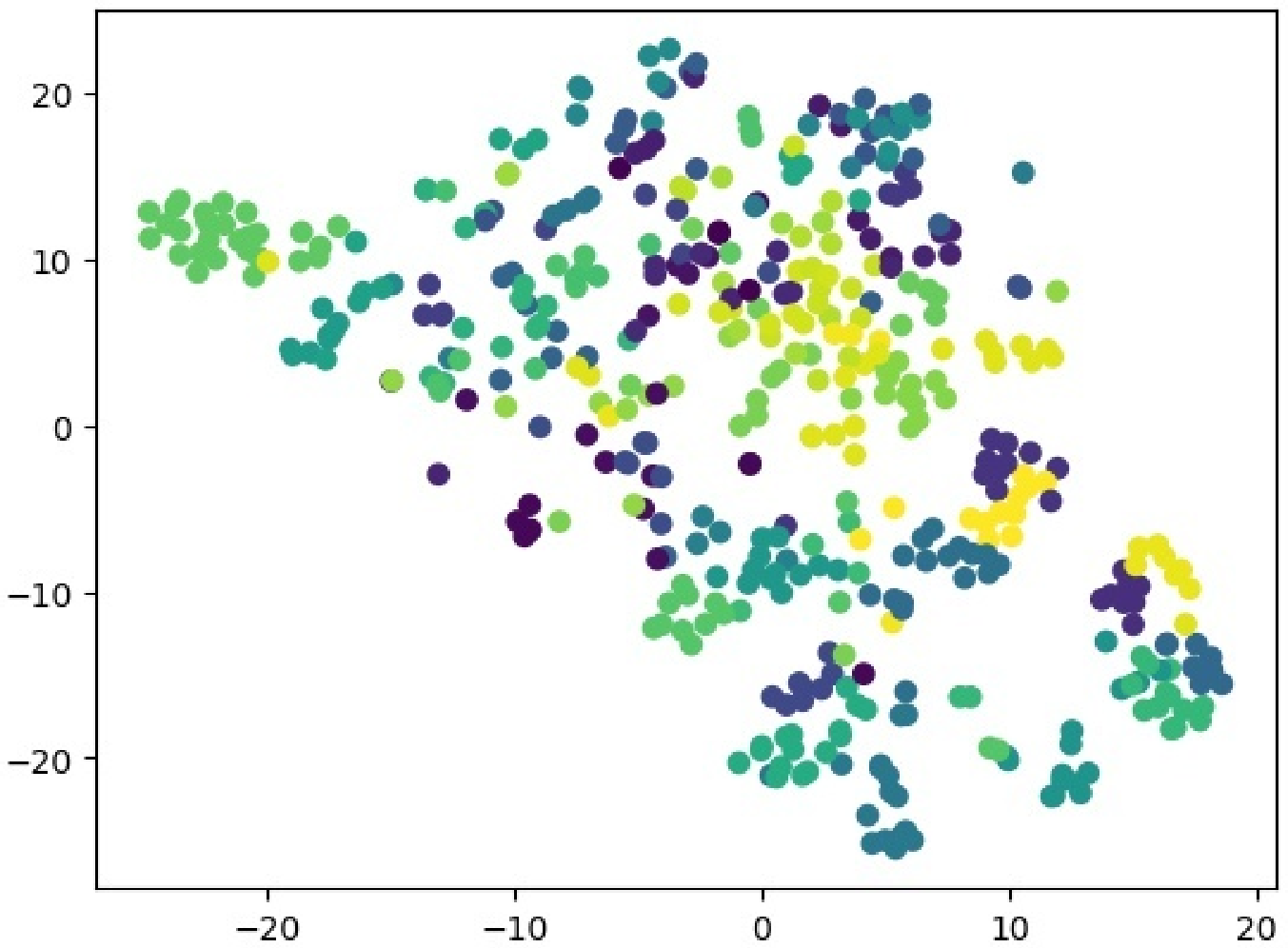}
		\caption{convLSTM}
        \label{fig:tsne_convLSTM}
        \end{subfigure}
        \begin{subfigure}[b]{0.32\textwidth}
        \includegraphics[scale=0.23]{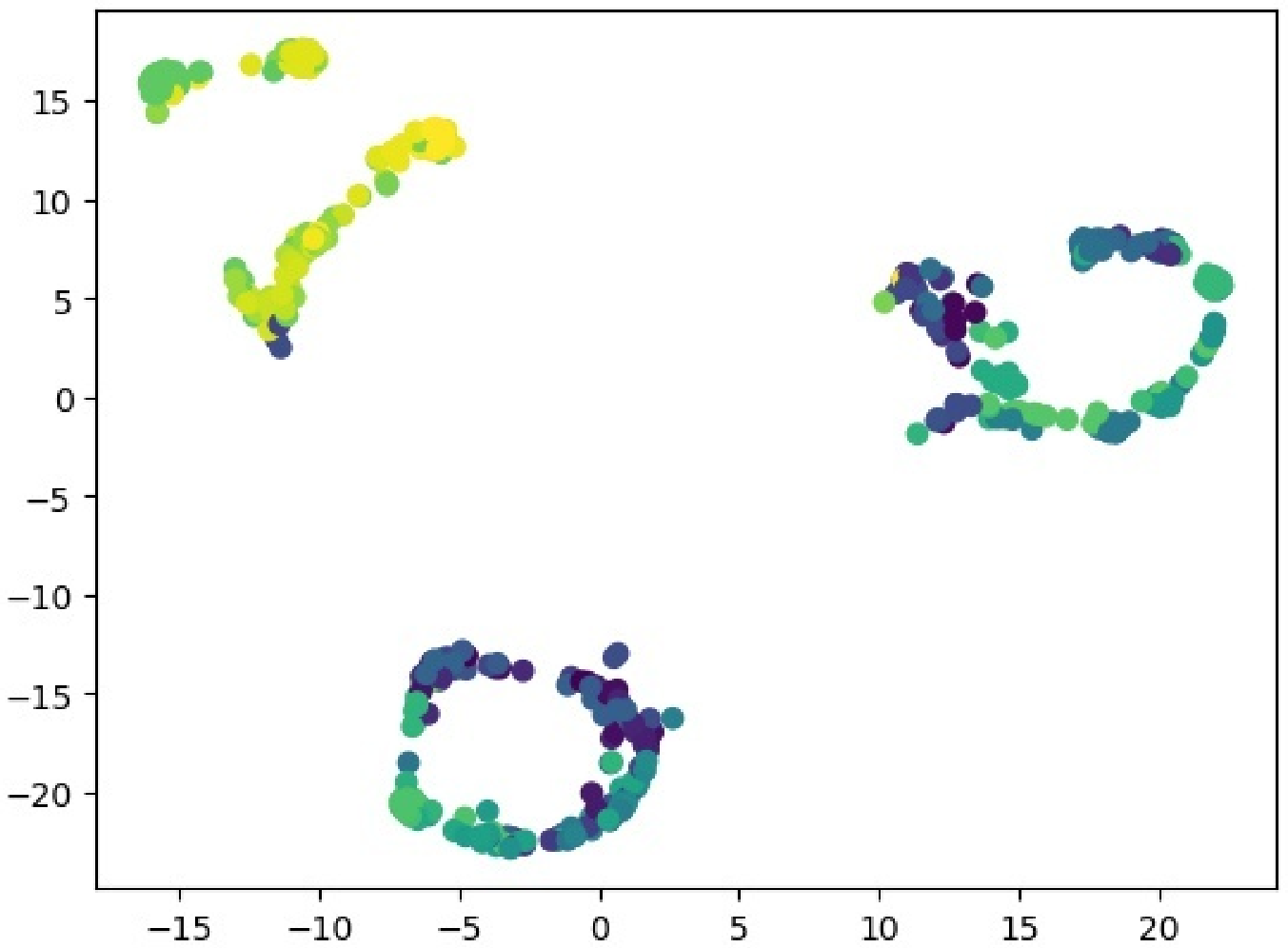}
		\caption{LRCN}
        \label{fig:tsne_LRCN}
        \end{subfigure}
        \begin{subfigure}[b]{0.32\textwidth}
        \includegraphics[scale=0.23]{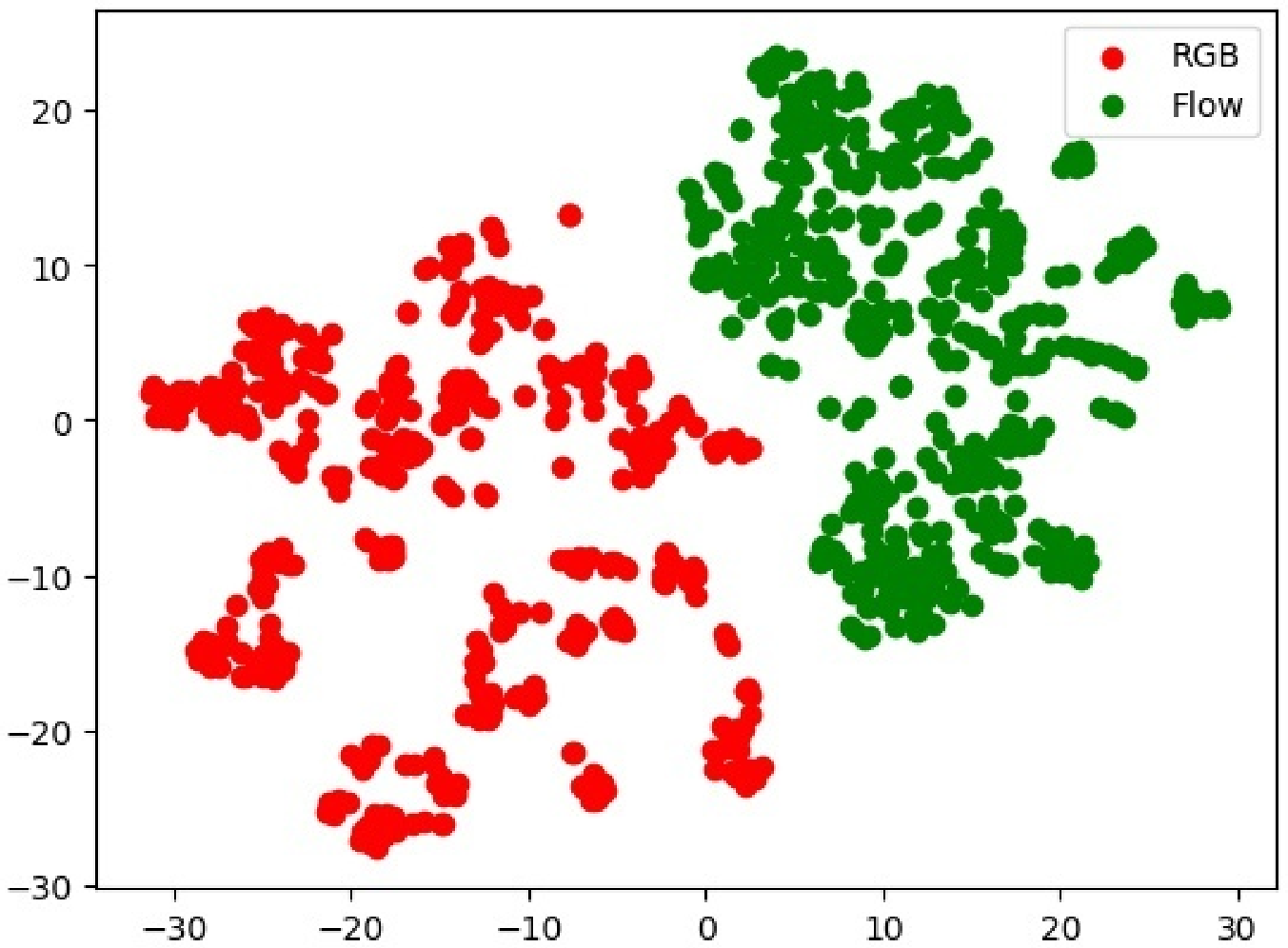}
		\caption{RGB vs Flow}
        \label{fig:tsne_rgb_flow}
        \end{subfigure}
   		\vspace*{2mm}
        \caption{t-SNE embedding of the features from (a) convLSTM model, (b) LRCN model and (c) spatial and temporal stream networks for videos from GTEA 61 dataset. In (a) and (b), features from videos of the same class have the same color.}
		\label{fig:tsne}
	\end{figure}

We first evaluated the performance of various configurations using the fixed split of GTEA 61 dataset. The results obtained are listed in Table~\ref{tab:conf_comp}. We first compare the performance of fully-connected LSTM with convLSTM. For the LSTM model, we follow the LRCN approach proposed by Donahue \etal \cite{donahue2015long} in which the output of the final convolutional layer of ResNet-34 is reshaped to obtain a feature descriptor describing the video frame. Feature descriptors thus generated from the frames of a video are then encoded using an LSTM with 512 hidden units. With the convLSTM module, as explained in section \ref{subsec:spat_att}, we are able to keep the spatial dimensions and a memory tensor is propagated across time which enables in capturing the evolution of spatio-temporal changes with time. The advantages of using convLSTM can also be validated by comparing the recognition performance obtained with the respective models. We also plot the t-SNE embedding \cite{maaten2008visualizing} of the features extracted from the output of both convLSTM and LRCN models in Figure~\ref{fig:tsne_convLSTM} and \ref{fig:tsne_LRCN}. In Figure~\ref{fig:tsne_convLSTM}, the features are more spread out and those from the videos belonging to the same class are clustered together. From these observations, we decide to use convLSTM as the temporal encoding module. Next, we analyzed the performance gain obtained by adding the proposed spatial attention mechanism and an improvement of 12\% is observed by making the network attend to the relevant regions in the video frames.

Camera motion is found to be one of the problems degrading the performance of action recognition systems. In order to combat its effects, action recognition~\cite{TSN2016ECCV} and egocentric action recognition methods \cite{li2015delving} propose to use stacked warp optical flow images in the temporal stream network. Warp optical flow is obtained by subtracting the camera motion from the optical flow. This is further validated by our experiments where an improvement of 4\% is gained by compensating the camera motion in the optical flow images and as a result we choose to apply stacked warp optical flow images in the temporal stream of the network.

As explained in section \ref{subsec:impl_det}, we perform two different approaches for the fusion of spatial and temporal stream network outputs. In Table~\ref{tab:conf_comp}, two-stream (average) denotes the fusion approach where the class scores obtained from each of the two streams are averaged and two-stream (joint train) shows the result obtained by adding a new fully-connected layer on top of the two individual network\rev{ followed by joint training of the fully-connected layers.}{s followed by finetuning up to Layer 4.} By following the joint training approach, a \rev{30}{10}
\% performance boost is gained. This can be explained with Figure~\ref{fig:tsne_rgb_flow} which plots the t-SNE embedding of the features obtained from the individual stream networks for videos from GTEA 61 dataset. From the Figure, it can be seen that the spatial stream and temporal stream features occupy in different regions of the feature space which indicates that they possess some complementary information that can assist in identifying one activity class from another. This resulted in the improved performance in the case of joint training since the network is allowed to learn to combine these complementary features in a way that each activity class becomes more discernible from one another. 

\begin{table}[t]
	\begin{center}
    \begin{tabular}{|c|c|c|c|c|}
		\hline
		  Methods & \parbox{0.8in}{\centering GTEA 61$^*$} & \parbox{0.8in}{\centering GTEA 61} & \parbox{0.8in}{\centering GTEA 71} & \parbox{0.8in}{\centering GTEA Gaze+} \\
         \hline
         Li \etal \cite{li2015delving}$^{**}$ & 66.8 & 64 & 62.1 & \rev{57.4}{60.5} \\
         \hline
         Ma \etal \cite{ma2016deeper}$^{**}$ & 75.08 & 73.02 & 73.24 & {\bf 66.4} \\
         \hline \hline
         Two stream \cite{simonyan2014two} & 57.64 & 51.58 & 49.65 & 58.77 \\
         \hline
         TSN \cite{TSN2016ECCV}  & 67.76 & 69.33 & 67.23 & 55.25 \\ 
         \hline \hline
          \textbf{Ours} & \rev{97.41}{\bf 77.59} & \rev{95.62}{\bf 79} & \rev{98.48}{\bf 77} & \rev{71.8}{60.13} \\
                               \hline
	\end{tabular}
    \end{center}
    \caption{Comparison with state-of-the-art methods on popular egocentric datasets, we report recognition accuracy in \%. ($^*$: fixed split\rev{}{; $^{**}$: trained with strong supervision})}
	\label{tab:res_table}
\end{table}

\subsection{Comparison with state-of-the-art techniques}
\label{subsec:comp}

Table \ref{tab:res_table} and \ref{tab:egtea} compares the proposed method with state-of-the-art techniques. The first block in Table~\ref{tab:res_table} shows methods specifically proposed for egocentric activity recognition, the second block shows methods proposed for third person action recognition. Regarding Table~\ref{tab:egtea}, EGTEA Gaze+ is a recently developed large scale egocentric video dataset comprising of 10325 activity instances from 106 classes. As the dataset is relatively new and no standard benchmarks are available, we compare recognition performance with state-of-the-art deep learning techniques for third person action recognition. Since this dataset could emerge as a future benchmark in egocentric activity recognition, we list the recognition accuracy obtained for each of the provided test-train splits as well in Table~\ref{tab:egtea} for serving as a baseline in future research. All the compared methods in Table~\ref{tab:res_table} and \ref{tab:egtea} use RGB images as well as optical flow images in order to classify the videos into the corresponding class categories.

\begin{table}[t]
\begin{center}
\begin{tabular}{|c|c|c|c|c|}
\hline
Methods & \parbox{0.8in}{\centering Split 1} & \parbox{0.8in}{\centering Split 2} & \parbox{0.8in}{\centering Split 3} & \parbox{0.8in}{\centering Average} \\
\hline
Two Stream$^*$ \cite{simonyan2014two} & 43.78 & 41.47 & 40.28 & 41.84 \\ 
\hline
I3D$^*$ \cite{carreira2017quo} & 54.19 & 51.45 & 49.41 & 51.68 \\
\hline
TSN \cite{TSN2016ECCV} & 58.01 & 55.01 & 54.78 & 55.93 \\
\hline \hline
\textbf{Ours} & \rev{81.90}{\bf 62.17} & \rev{79.08}{\bf 61.47} & \rev{79.12}{\bf 58.63} & \rev{80.03}{\bf 60.76} \\
\hline
\end{tabular}
\end{center}
\caption{Recognition accuracy (in \%) obtained for different train-test splits of EGTEA Gaze+ dataset compared against state-of-the-art techniques. (*: Result provided by the dataset developers)}
\label{tab:egtea}
\end{table}

The method proposed by Li \etal \cite{li2015delving} uses hand segmentation for the first two datasets along with gaze information in the case of Gaze+ dataset for detecting the location of the objects being handled, while Ma \etal \cite{ma2016deeper} trains a network for hand segmentation and object localization for obtaining explicit information about the objects and their location. In the proposed method, we make use of the prior knowledge inflated in the network to identify the location of the object that is relevant in identifying the activity class. From both Table~\ref{tab:res_table} and \ref{tab:egtea}, we can see that the proposed method achieves a notable performance boost, except in the case of Gaze+ dataset, which anyway validates its efficacy in performing egocentric activity recognition. For Gaze+ dataset, the proposed method is resulting in lower performance than the methods proposed in \cite{ma2016deeper} and \cite{li2015delving}. This may be attributed to the nature of the dataset. In Gaze+ dataset, objects ``spoon'', ``fork'' and ``knife'' are combined to an aggregated object ``spoonForkKnife'' and objects ``cup'', ``plate'' and ``bowl'' are combined as ``cupPlateBowl''. Their corresponding activities are clubbed together as a single activity class. For example, activities such as ``take knife'', ``take fork'' and ``take spoon'' are combined to form a single activity class titled ``take spoonForkKnife''. In the proposed method, the network is trained to identify the relevant object's location from the activity label alone, and combining different objects to form a new \rev{pseudo}{meta}-object class weakens the correlation between activity class label and the visual information present in the frame. It therefore hampers finetuning of activation mapping during learning. In the supplementary material we provide confusion matrices for the worst-performing splits, to confirm that the proposed method struggles to classify the activities involving meta-object labels.
\rev{This prevents the proposed method from utilizing the semantics of the activity label for identifying relevant objects related to the activity under consideration.}{} 
In the case of methods proposed by Ma \etal \cite{ma2016deeper} and Li \etal \cite{li2015delving}, the network is trained using explicit supervision in the form of object instance location, thereby resulting in improved performance compared to the proposed method. It should also be noted that a significant improvement is observed when comparing the proposed method with Two-Stream~\cite{simonyan2014two} on EGTEA Gaze+ dataset which subsumes Gaze+ dataset. In EGTEA Gaze+ dataset the above mentioned objects are not combined together to form a single class and the network more easily succeeds in identifying the regions in the video frames related to the activity.

\begin{figure}[t]
		\centering      
        \begin{subfigure}[b]{0.47\textwidth}
			\includegraphics[scale=0.25]{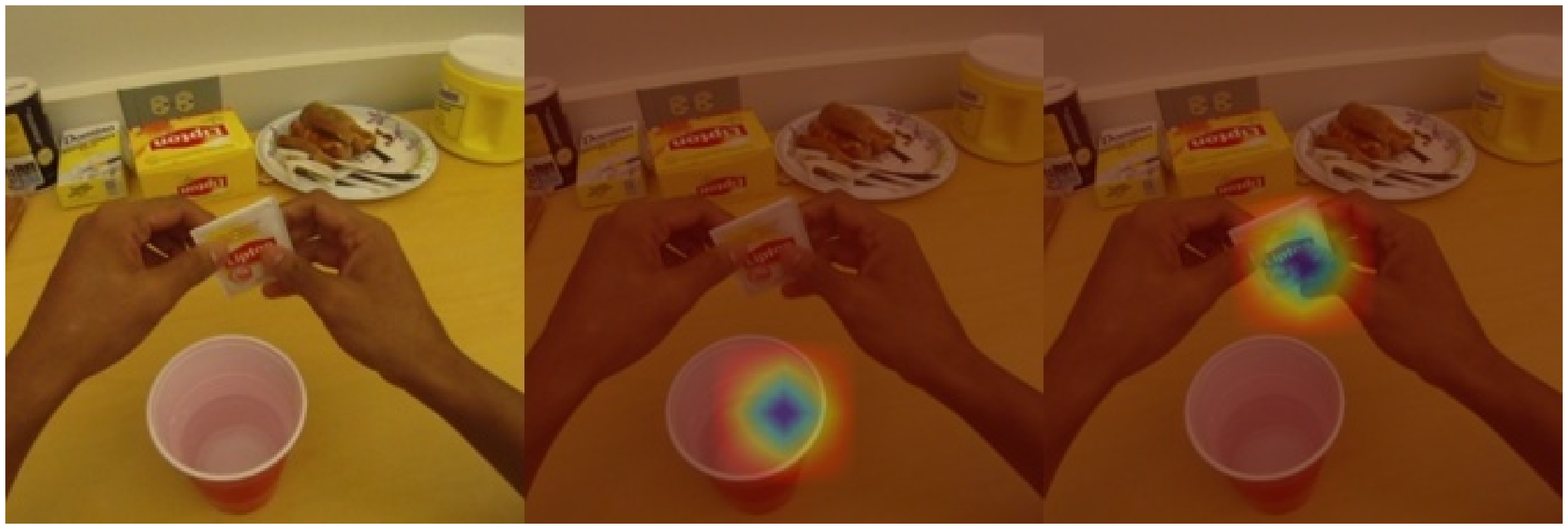}
            \caption{Open tea}
			\label{fig_ex_comp1}
		\end{subfigure}
        \begin{subfigure}[b]{0.47\textwidth}
			\includegraphics[scale=0.25]{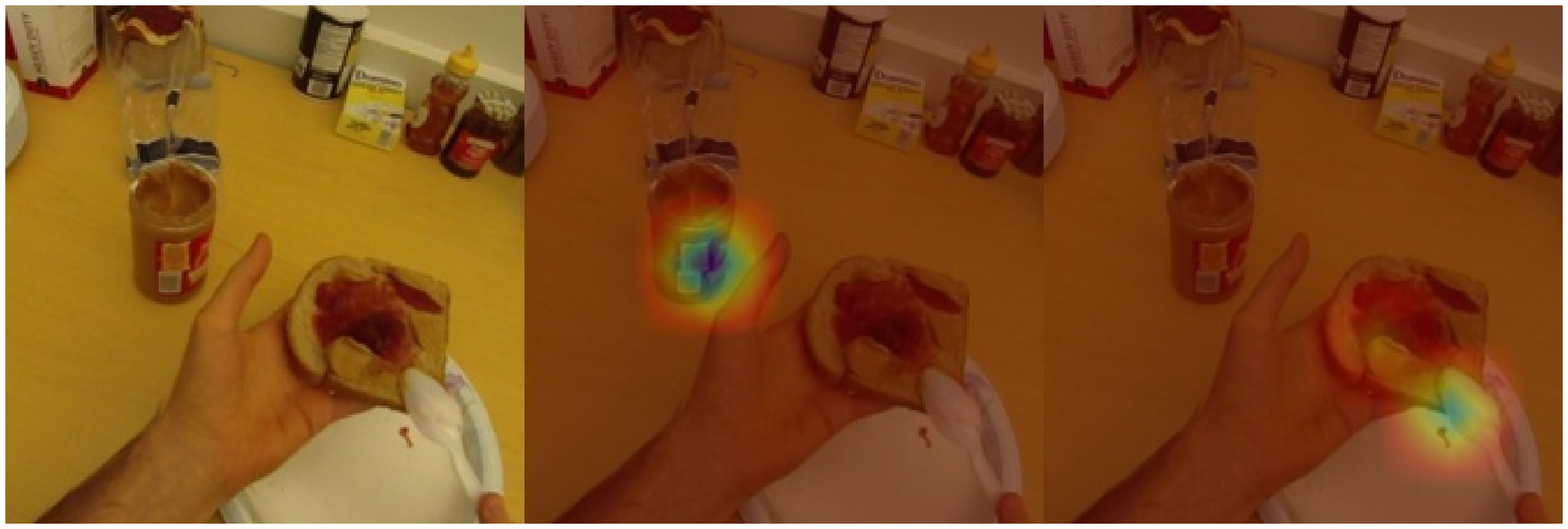}
            \caption{Spread peanut}
			\label{fig_ex_comp2}
		\end{subfigure}\\
        \vskip .5mm
        \begin{subfigure}[b]{0.47\textwidth}
			\includegraphics[scale=0.25]{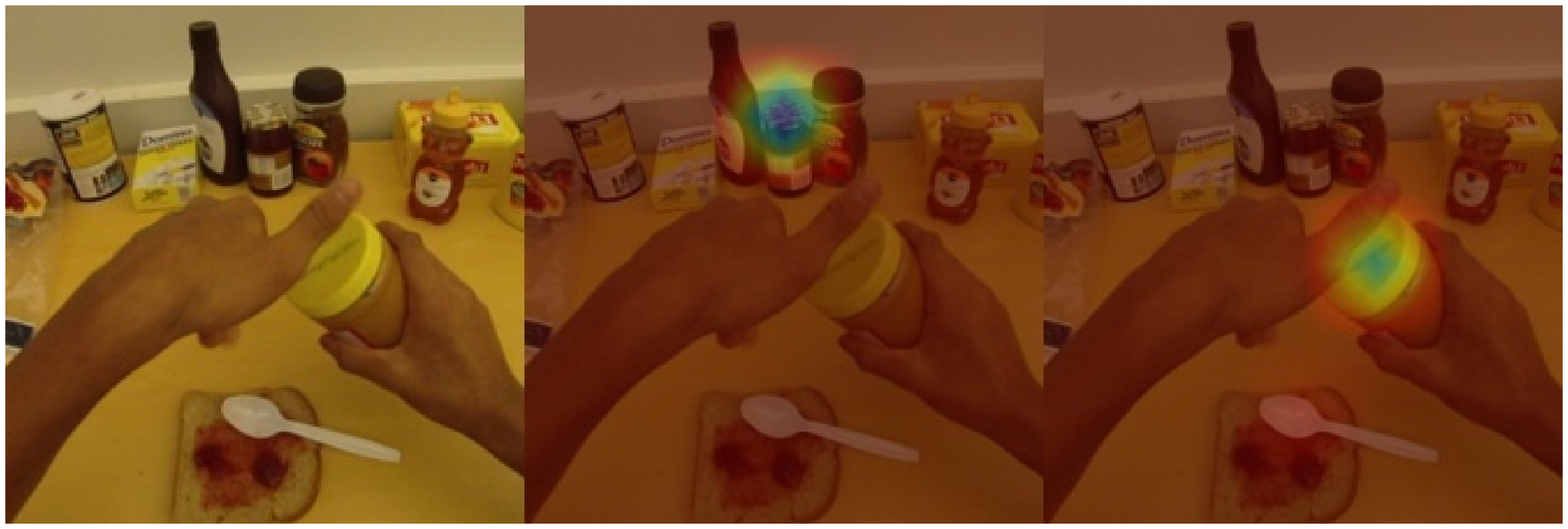}
            \caption{Open peanut}
			\label{fig_ex_comp3}
		\end{subfigure}
        \begin{subfigure}[b]{0.47\textwidth}
			\includegraphics[scale=0.25]{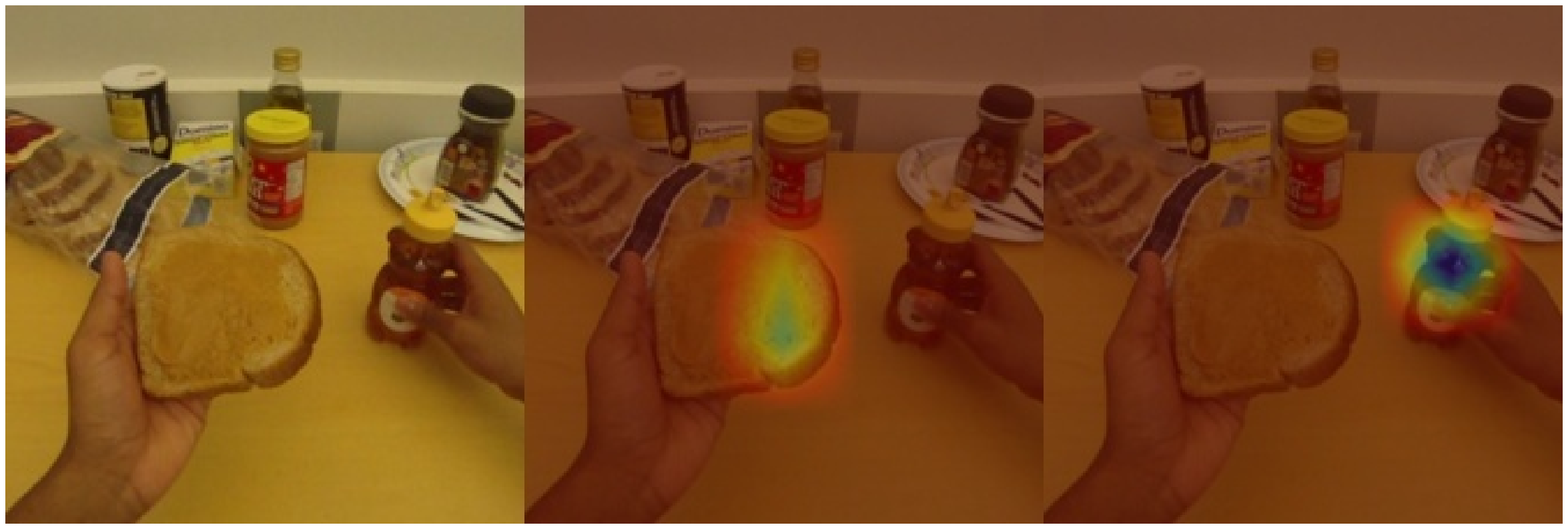}
            \caption{Take honey}
			\label{fig_ex_comp4}
		\end{subfigure}
		\vspace*{2mm}
        \caption{Spatial attention maps obtained for frames from GTEA 61 dataset. First image shows the original frame, second image shows the attention map generated with ResNet-34 trained on imagenet and the last image shows the attention map obtained using the network trained for activity recognition (after stage 2 training).}
		\label{fig:fig_ex_comp}
	\end{figure}

In Figure~\ref{fig:fig_ex}, we saw that a generic image recognition prior inflated in the network need not be representative of the activity class all the time. In order to compensate this problem, we train the fully-connected layer together with the convolutional layers of the final layer of ResNet (orange blocks in Figure~\ref{fig:block_dia}) in stage 2. Figure~\ref{fig:fig_ex_comp} shows the spatial attention map obtained after this second stage of training. From the Figure, we can see that the network learns to correctly identify the object in the images that is representative of the activity class under consideration. More examples are provided in the supplementary material.

Another point worth mentioning is that the state-of-the-art-techniques for action recognition from third person videos, listed in Table~\ref{tab:res_table} and \ref{tab:egtea}, are performing sub-par compared to the methods proposed for egocentric videos. This shows that these methods cannot be considered as standardized techniques for action recognition from videos in general and underlines the importance of developing methods tailored for egocentric videos.

\section{Conclusion}
\label{sec:conc}

In this work, we have successfully developed a novel deep learning approach for egocentric activity recognition. We validate the performance of the proposed method on four standard egocentric activity datasets and the proposed method achieved state-of-the-art recognition performance. The proposed approach takes advantage of the prior information encoded in a convolutional neural network pre-trained for image classification to generate object specific spatial attention. The spatially attended image features are then temporally encoded using a convolutional long short-term memory module. Detailed analysis show that the proposed spatial attention mechanism is able to correctly locate the objects that are representative of the activity class under consideration. We have also performed a detailed ablation analysis on the various network configurations
. As a future work, we will explore the possibility of adding a temporal attention mechanism since not all frames present in a video are equally representative of the concerned activity. \rev{}{We will further evaluate our method on the recently introduced EPIC-Kitchens dataset~\cite{Damen2018EPICKITCHENS} and on video recognition problems from third-person views}.


\bibliography{egbib}

\newpage

\section*{Supplementary material}

This supplementary material of our BMVC submission shows: 
\begin{itemize}
	\item More examples of the proposed spatial attention map generation technique
	\item The confusion matrix obtained for subjects P1 and P3 of the GTEA Gaze+ dataset
\end{itemize}

\section{GTEA 61 attention maps}
\label{sec:2}
Here, we show additional images with the attention map generated by the network as discussed in section 4.3 of the paper. The videos are from the GTEA 61 dataset mentioned in our paper. In the figures, each column consists of the frames extracted from the same video. In each image, the first one represents the input image, second one, the attention map obtained from the imagenet pre-trained ResNet-34 network and the third image shows the attention map obtained after the proposed fine-tuning technique(after stage 2 training of our network). From the figures, it can be seen that the network learns to attend to the relevant objects that characterize each activity and gets improved after the fine-tuning step. 
\begin{figure}[h]
	\centering      
	\begin{subfigure}[b]{0.32\textwidth}
		\includegraphics[scale=0.17]{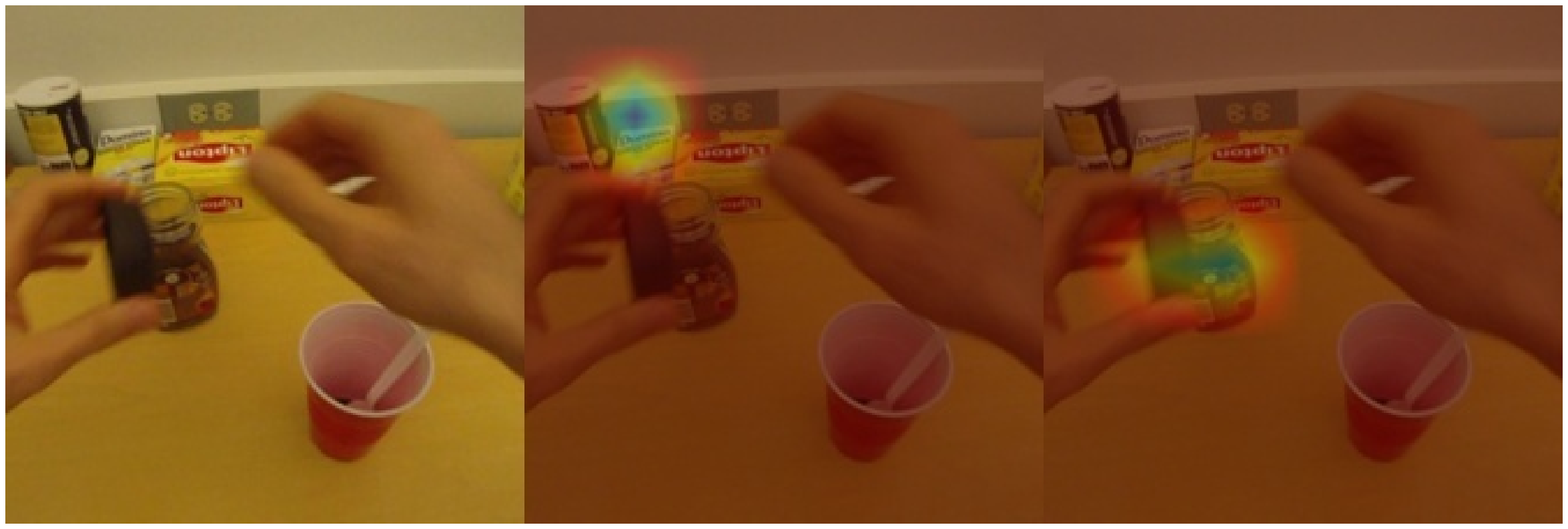}
	\end{subfigure}
	\begin{subfigure}[b]{0.32\textwidth}
		\includegraphics[scale=0.17]{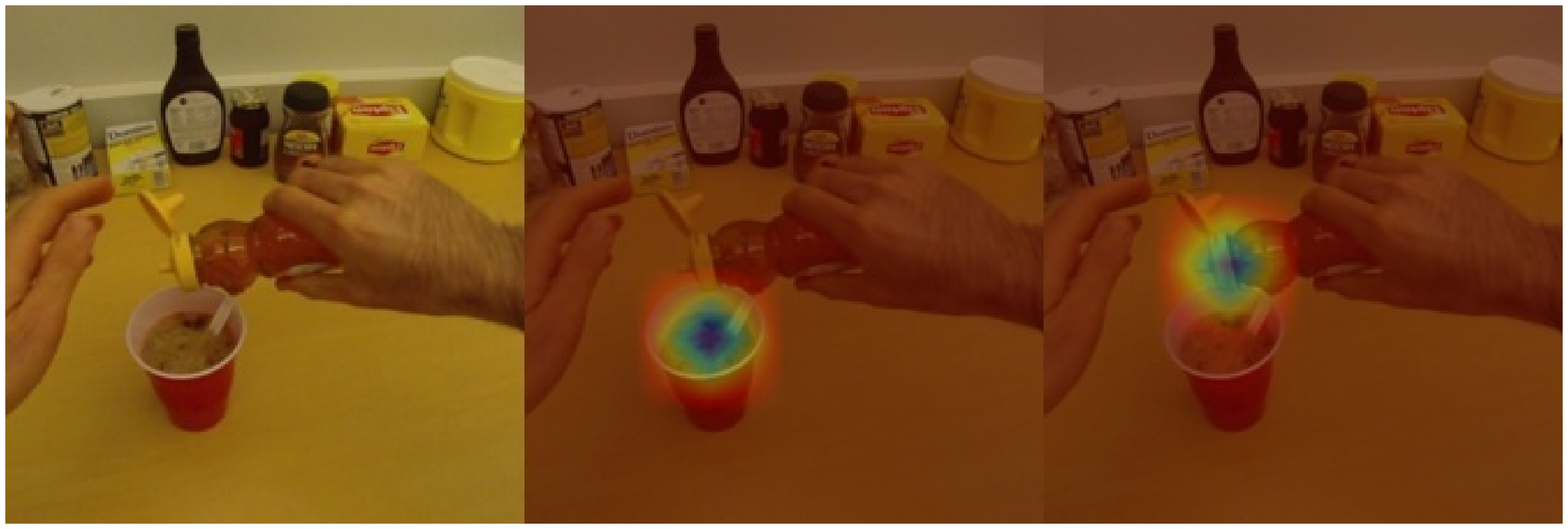}
	\end{subfigure}
	\begin{subfigure}[b]{0.32\textwidth}
		\includegraphics[scale=0.17]{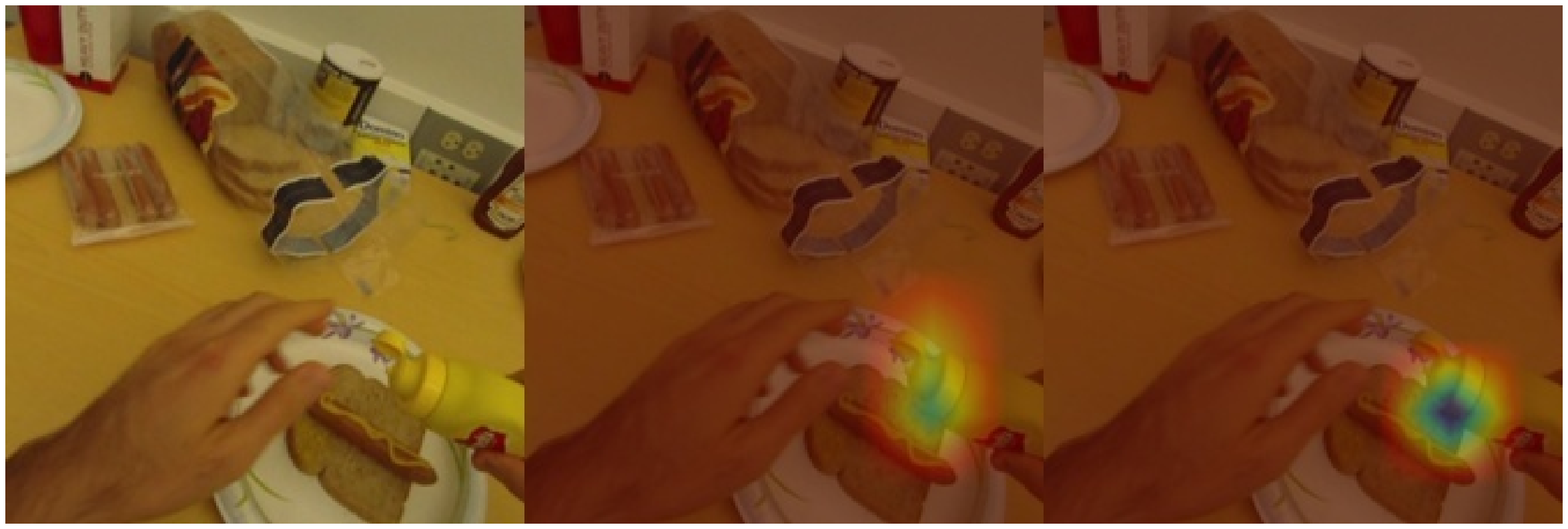}
	\end{subfigure}\\
	\vskip 2mm
	\begin{subfigure}[b]{0.32\textwidth}
		\includegraphics[scale=0.17]{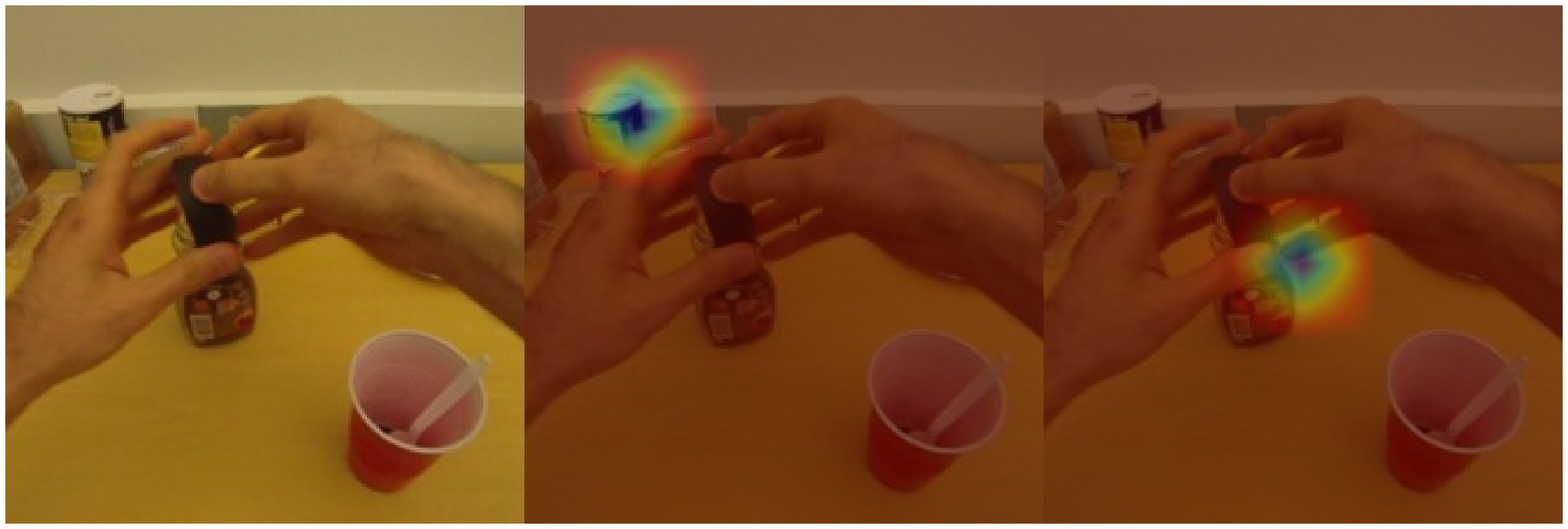}
	\end{subfigure}
	\begin{subfigure}[b]{0.32\textwidth}
		\includegraphics[scale=0.17]{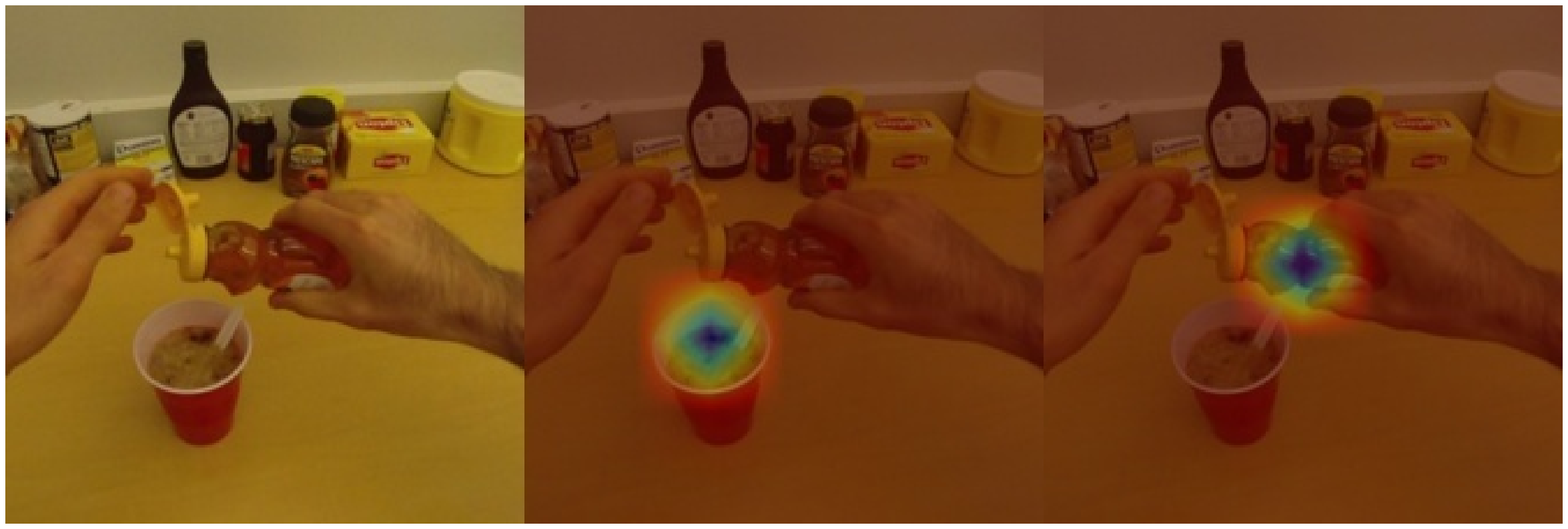}
	\end{subfigure}
	\begin{subfigure}[b]{0.32\textwidth}
		\includegraphics[scale=0.17]{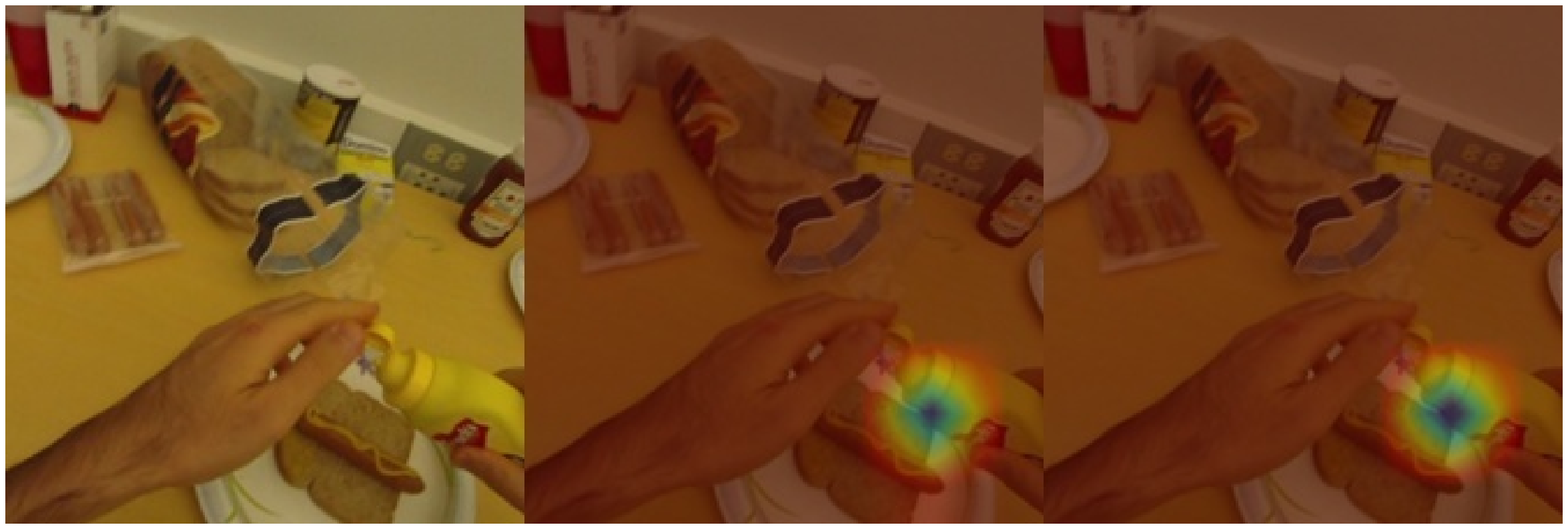}
	\end{subfigure}\\
	\vskip 2mm
	\begin{subfigure}[b]{0.32\textwidth}
		\includegraphics[scale=0.17]{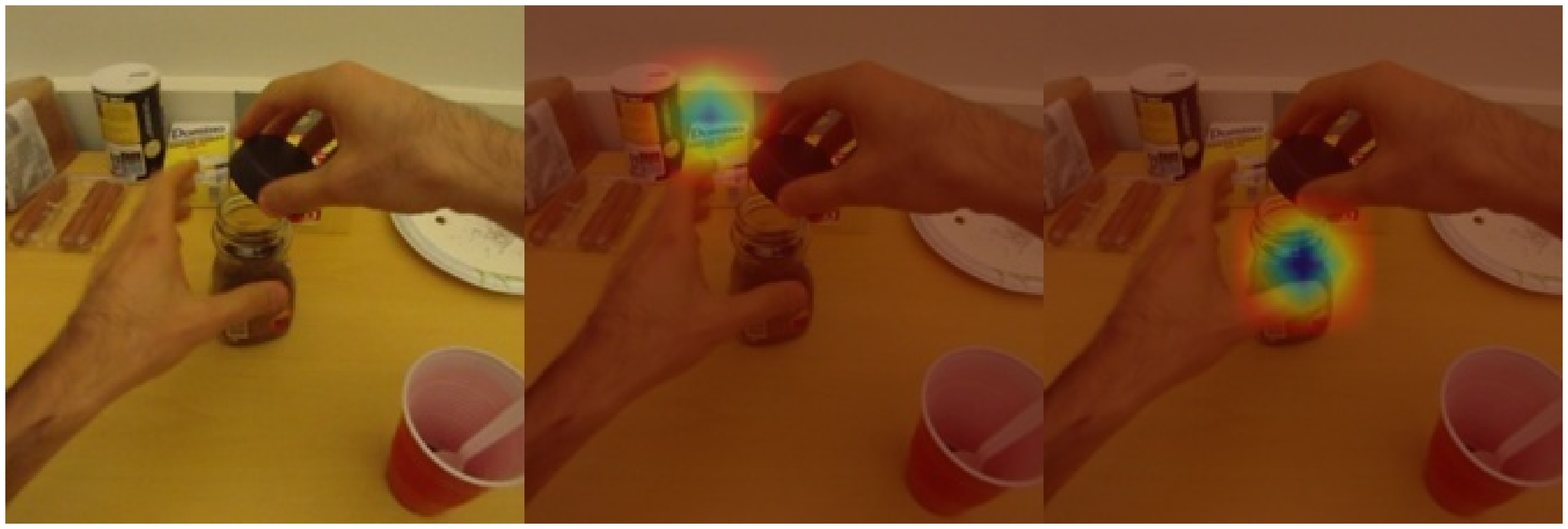}
	\end{subfigure}
	\begin{subfigure}[b]{0.32\textwidth}
		\includegraphics[scale=0.17]{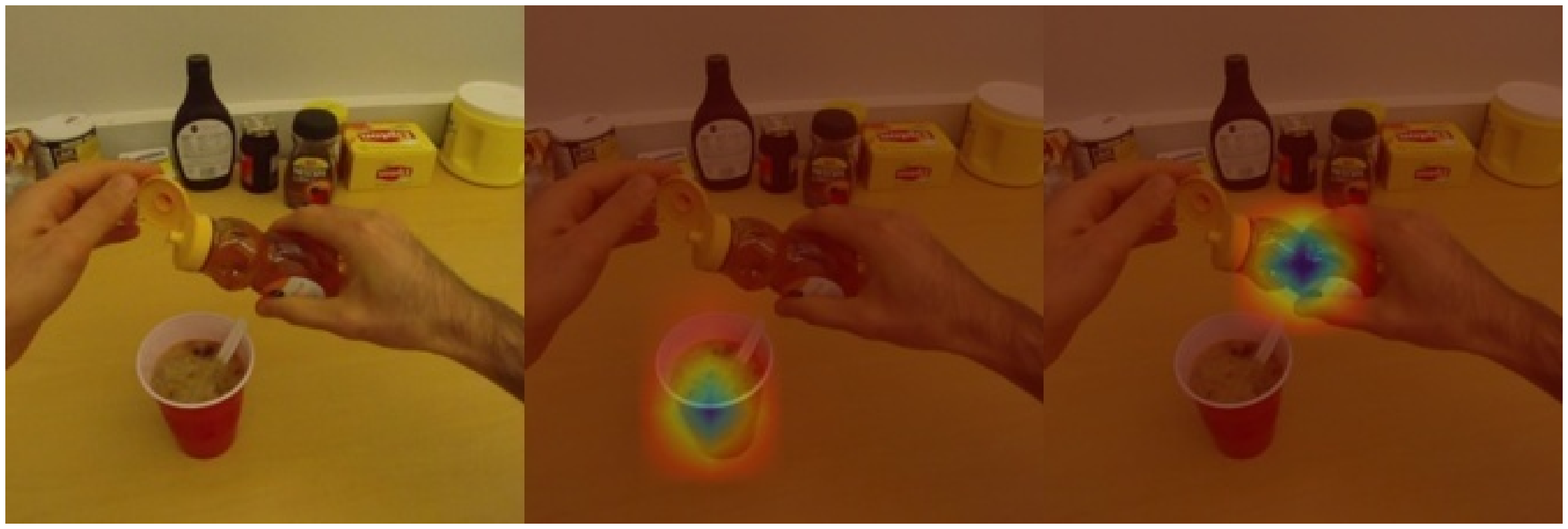}
	\end{subfigure}
	\begin{subfigure}[b]{0.32\textwidth}
		\includegraphics[scale=0.17]{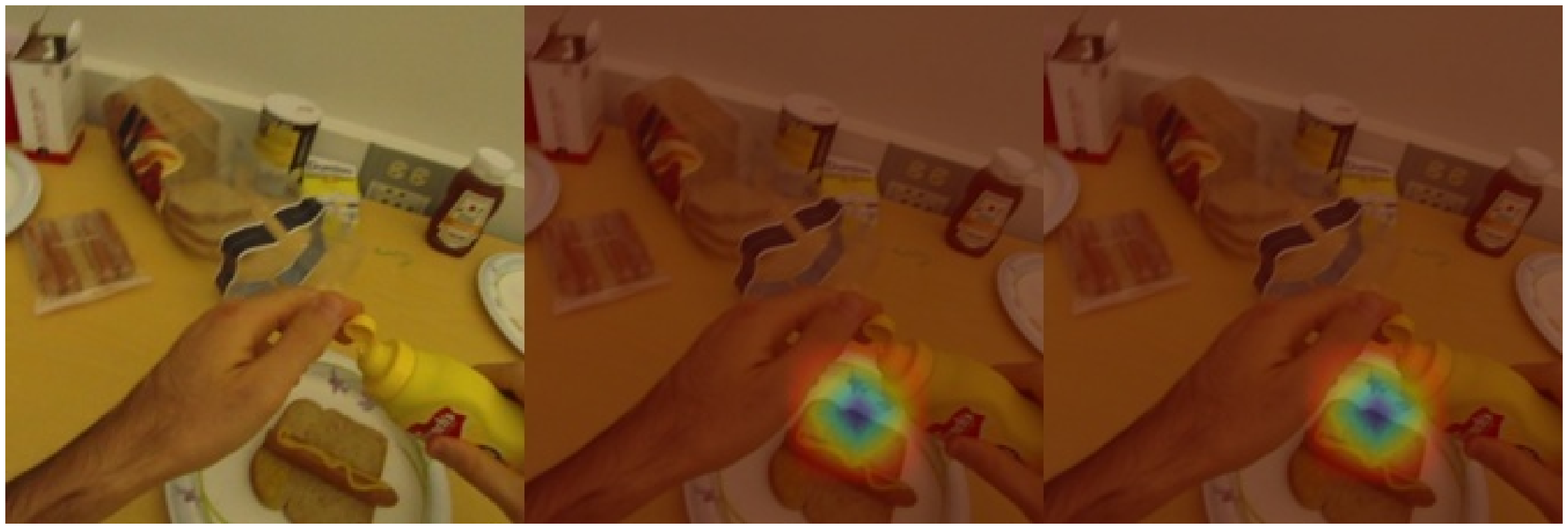}
	\end{subfigure}\\
	\vskip 2mm
	\begin{subfigure}[b]{0.32\textwidth}
		\includegraphics[scale=0.17]{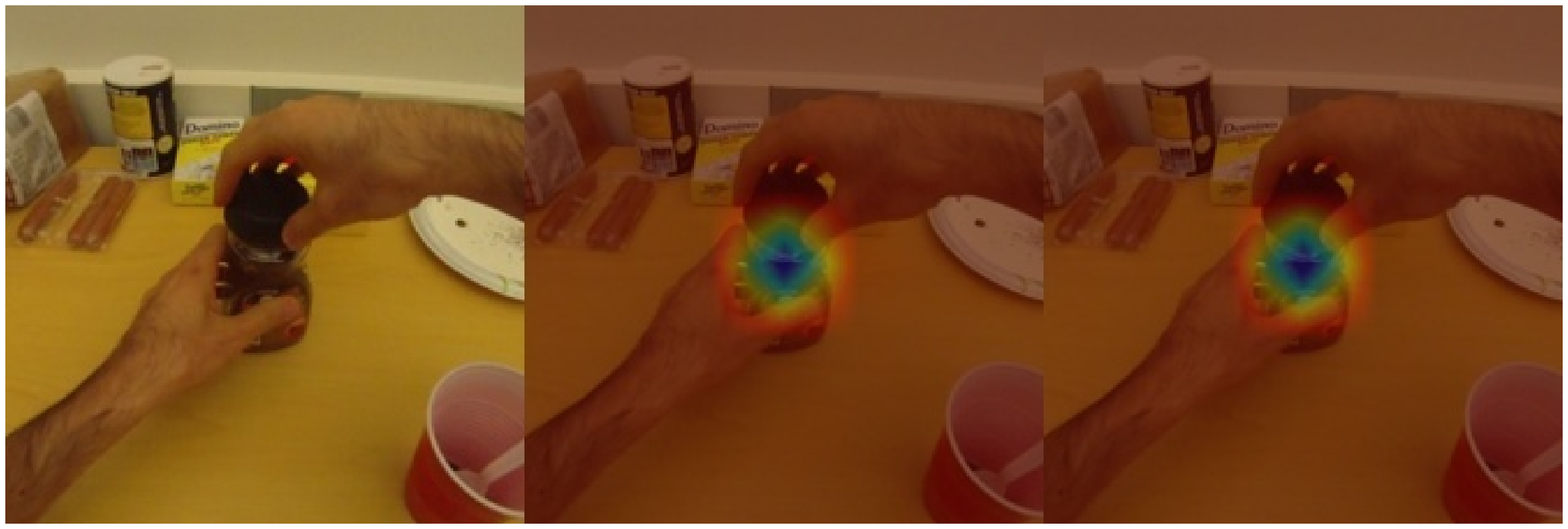}
	\end{subfigure}
	\begin{subfigure}[b]{0.32\textwidth}
		\includegraphics[scale=0.17]{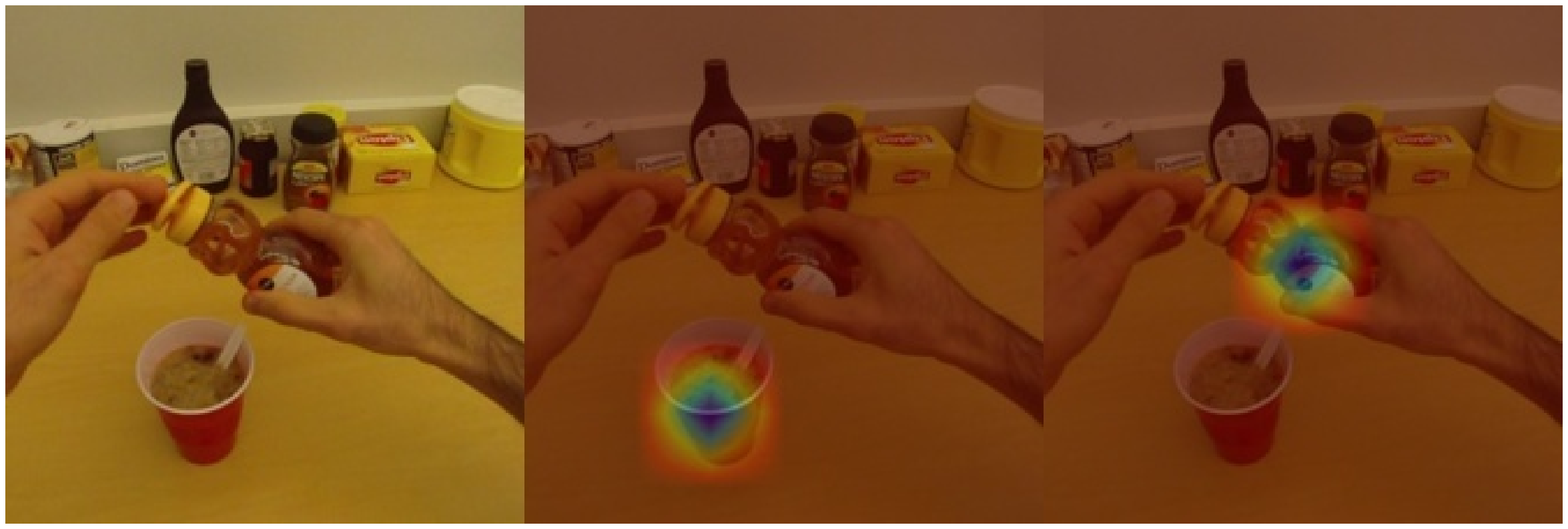}
	\end{subfigure}
	\begin{subfigure}[b]{0.32\textwidth}
		\includegraphics[scale=0.17]{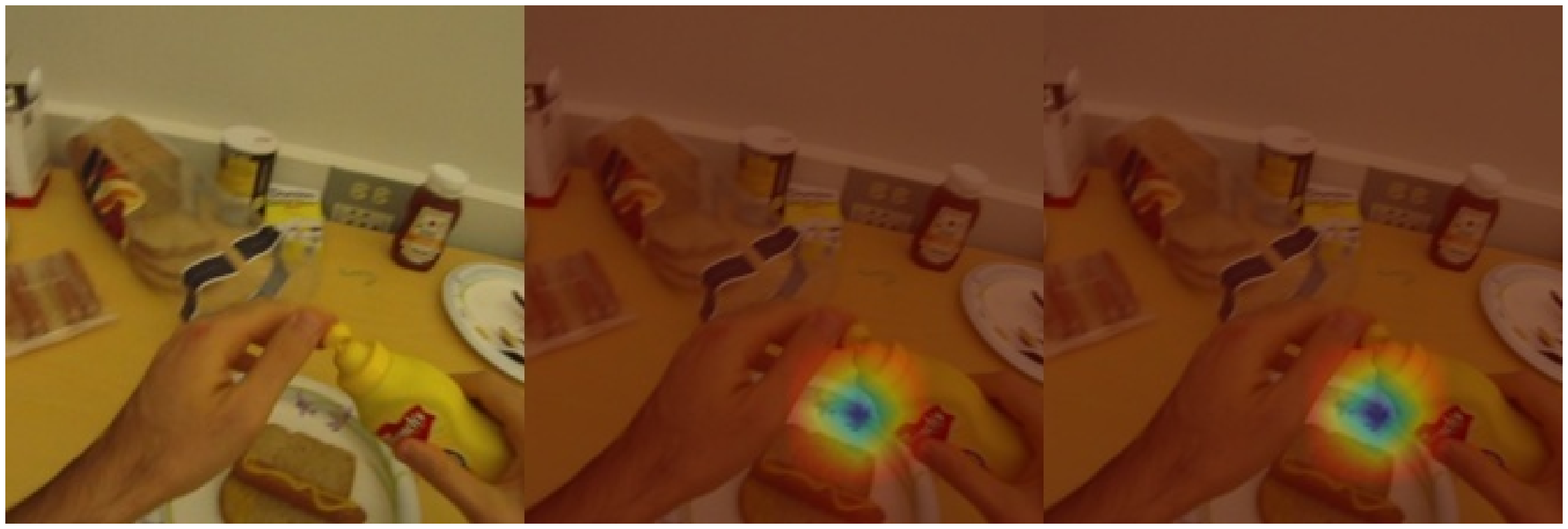}
	\end{subfigure}\\
	\vskip 2mm
	\begin{subfigure}[b]{0.32\textwidth}
		\includegraphics[scale=0.17]{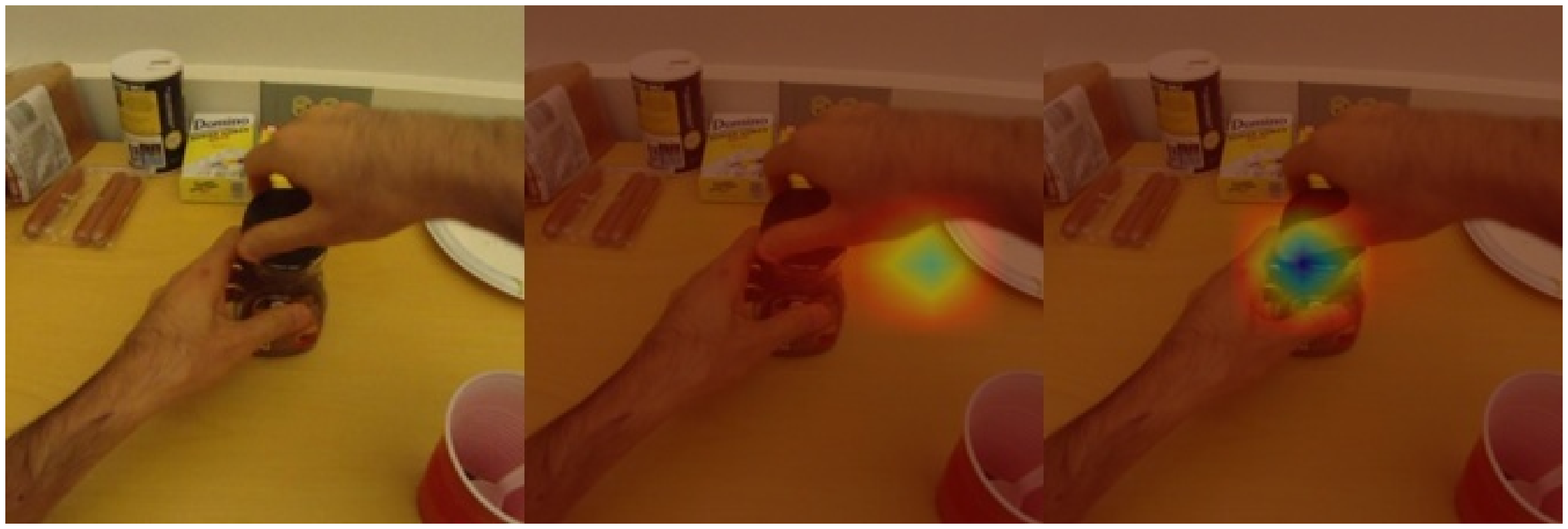}
	\end{subfigure}
	\begin{subfigure}[b]{0.32\textwidth}
		\includegraphics[scale=0.17]{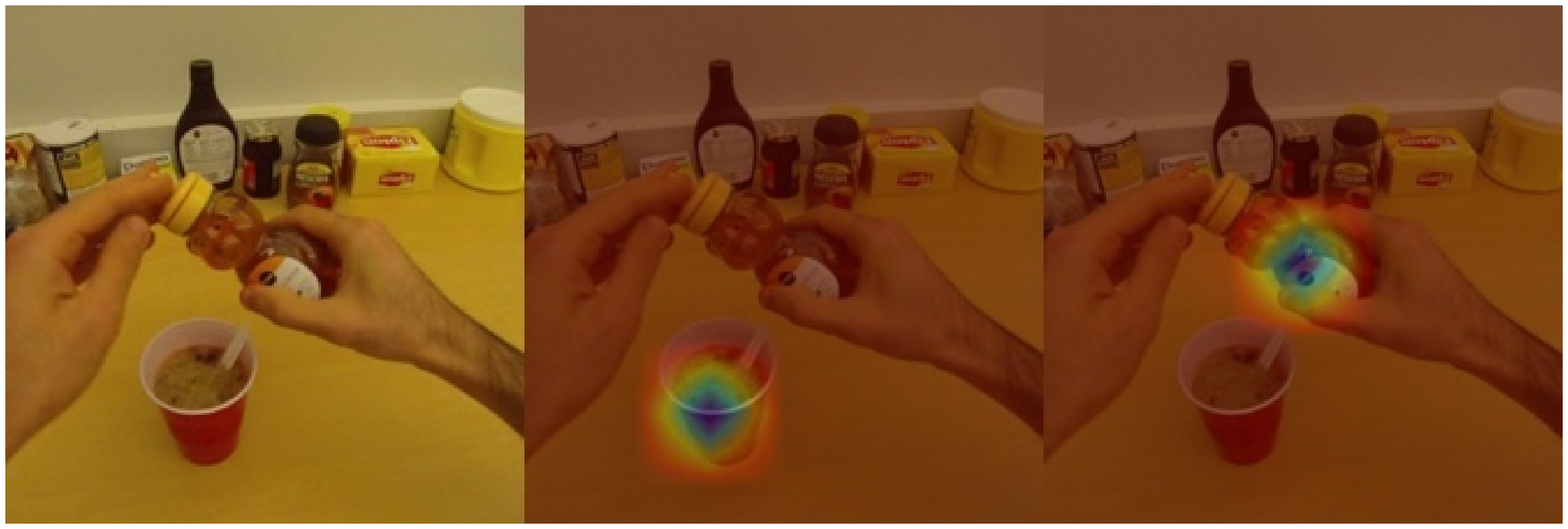}
	\end{subfigure}
	\begin{subfigure}[b]{0.32\textwidth}
		\includegraphics[scale=0.17]{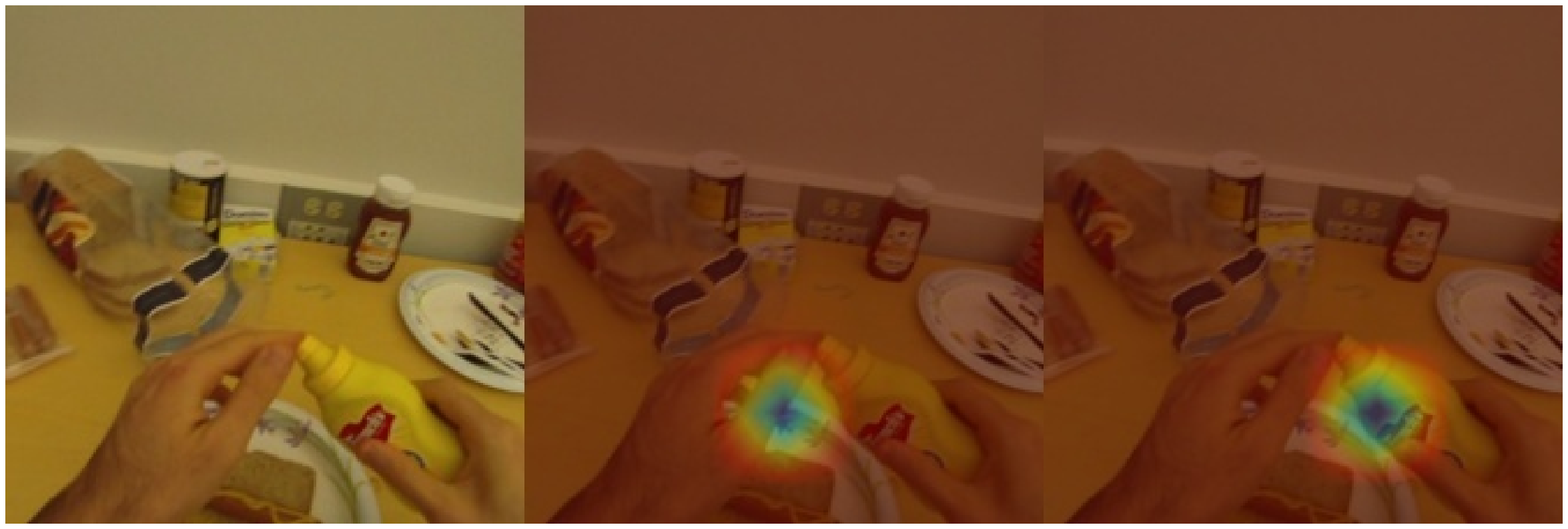}
	\end{subfigure}\\
	\vskip 2mm
	\begin{subfigure}[b]{0.32\textwidth}
		\includegraphics[scale=0.17]{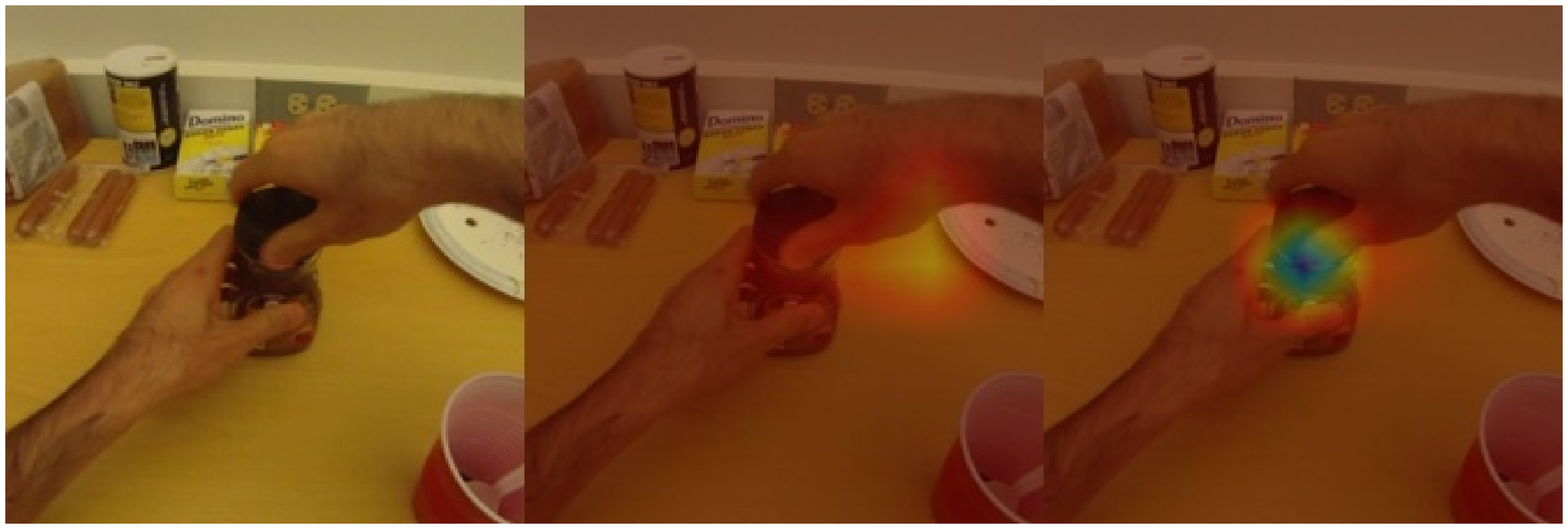}
	\end{subfigure}
	\begin{subfigure}[b]{0.32\textwidth}
		\includegraphics[scale=0.17]{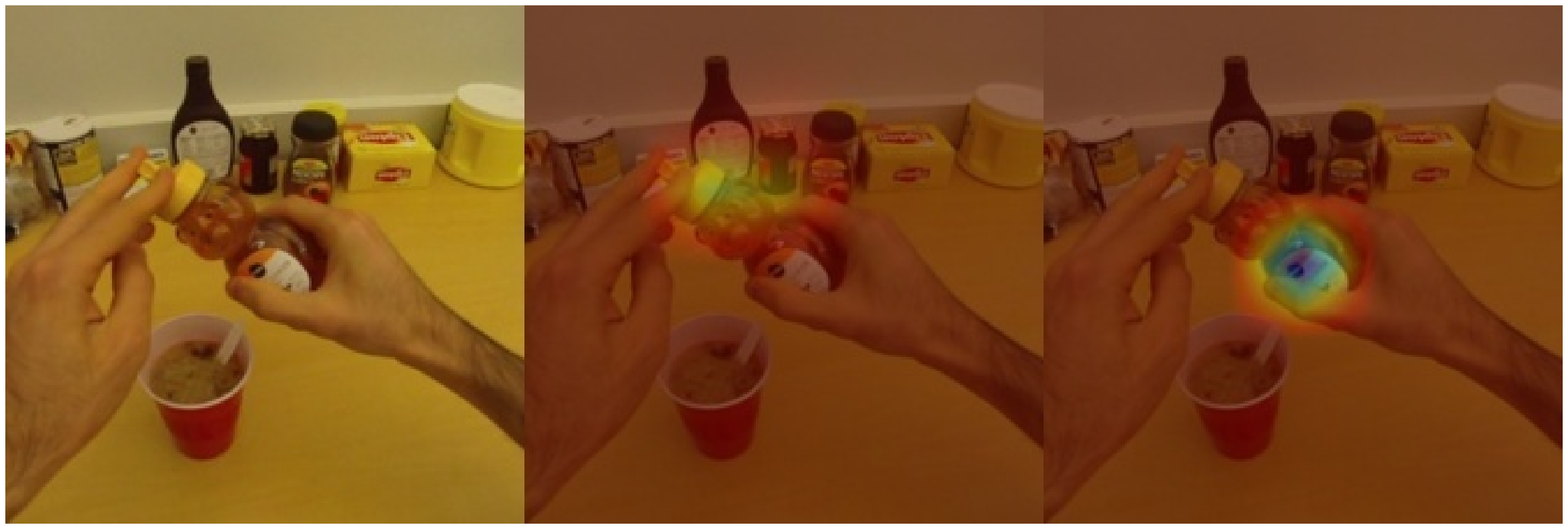}
	\end{subfigure}
	\begin{subfigure}[b]{0.32\textwidth}
		\includegraphics[scale=0.17]{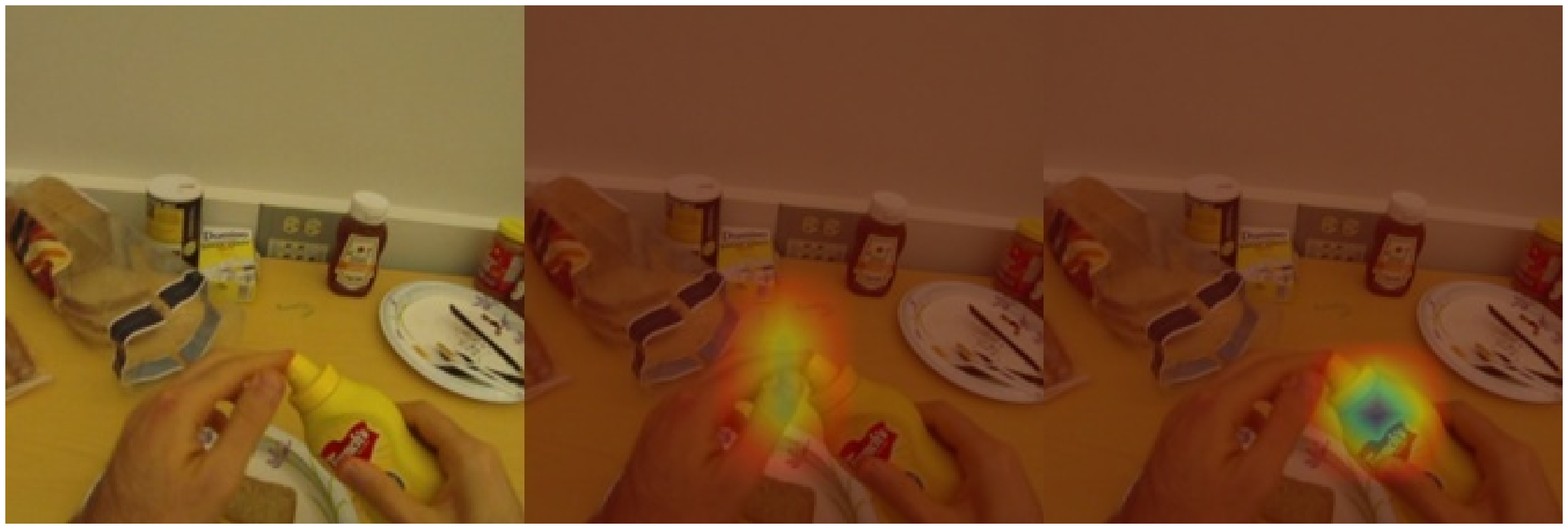}
	\end{subfigure}\\
	\vskip 2mm
	\begin{subfigure}[b]{0.32\textwidth}
		\includegraphics[scale=0.17]{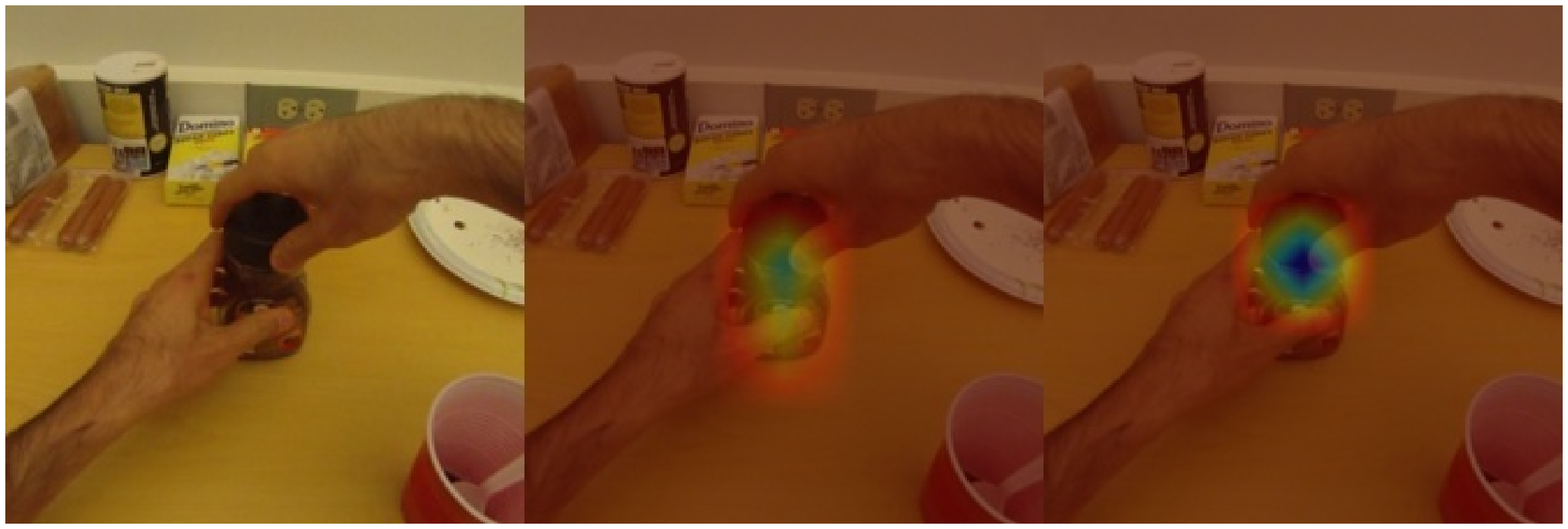}
	\end{subfigure}
	\begin{subfigure}[b]{0.32\textwidth}
		\includegraphics[scale=0.17]{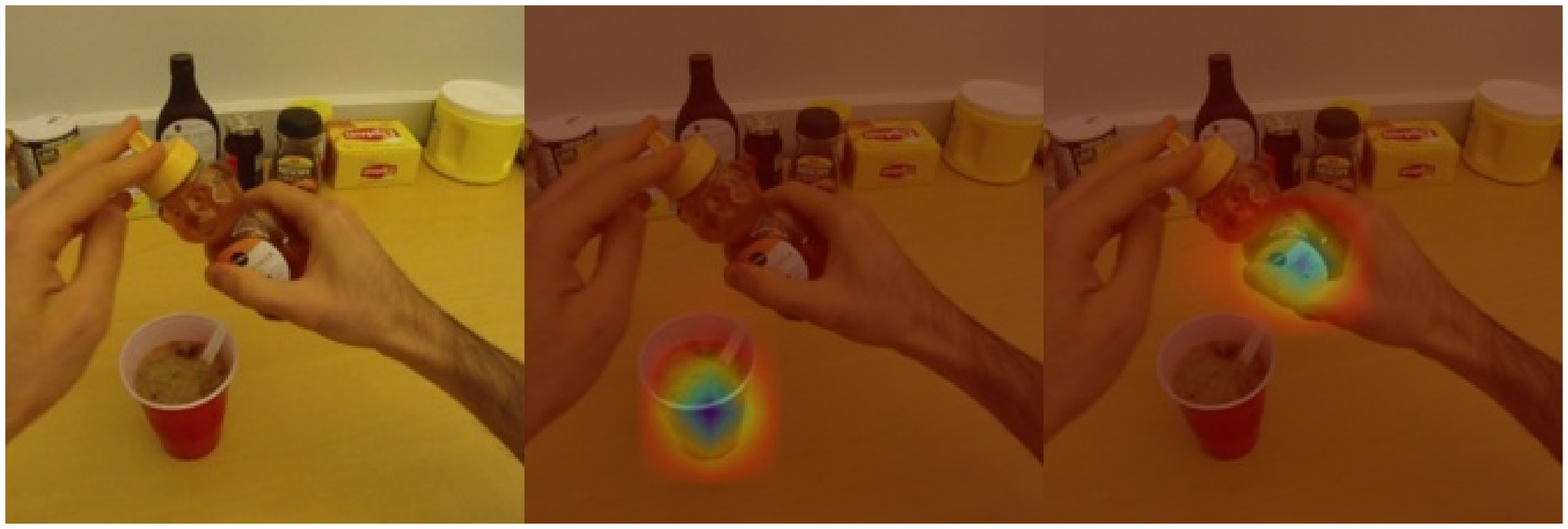}
	\end{subfigure}
	\begin{subfigure}[b]{0.32\textwidth}
		\includegraphics[scale=0.17]{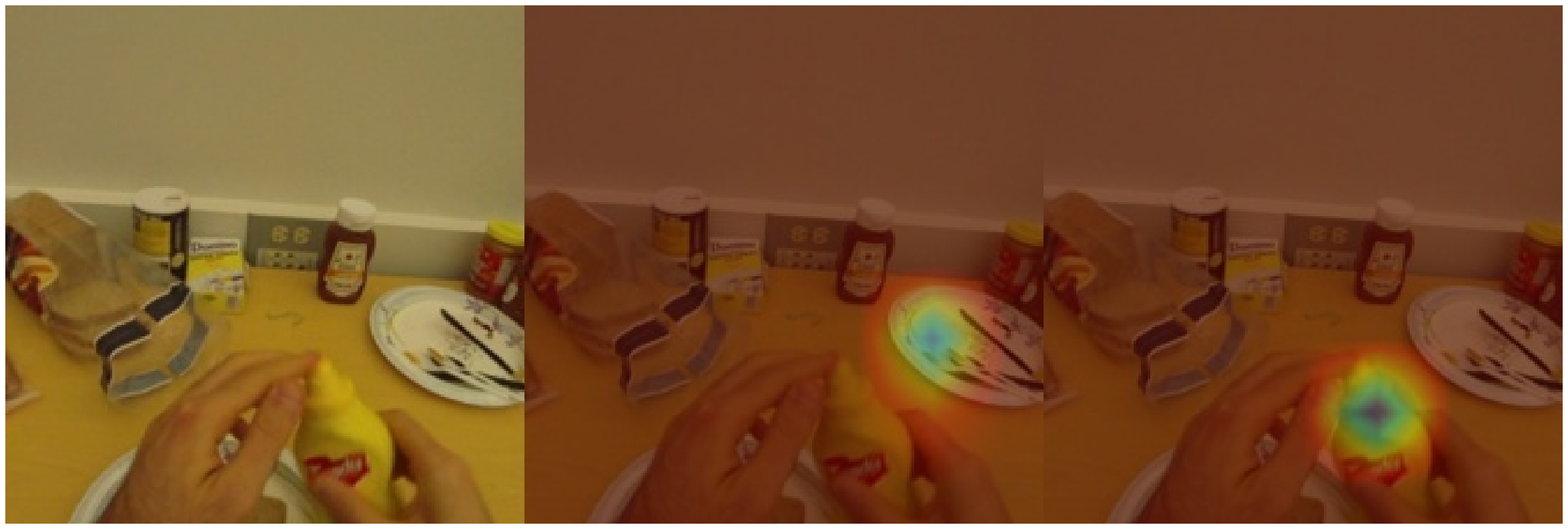}
	\end{subfigure}\\
	\vskip 2mm
	\begin{subfigure}[b]{0.32\textwidth}
		\includegraphics[scale=0.17]{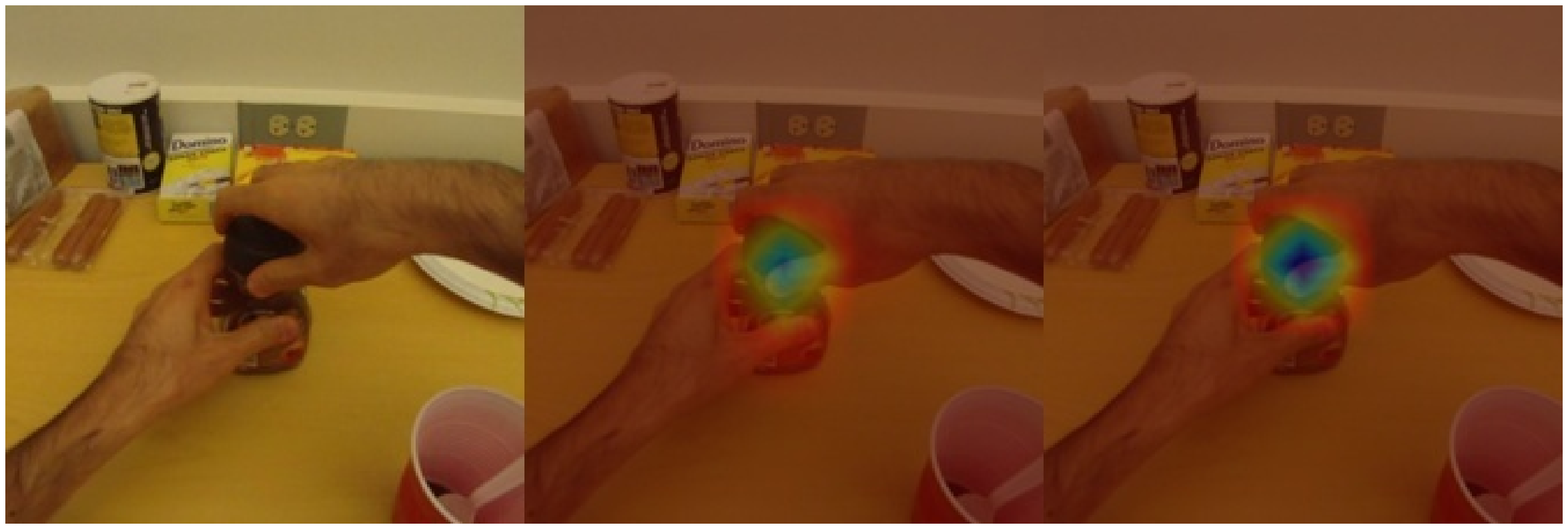}
		\caption{Close coffee}
	\end{subfigure}
	\begin{subfigure}[b]{0.32\textwidth}
		\includegraphics[scale=0.17]{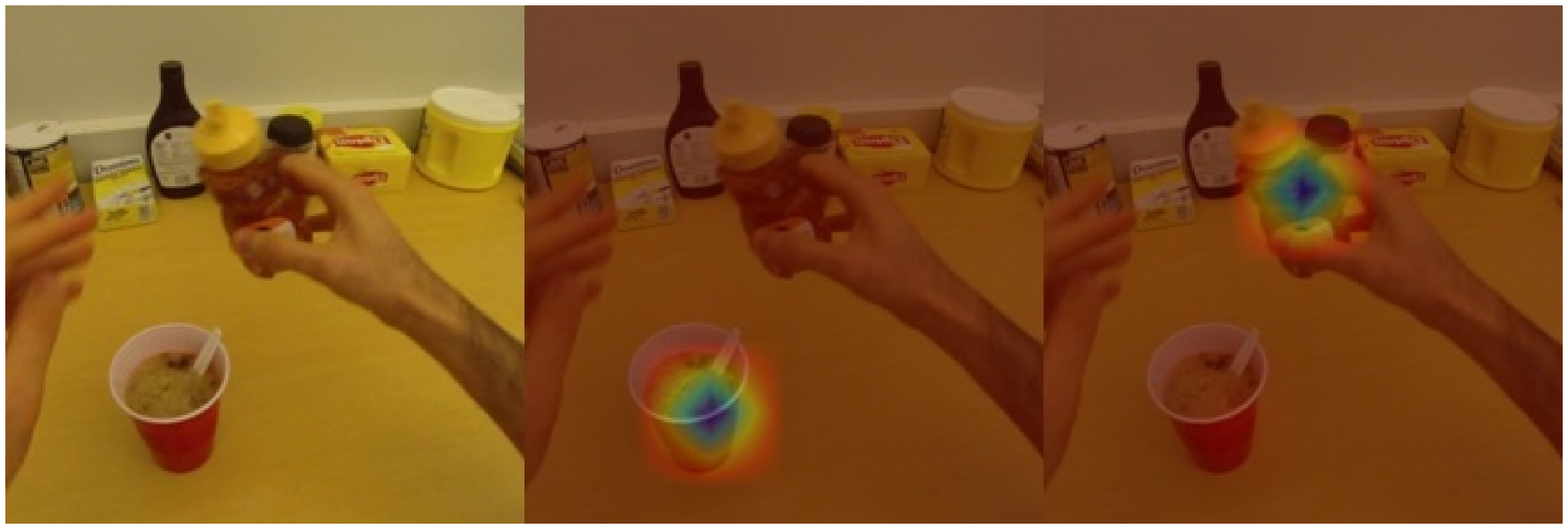}
		\caption{Close honey}
	\end{subfigure}
	\begin{subfigure}[b]{0.32\textwidth}
		\includegraphics[scale=0.17]{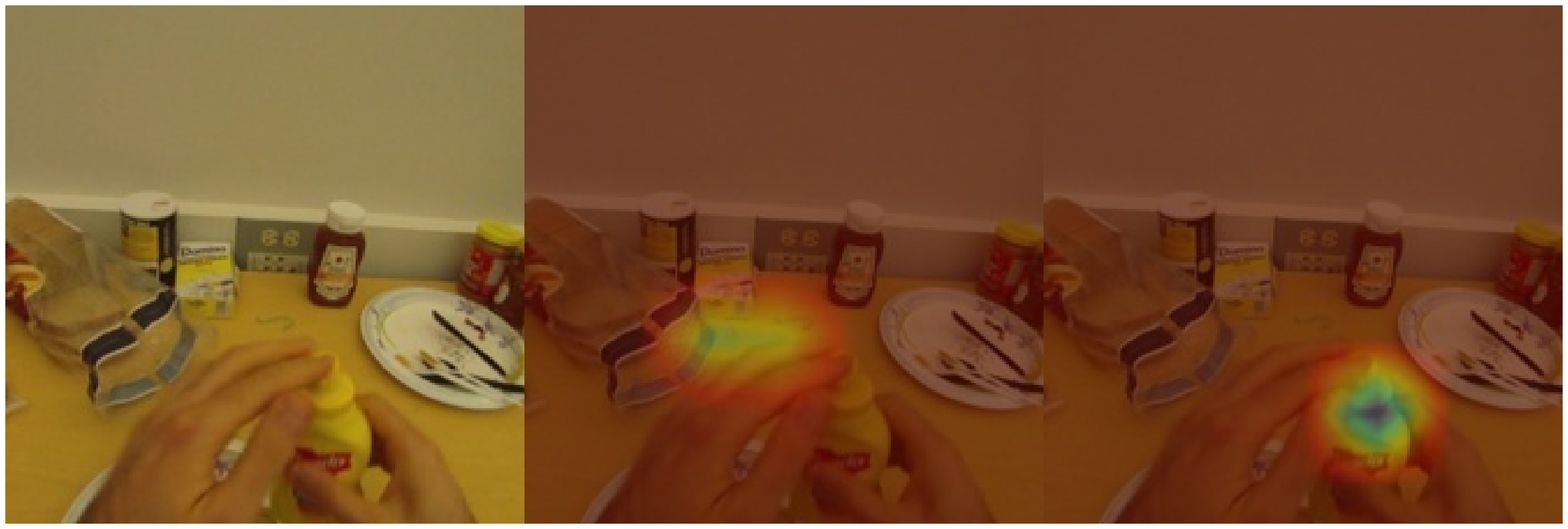}
		\caption{Close mustard}
	\end{subfigure}
	\vspace*{2mm}
	\caption{Spatial attention maps obtained for frames from GTEA(61) dataset. The clip identifiers are: (a) S1\_Coffee\_C1 (b) S1\_CofHoney\_C1 (c) S1\_Hotdog\_C1}
	\label{fig:fig_ex1}
\end{figure}

\begin{figure}[h]
	\centering      
	\begin{subfigure}[b]{0.32\textwidth}
		\includegraphics[scale=0.17]{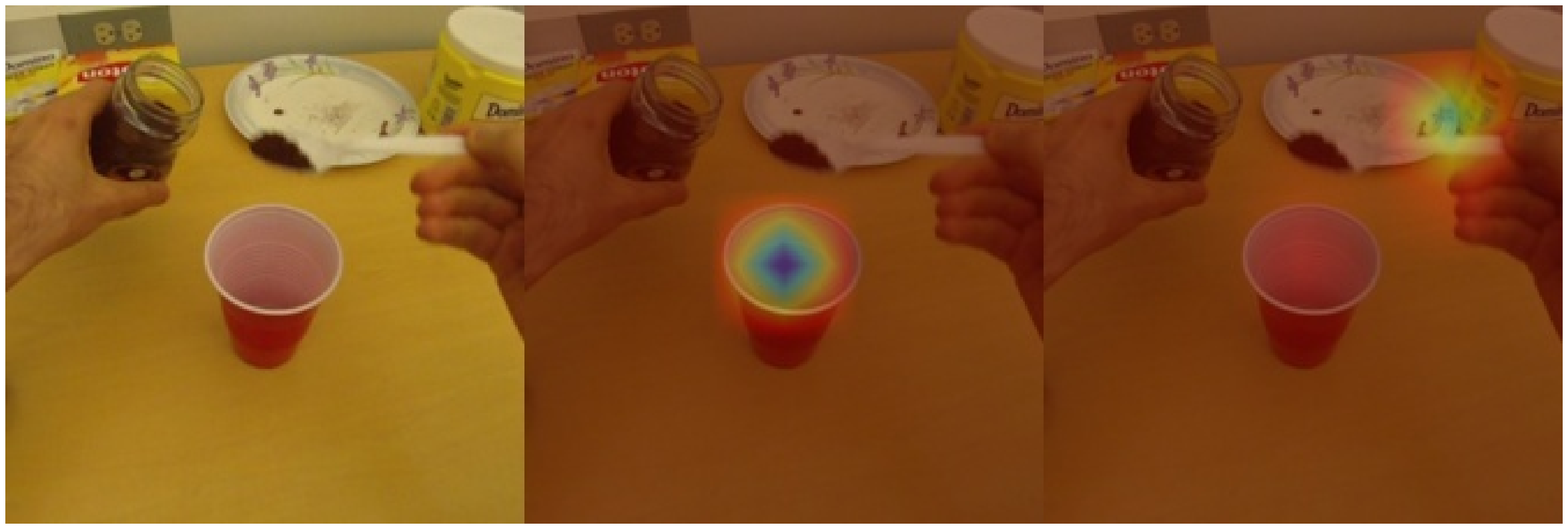}
	\end{subfigure}
	\begin{subfigure}[b]{0.32\textwidth}
		\includegraphics[scale=0.17]{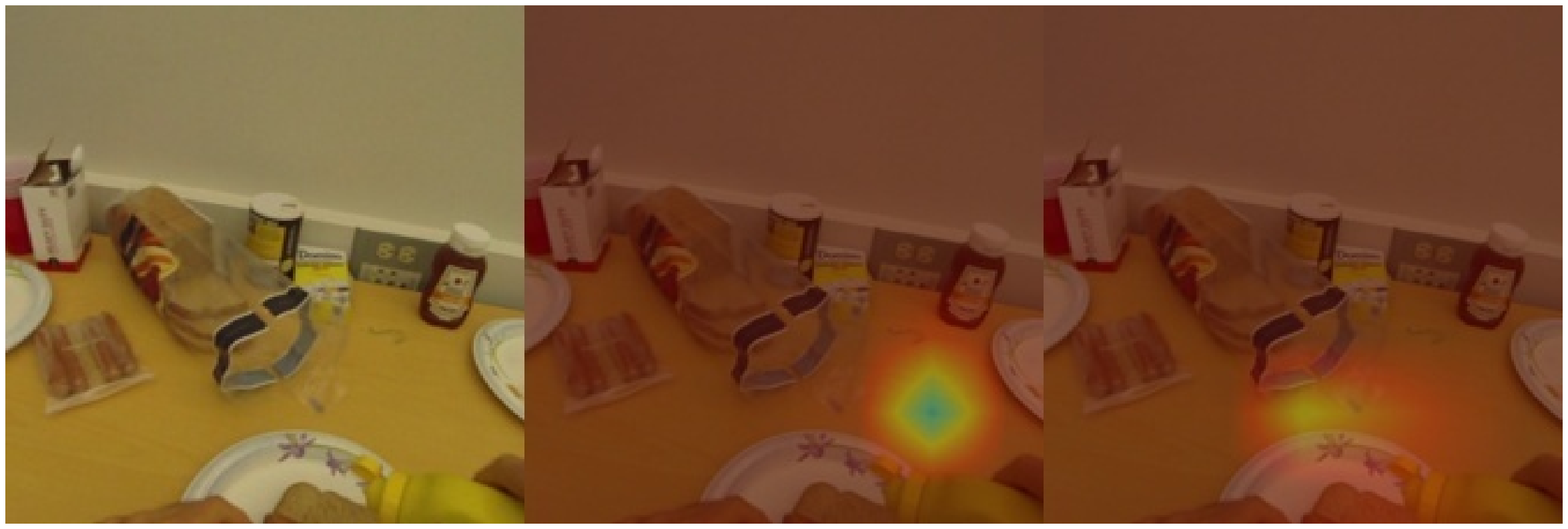}
	\end{subfigure}
	\begin{subfigure}[b]{0.32\textwidth}
		\includegraphics[scale=0.17]{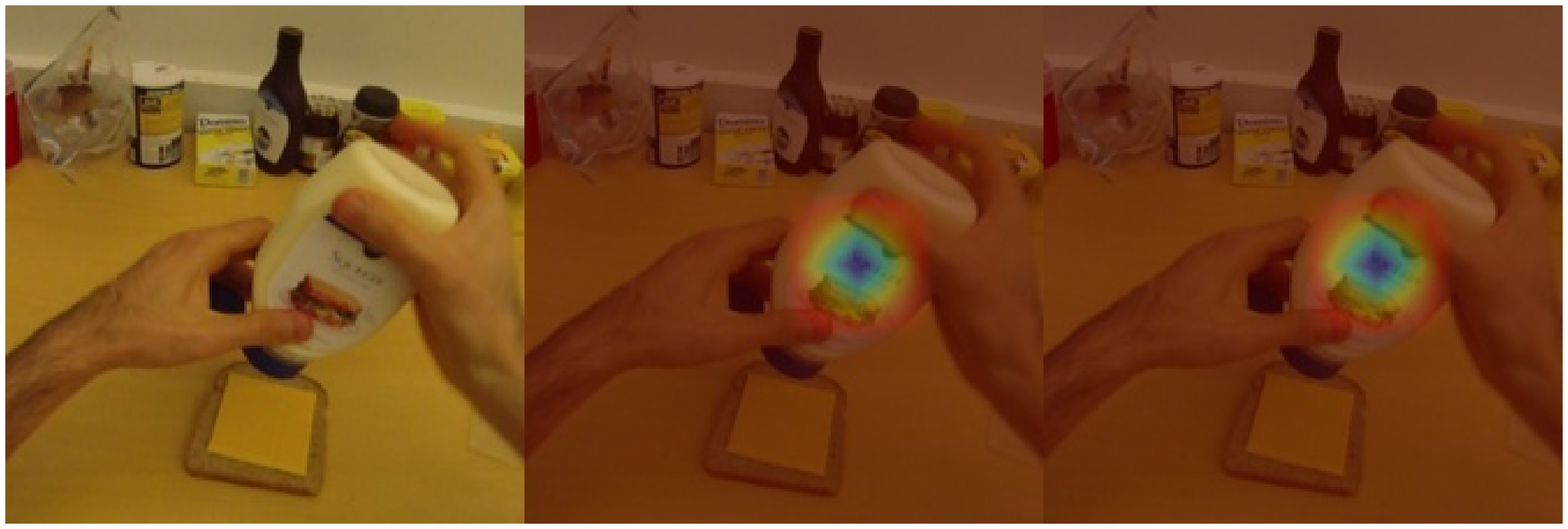}
	\end{subfigure}\\
	\vskip 2mm
	\begin{subfigure}[b]{0.32\textwidth}
		\includegraphics[scale=0.17]{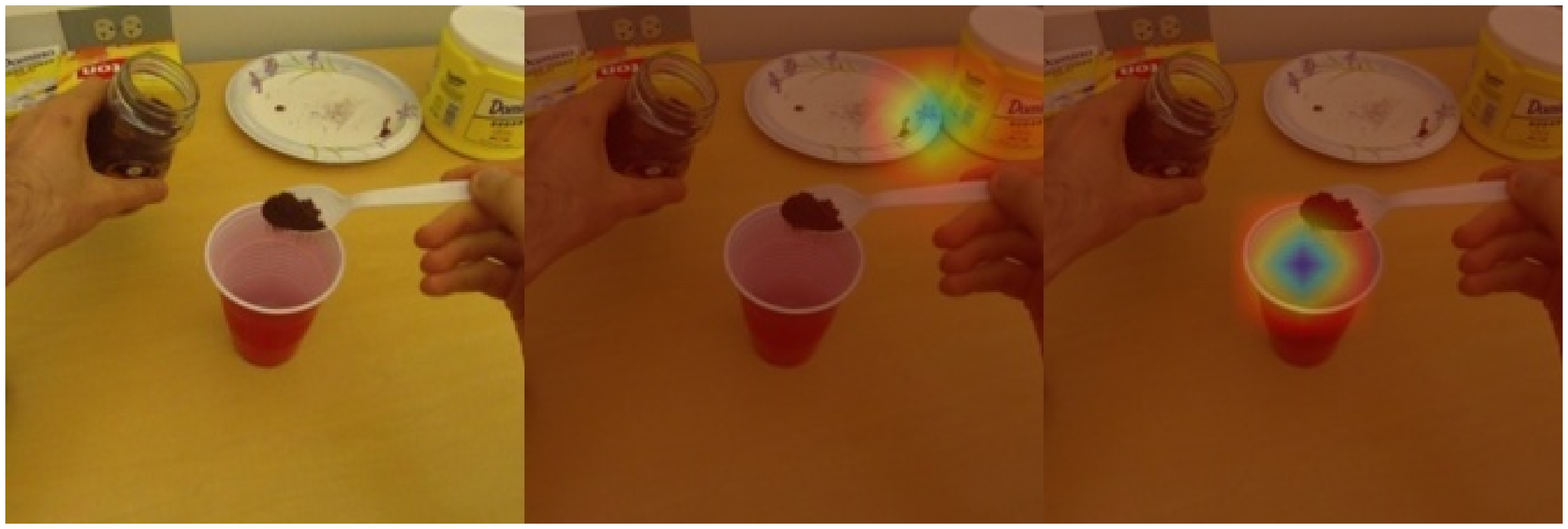}
	\end{subfigure}
	\begin{subfigure}[b]{0.32\textwidth}
		\includegraphics[scale=0.17]{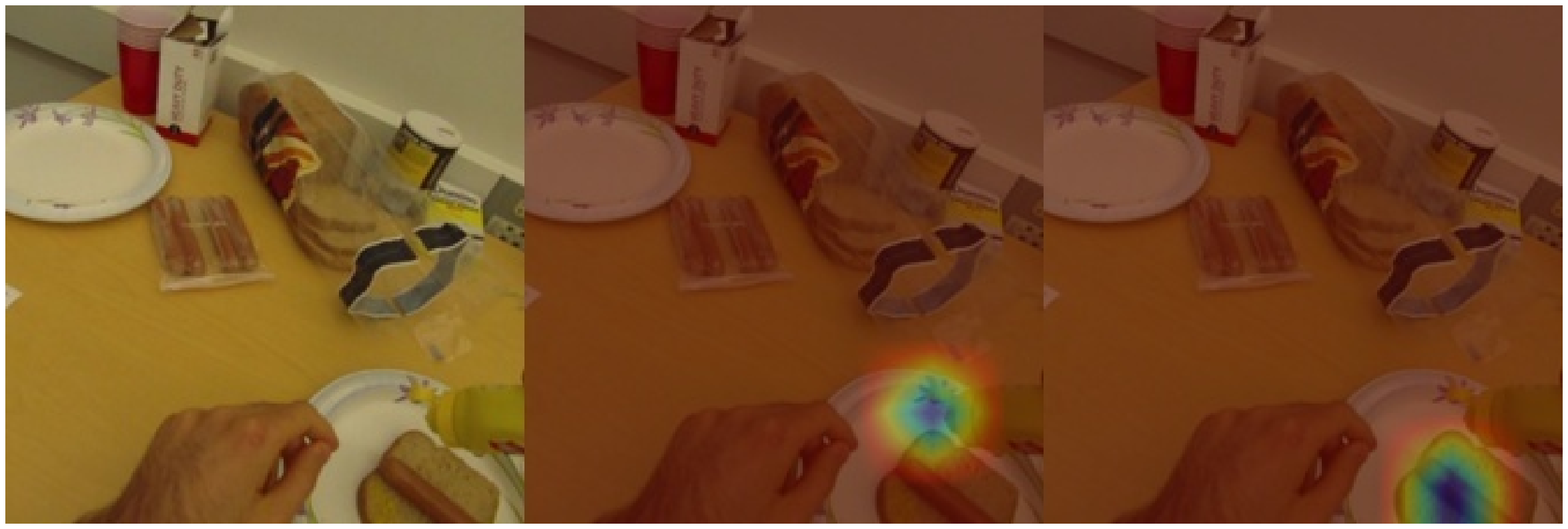}
	\end{subfigure}
	\begin{subfigure}[b]{0.32\textwidth}
		\includegraphics[scale=0.17]{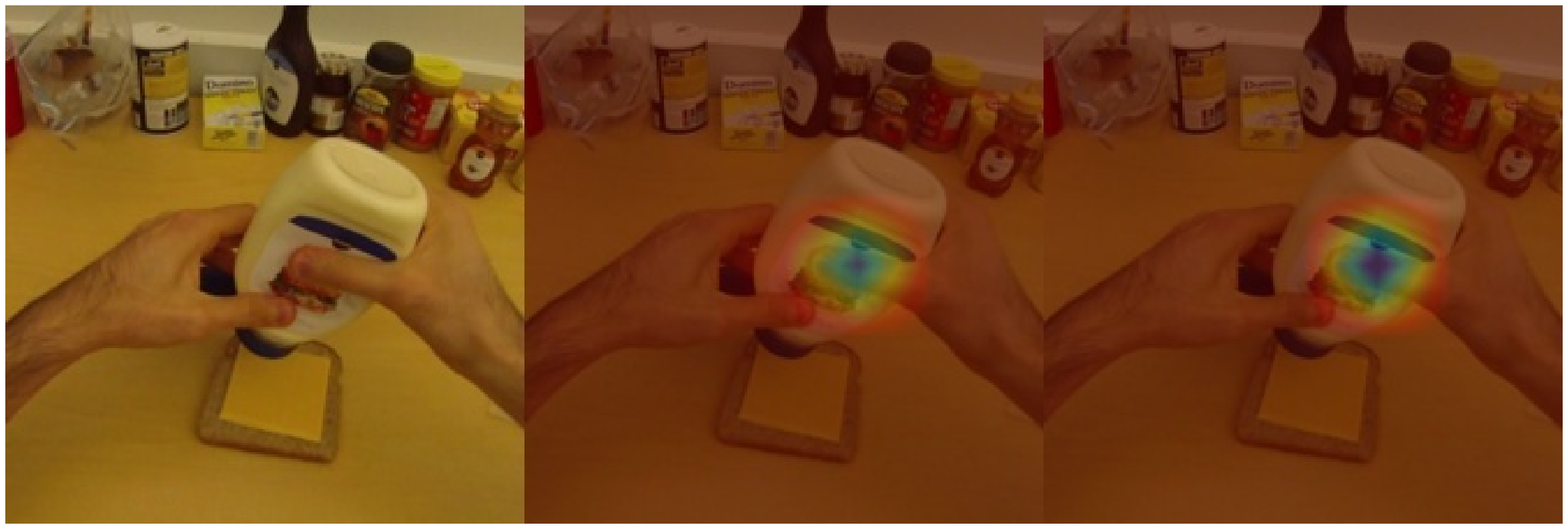}
	\end{subfigure}\\
	\vskip 2mm
	\begin{subfigure}[b]{0.32\textwidth}
		\includegraphics[scale=0.17]{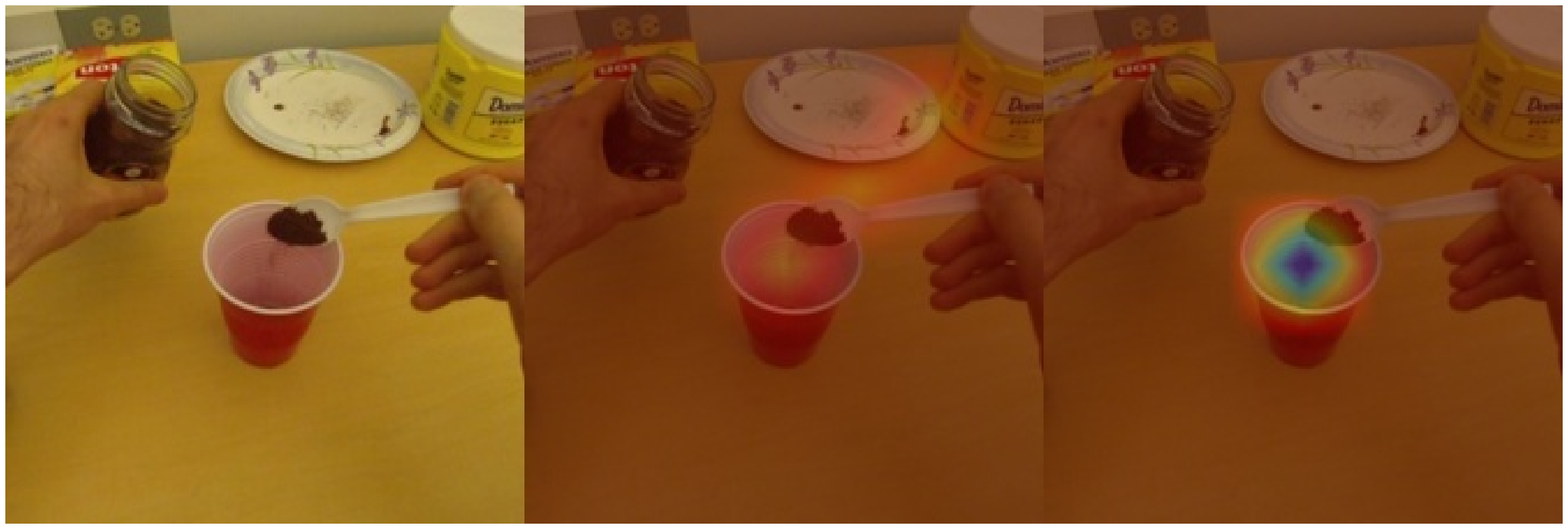}
	\end{subfigure}
	\begin{subfigure}[b]{0.32\textwidth}
		\includegraphics[scale=0.17]{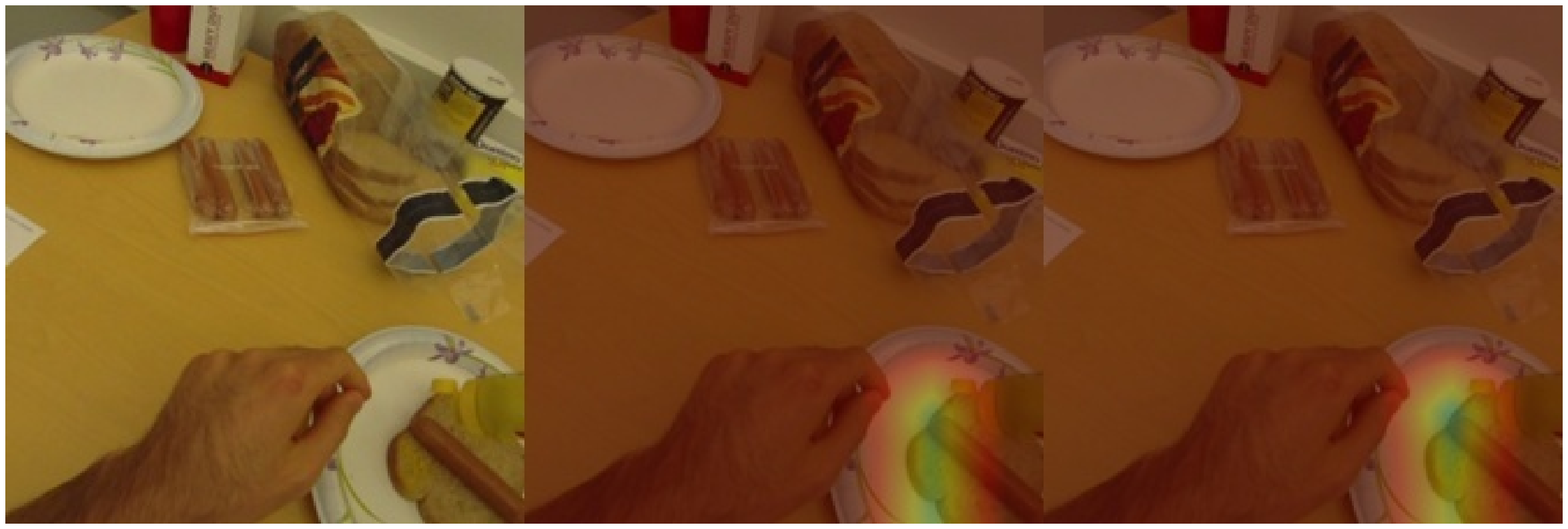}
	\end{subfigure}
	\begin{subfigure}[b]{0.32\textwidth}
		\includegraphics[scale=0.17]{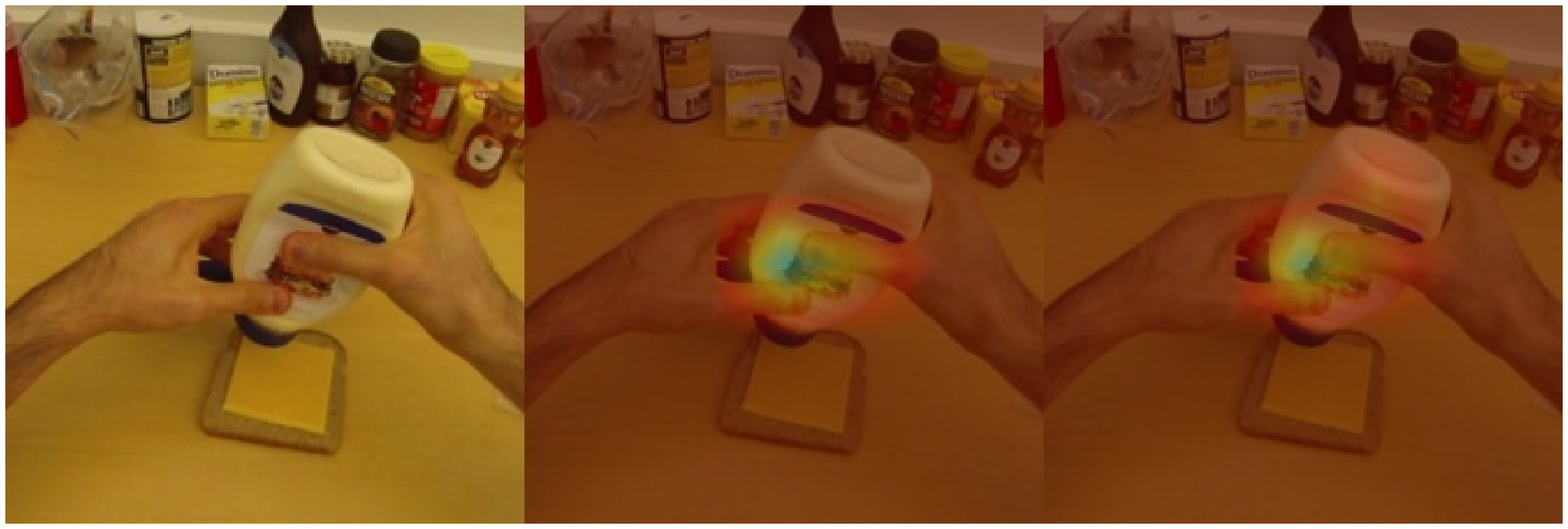}
	\end{subfigure}\\
	\vskip 2mm
	\begin{subfigure}[b]{0.32\textwidth}
		\includegraphics[scale=0.17]{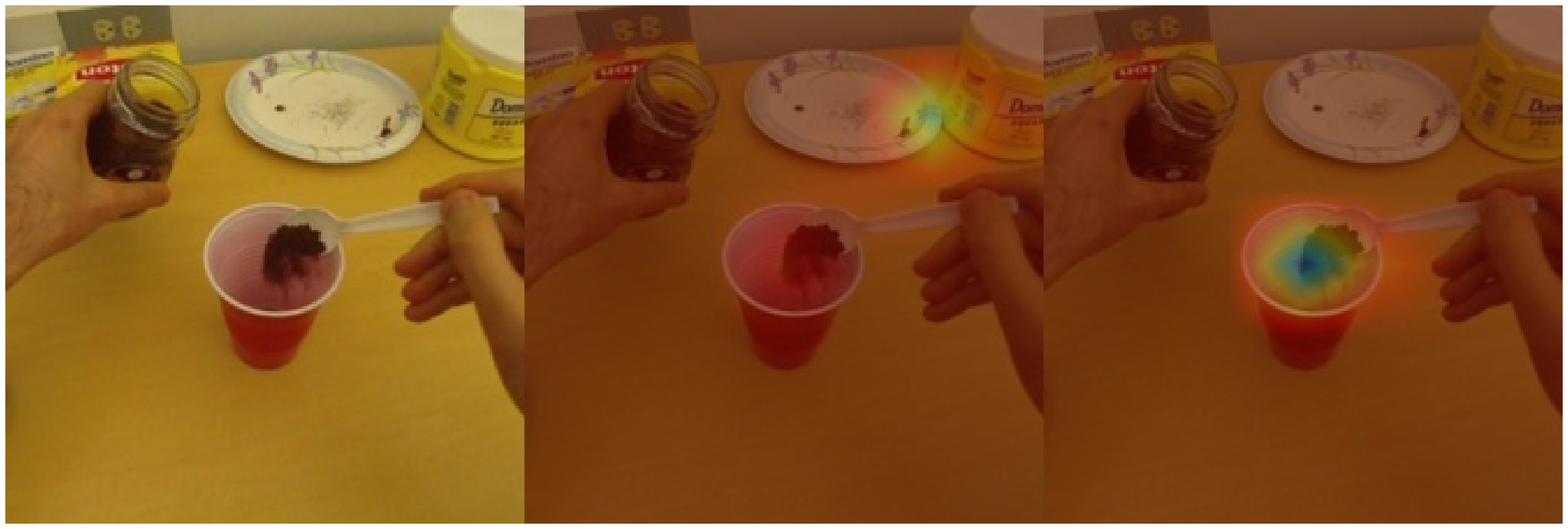}
	\end{subfigure}
	\begin{subfigure}[b]{0.32\textwidth}
		\includegraphics[scale=0.17]{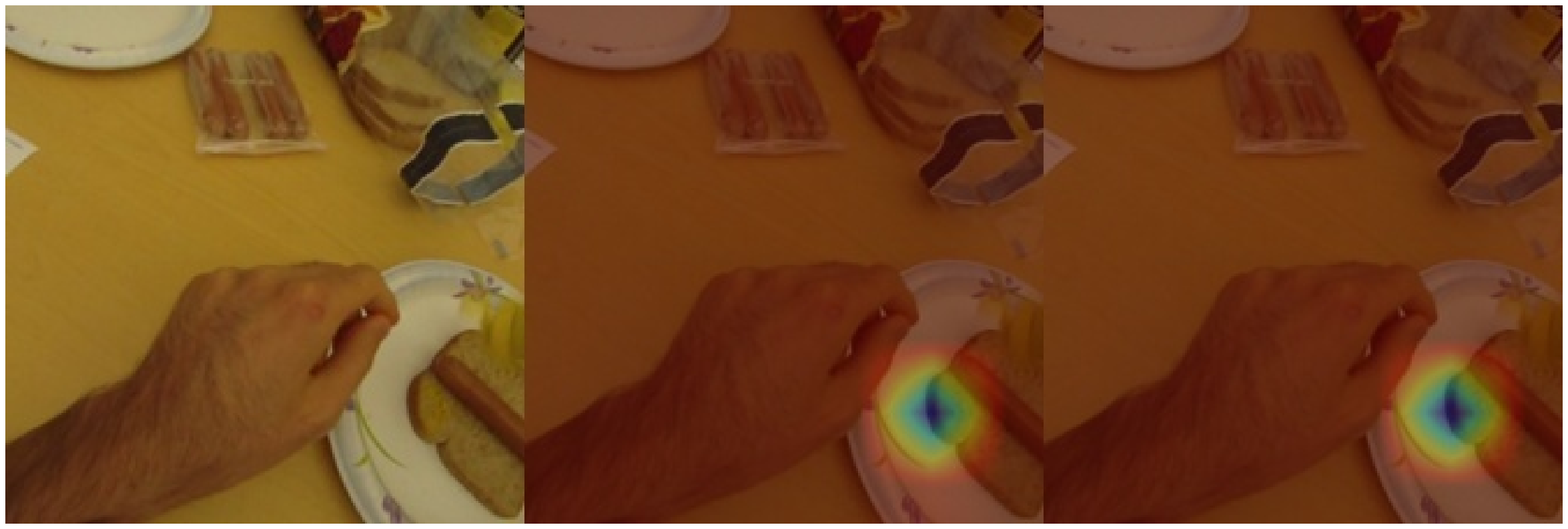}
	\end{subfigure}
	\begin{subfigure}[b]{0.32\textwidth}
		\includegraphics[scale=0.17]{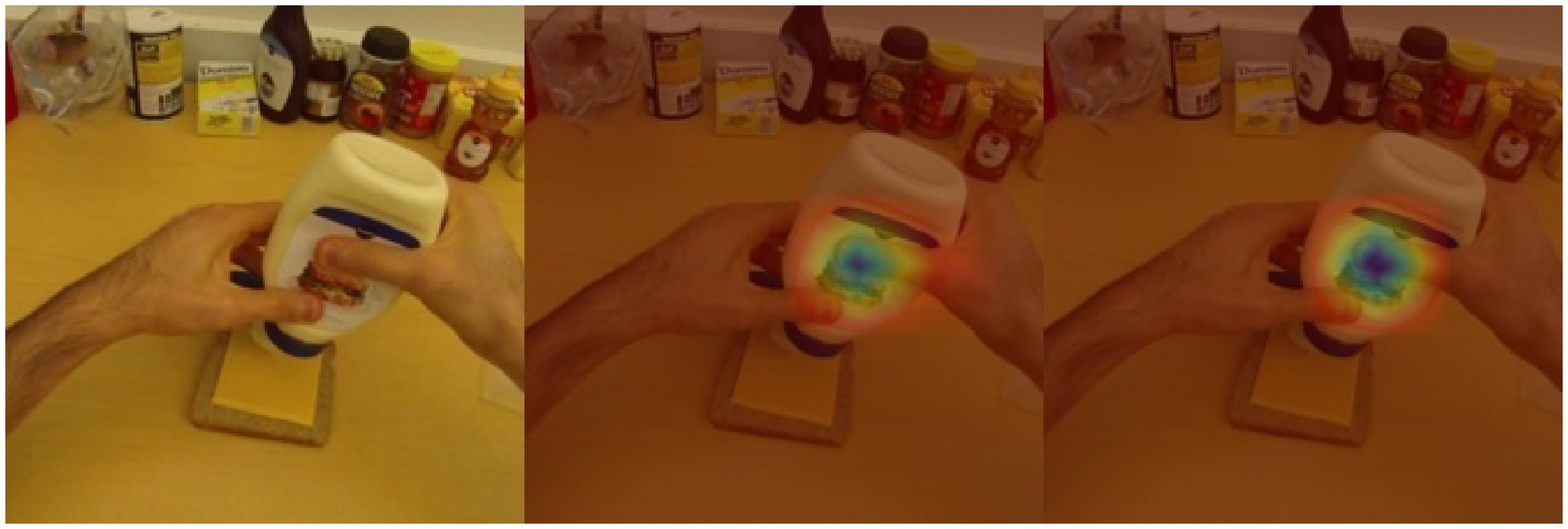}
	\end{subfigure}\\
	\vskip 2mm
	\begin{subfigure}[b]{0.32\textwidth}
		\includegraphics[scale=0.17]{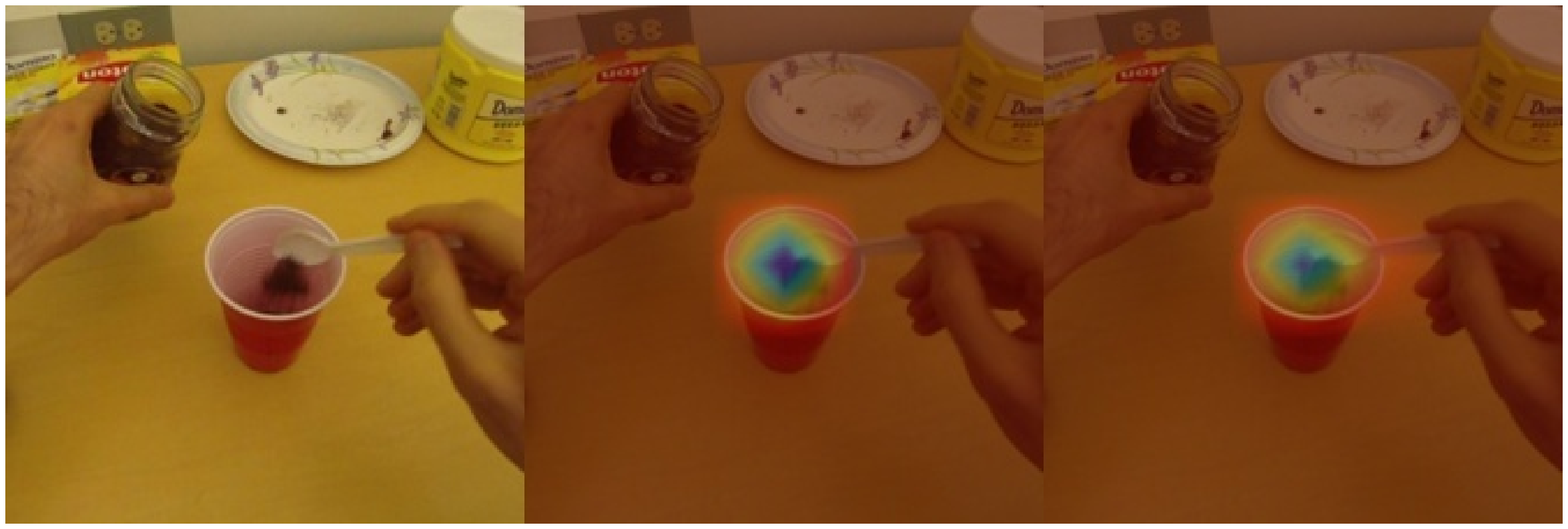}
	\end{subfigure}
	\begin{subfigure}[b]{0.32\textwidth}
		\includegraphics[scale=0.17]{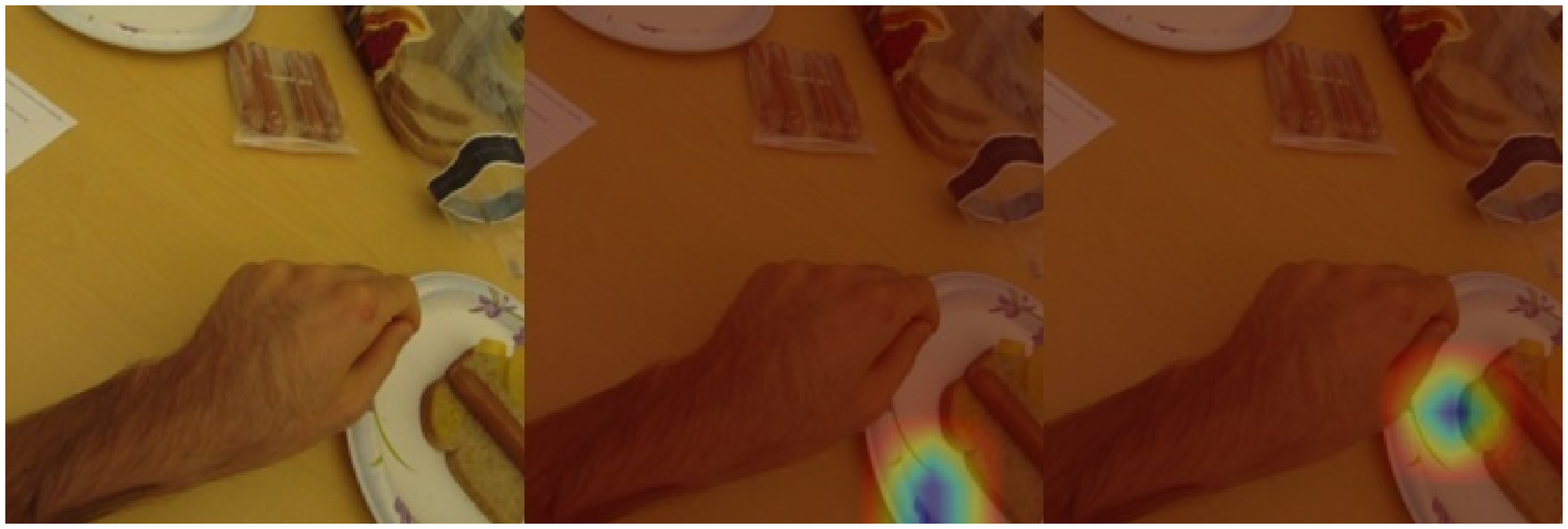}
	\end{subfigure}
	\begin{subfigure}[b]{0.32\textwidth}
		\includegraphics[scale=0.17]{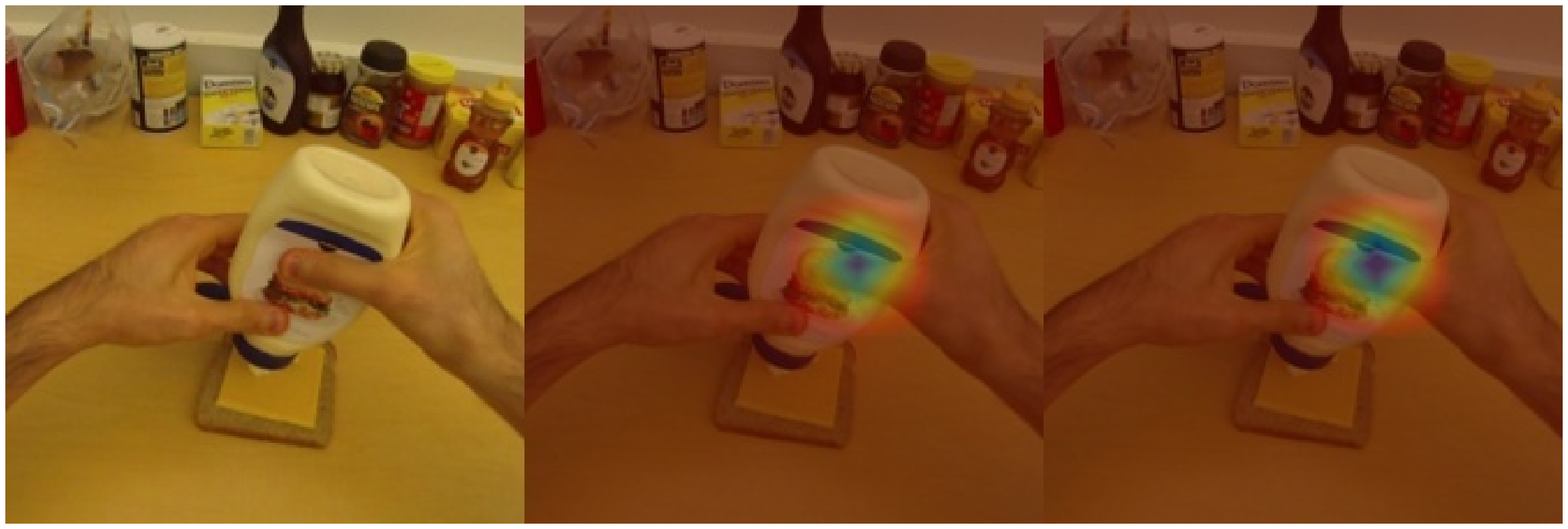}
	\end{subfigure}\\
	\vskip 2mm
	\begin{subfigure}[b]{0.32\textwidth}
		\includegraphics[scale=0.17]{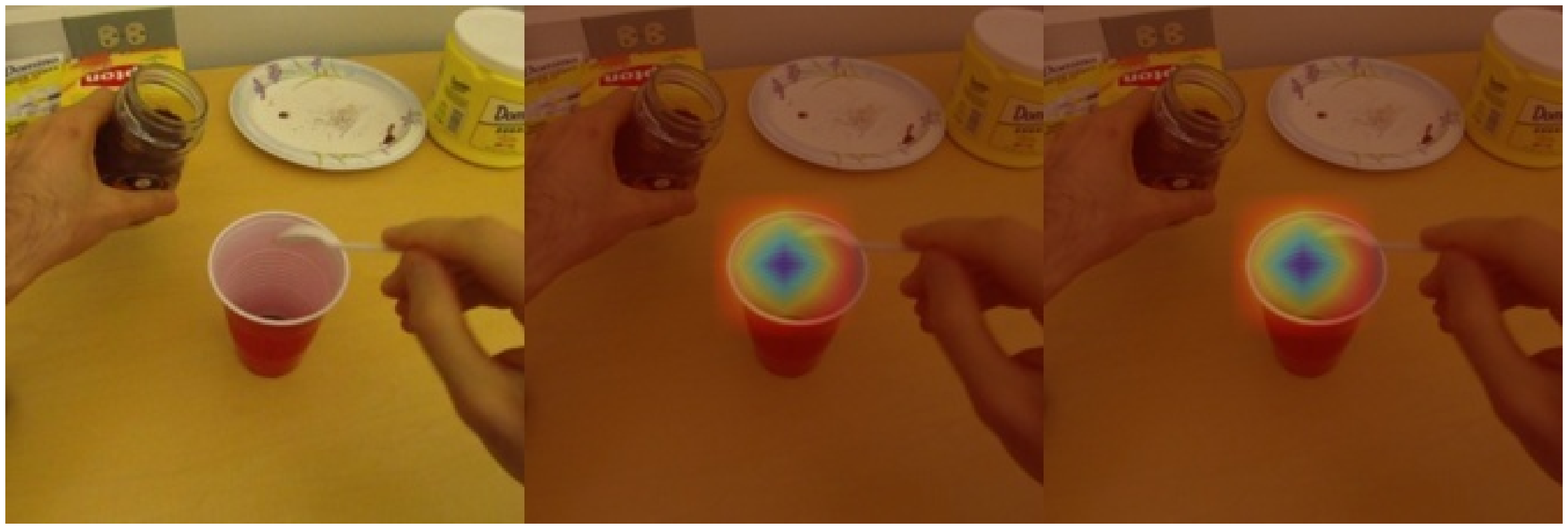}
	\end{subfigure}
	\begin{subfigure}[b]{0.32\textwidth}
		\includegraphics[scale=0.17]{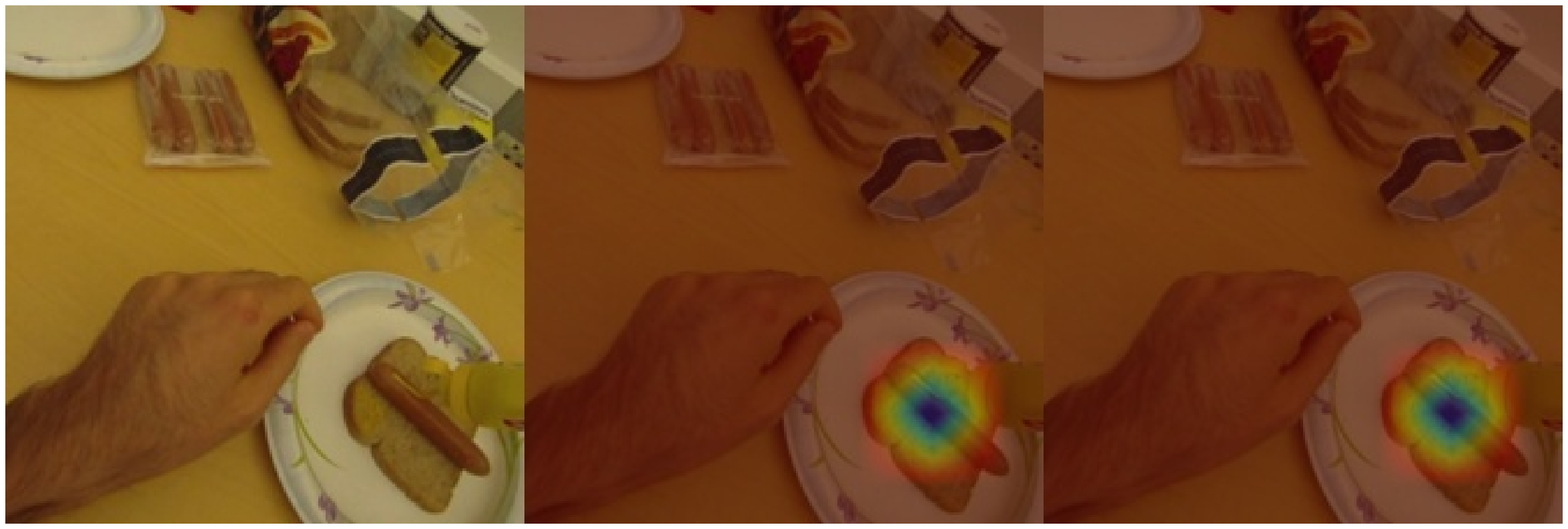}
	\end{subfigure}
	\begin{subfigure}[b]{0.32\textwidth}
		\includegraphics[scale=0.17]{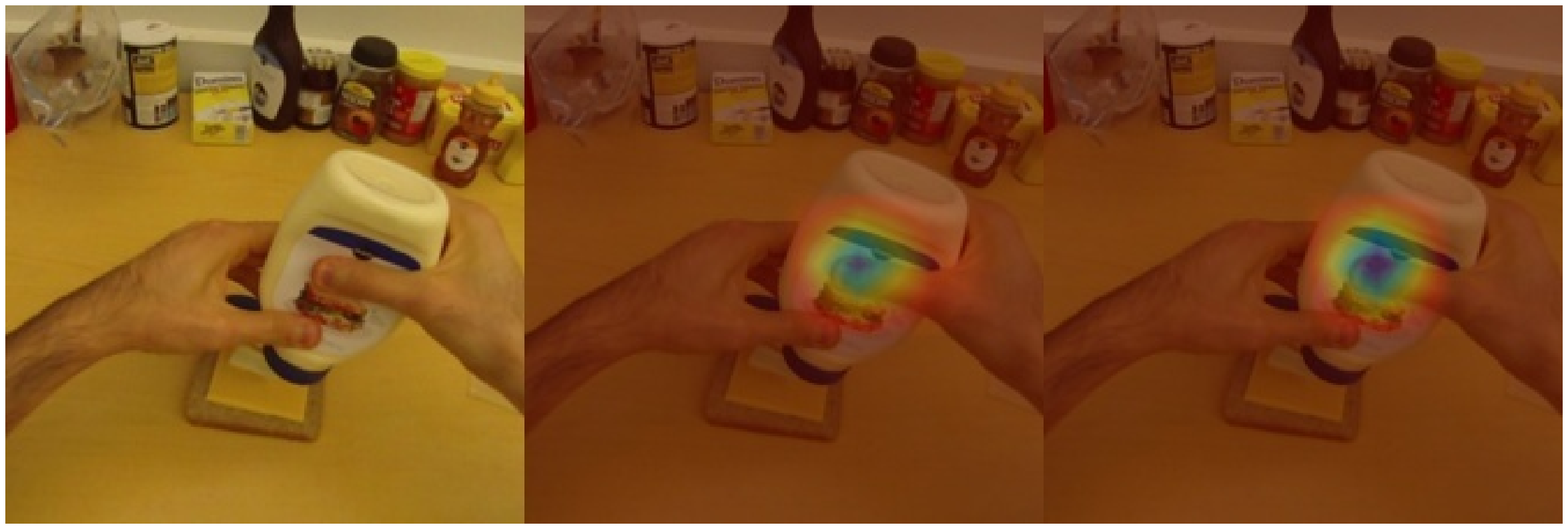}
	\end{subfigure}\\
	\vskip 2mm
	\begin{subfigure}[b]{0.32\textwidth}
		\includegraphics[scale=0.17]{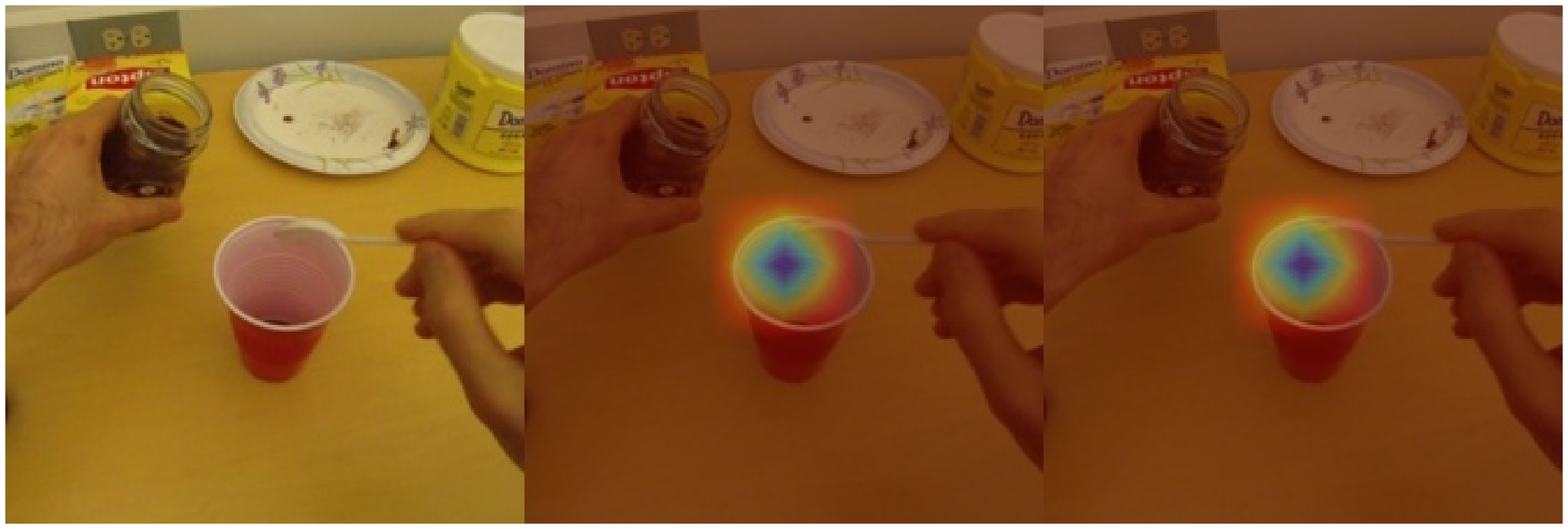}
	\end{subfigure}
	\begin{subfigure}[b]{0.32\textwidth}
		\includegraphics[scale=0.17]{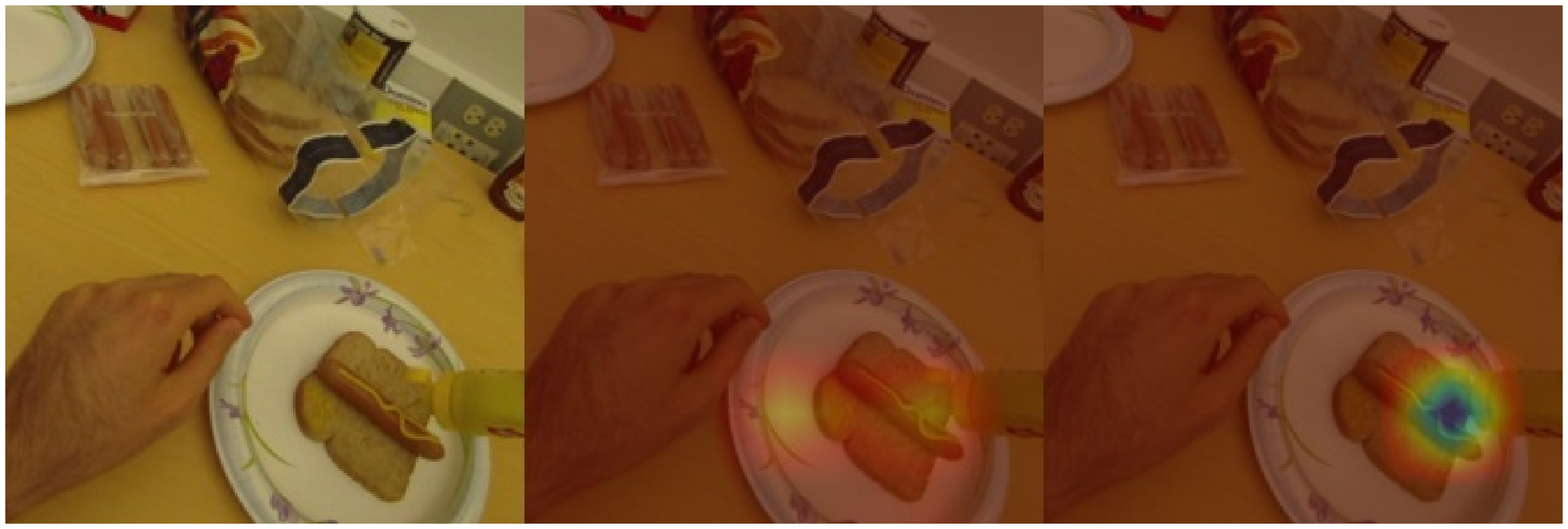}
	\end{subfigure}
	\begin{subfigure}[b]{0.32\textwidth}
		\includegraphics[scale=0.17]{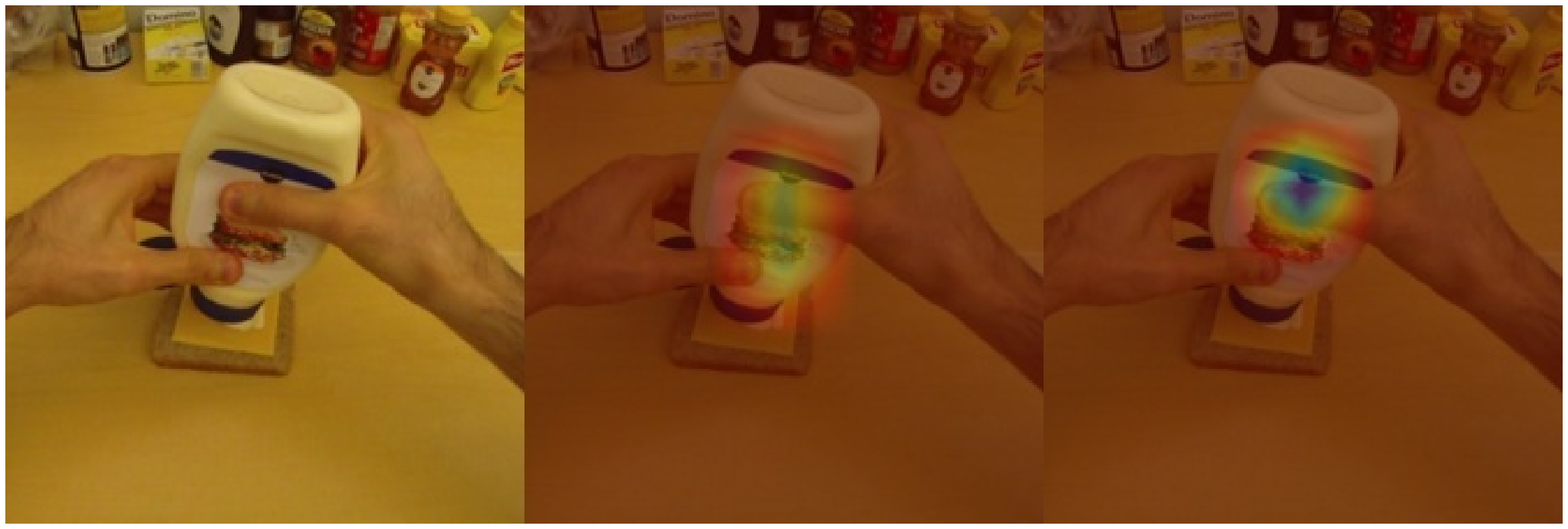}
	\end{subfigure}\\
	\vskip 2mm
	\begin{subfigure}[b]{0.32\textwidth}
		\includegraphics[scale=0.17]{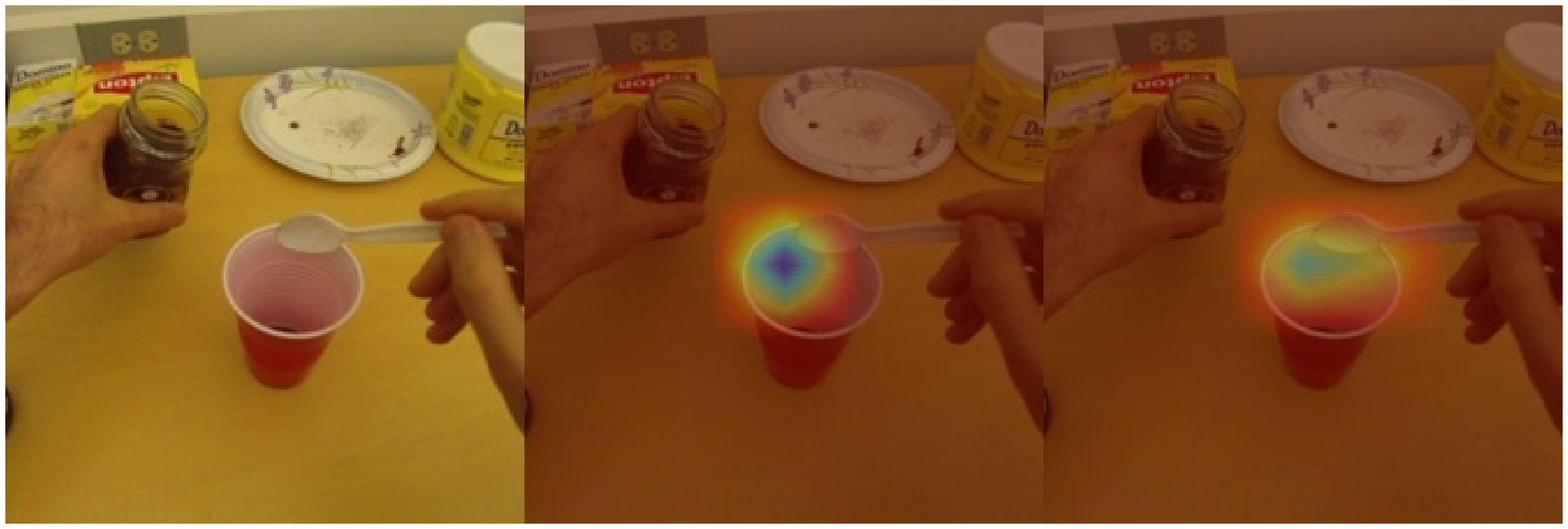}
		\caption{Pour coffee}
	\end{subfigure}
	\begin{subfigure}[b]{0.32\textwidth}
		\includegraphics[scale=0.17]{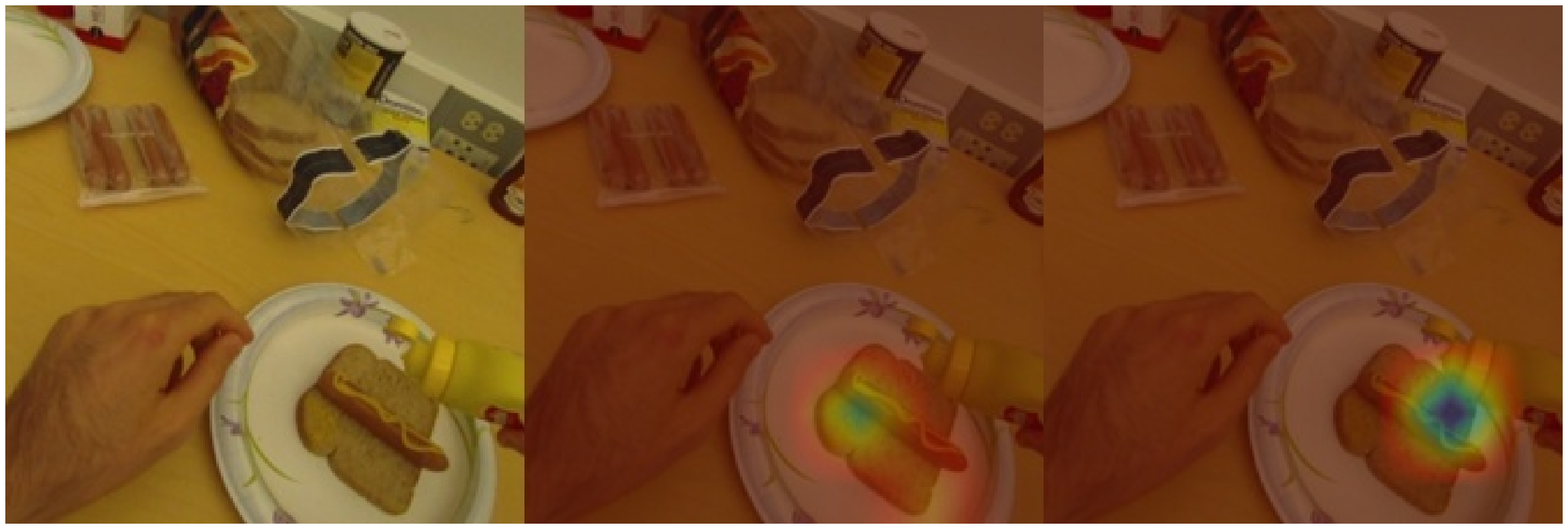}
		\caption{Pour mustard}
	\end{subfigure}
	\begin{subfigure}[b]{0.32\textwidth}
		\includegraphics[scale=0.17]{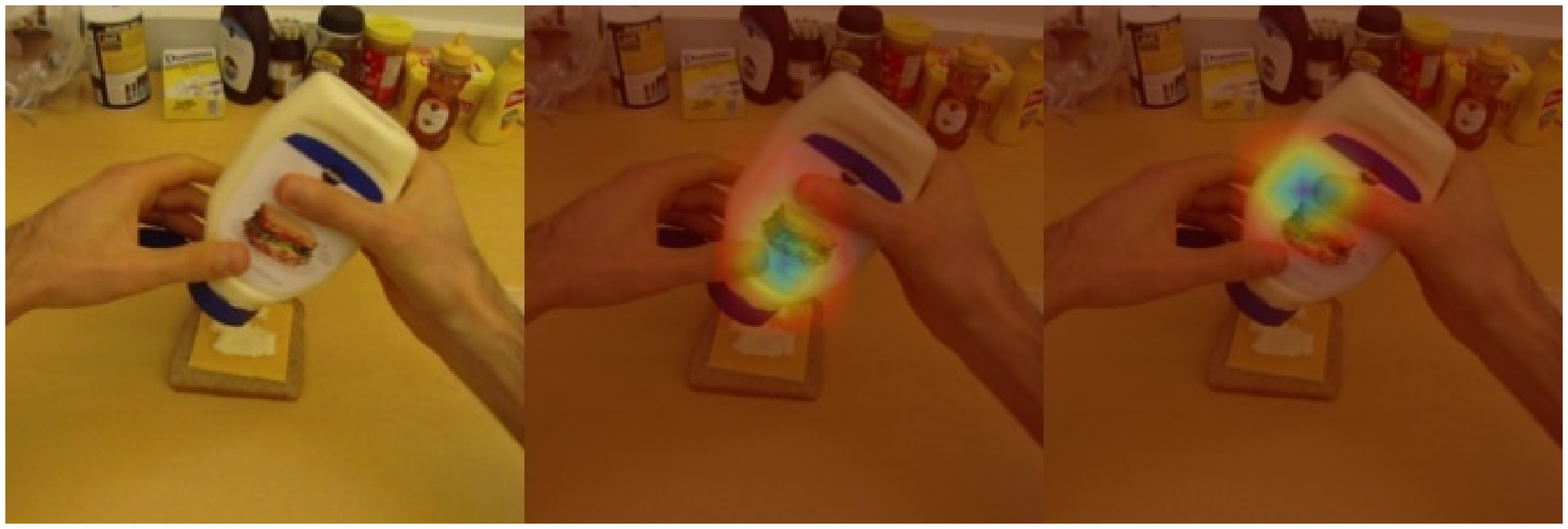}
		\caption{Pour mayonnaise}
	\end{subfigure}
			\vspace*{2mm}
	\caption{Spatial attention maps obtained for frames from GTEA(61) dataset. The clip identifiers are: (a) S1\_Coffee\_C1 (b) S1\_Hotdog\_C1 (c) S1\_Cheese\_C1}
	\label{fig:fig_ex2}
\end{figure}

\begin{figure}[h]
	\centering      
	\begin{subfigure}[b]{0.32\textwidth}
		\includegraphics[scale=0.17]{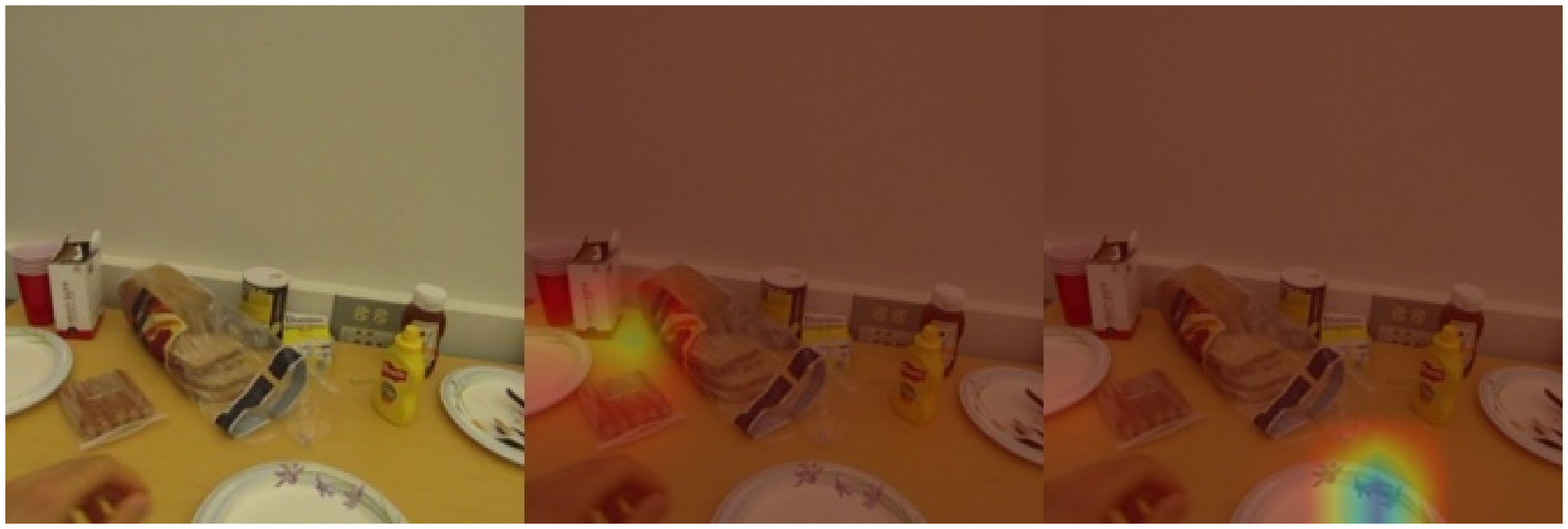}
	\end{subfigure}
	\begin{subfigure}[b]{0.32\textwidth}
		\includegraphics[scale=0.17]{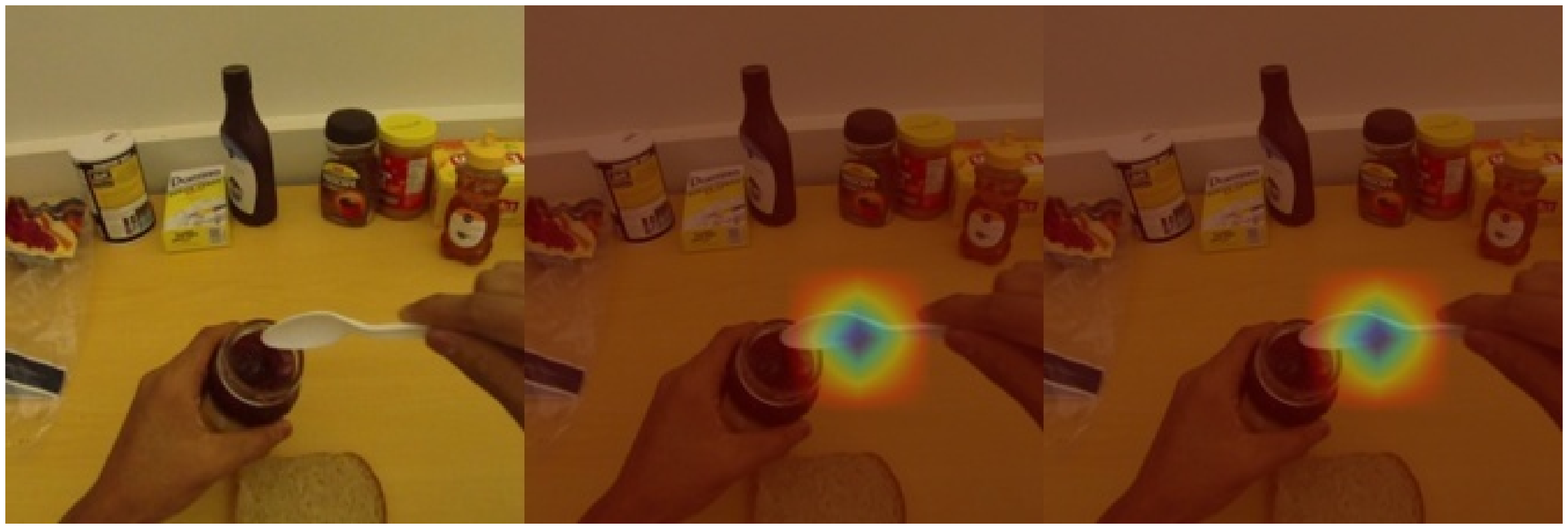}
	\end{subfigure}
	\begin{subfigure}[b]{0.32\textwidth}
		\includegraphics[scale=0.17]{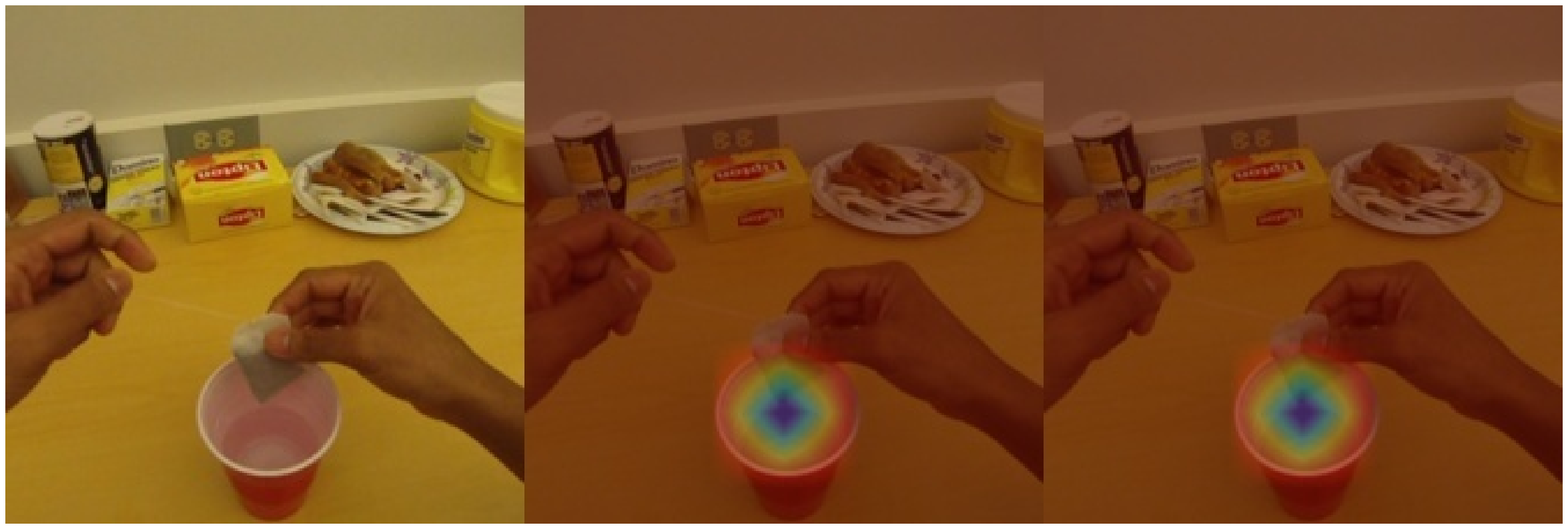}
	\end{subfigure}\\
	\vskip 2mm
	\begin{subfigure}[b]{0.32\textwidth}
		\includegraphics[scale=0.17]{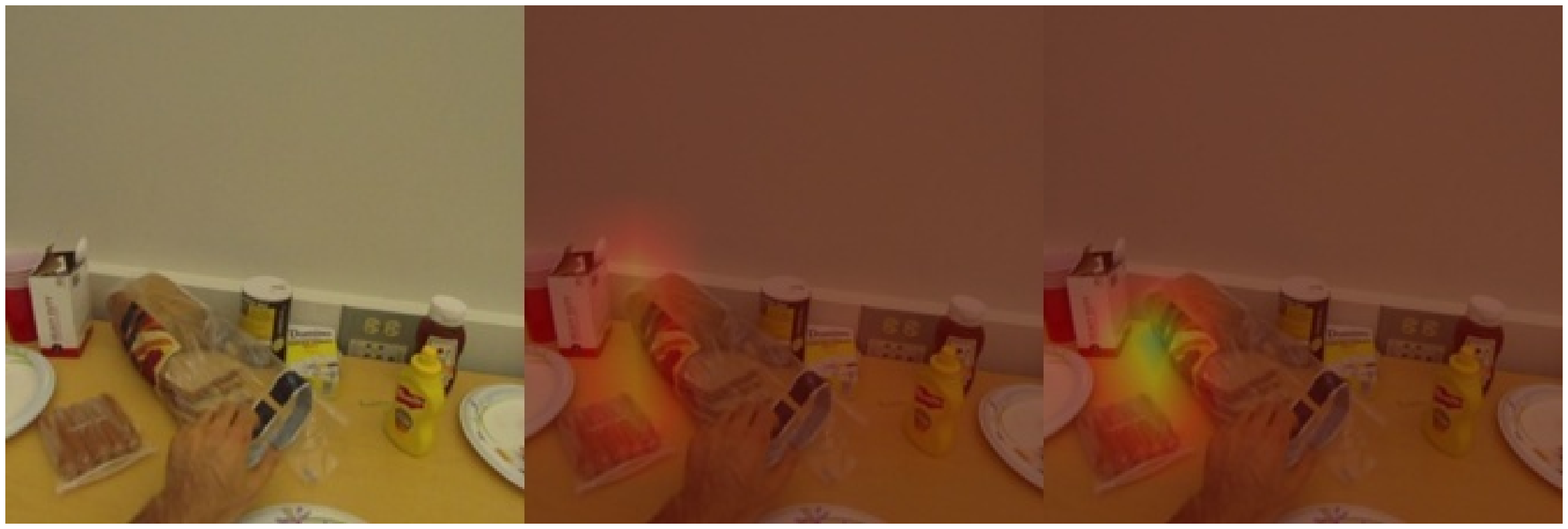}
	\end{subfigure}
	\begin{subfigure}[b]{0.32\textwidth}
		\includegraphics[scale=0.17]{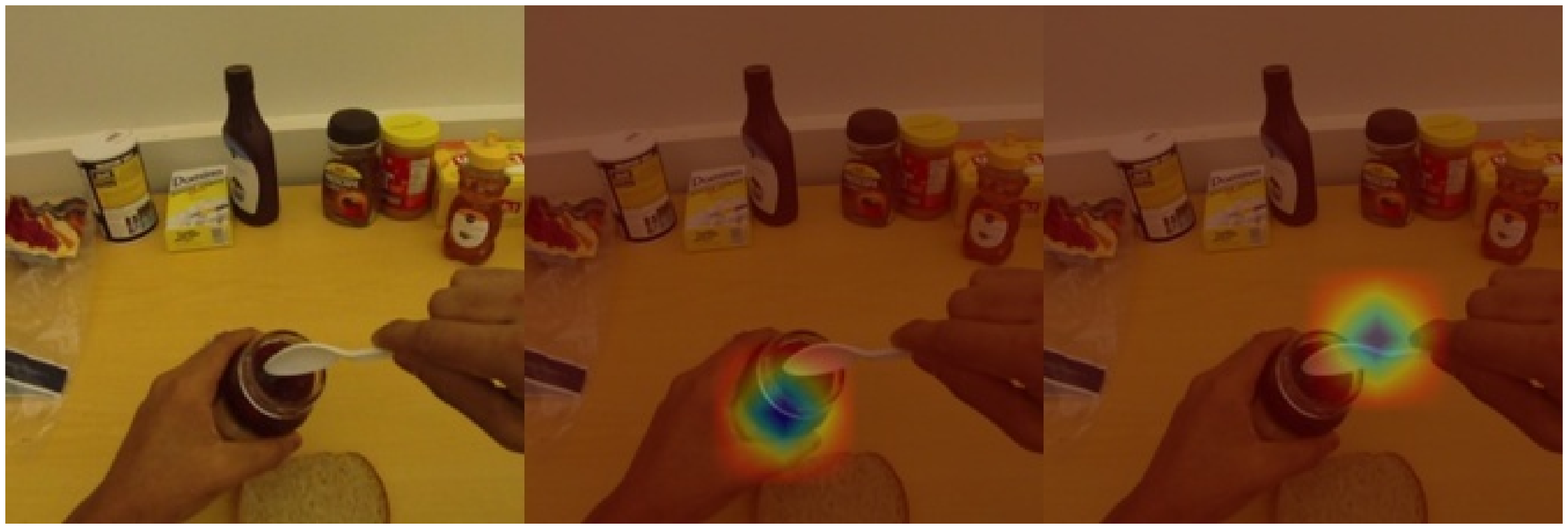}
	\end{subfigure}
	\begin{subfigure}[b]{0.32\textwidth}
		\includegraphics[scale=0.17]{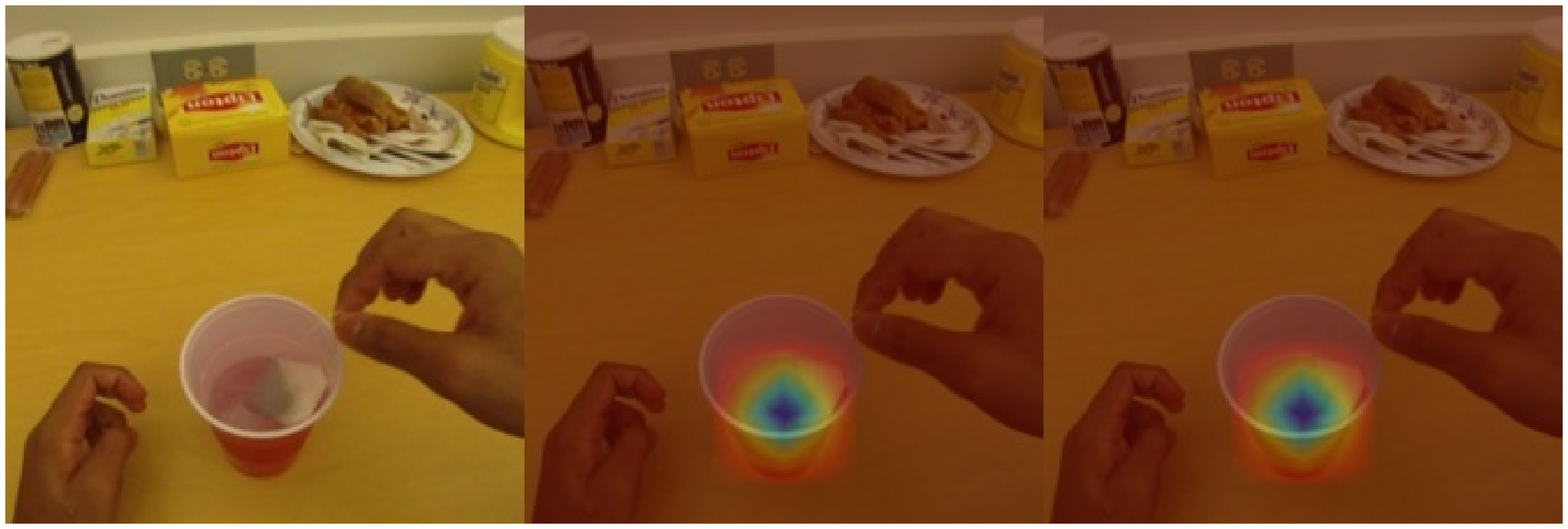}
	\end{subfigure}\\
	\vskip 2mm
	\begin{subfigure}[b]{0.32\textwidth}
		\includegraphics[scale=0.17]{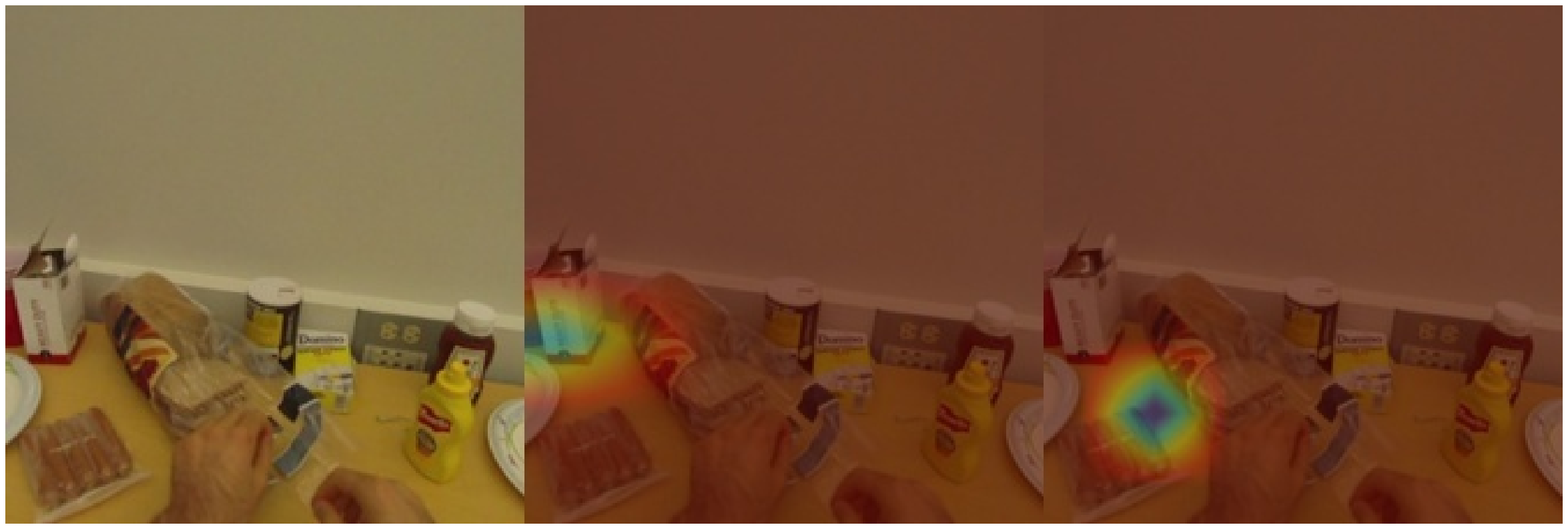}
	\end{subfigure}
	\begin{subfigure}[b]{0.32\textwidth}
		\includegraphics[scale=0.17]{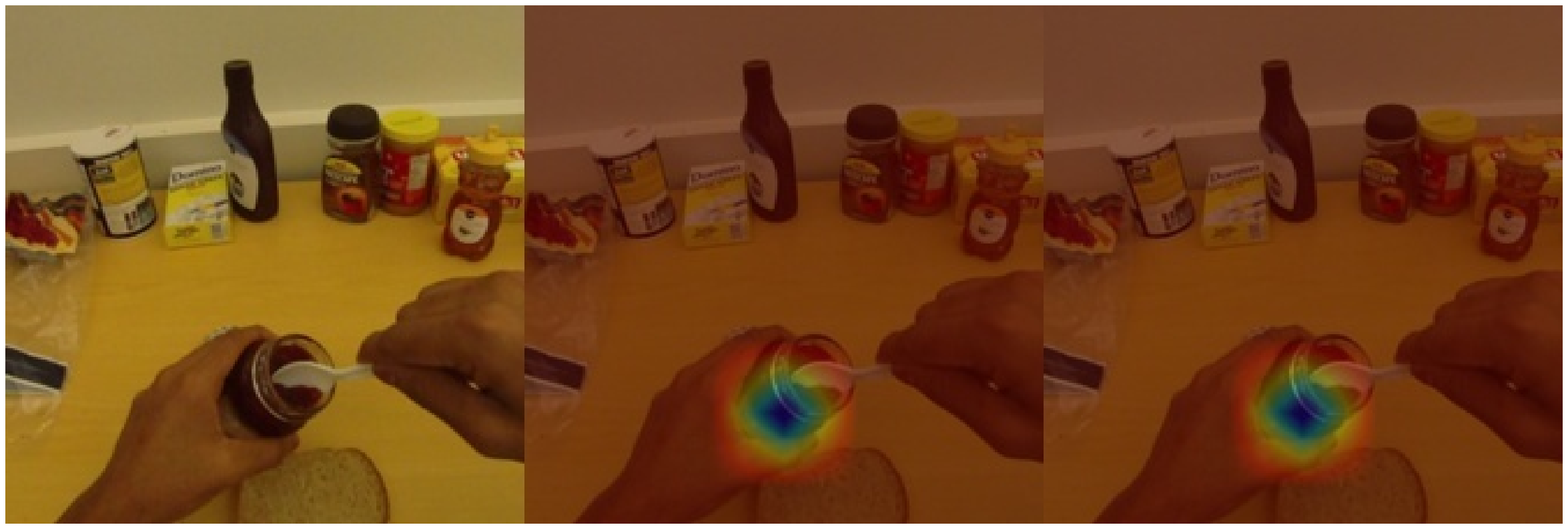}
	\end{subfigure}
	\begin{subfigure}[b]{0.32\textwidth}
		\includegraphics[scale=0.17]{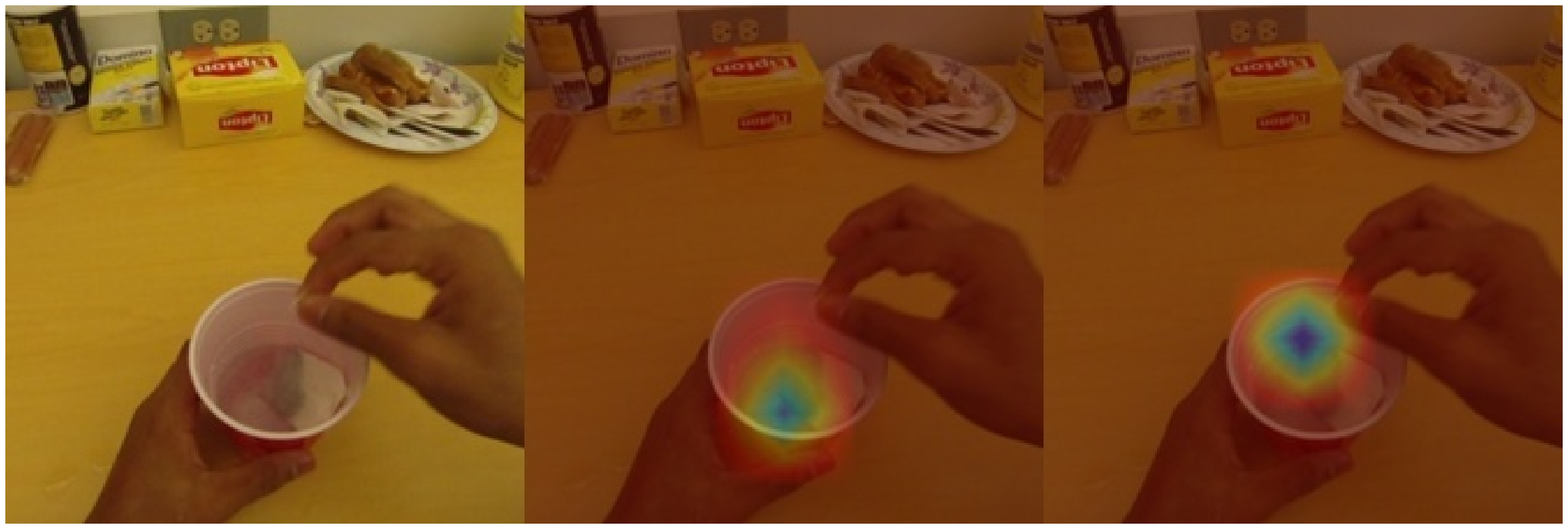}
	\end{subfigure}\\
	\vskip 2mm
	\begin{subfigure}[b]{0.32\textwidth}
		\includegraphics[scale=0.17]{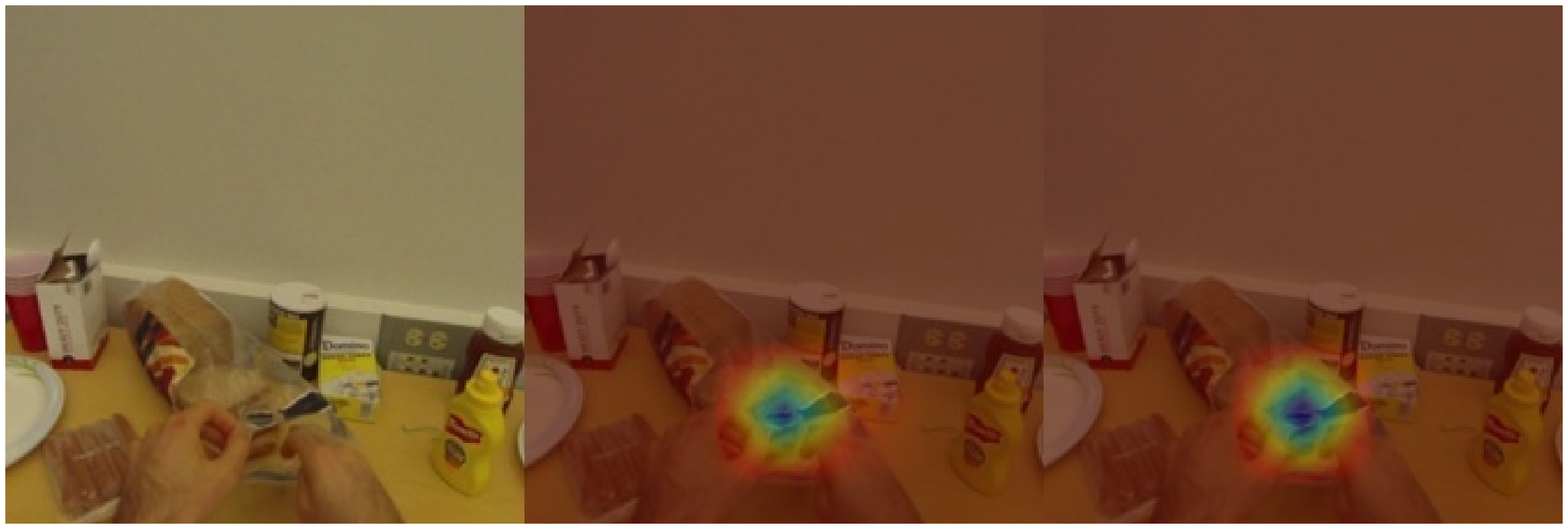}
	\end{subfigure}
	\begin{subfigure}[b]{0.32\textwidth}
		\includegraphics[scale=0.17]{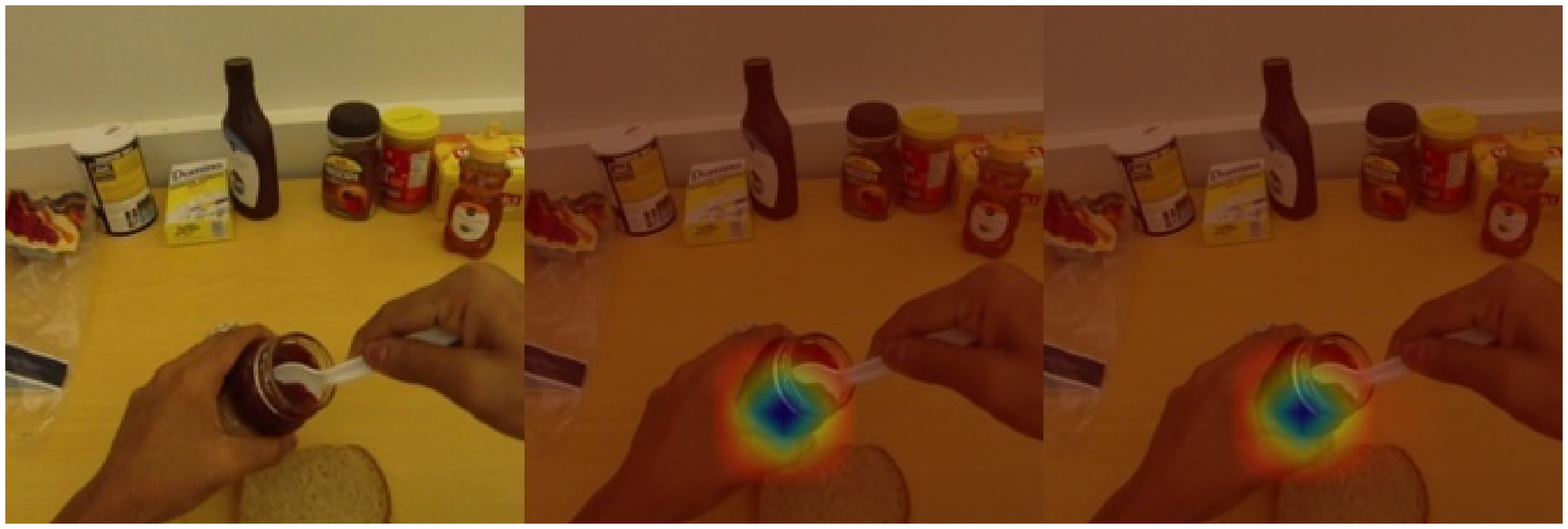}
	\end{subfigure}
	\begin{subfigure}[b]{0.32\textwidth}
		\includegraphics[scale=0.17]{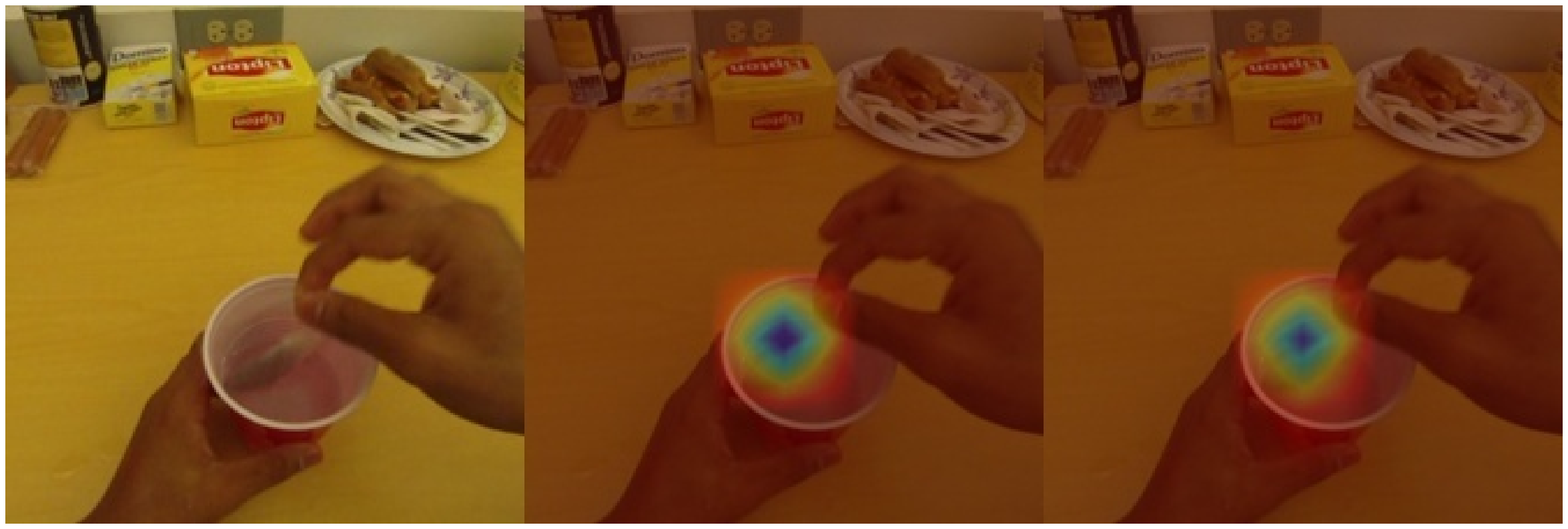}
	\end{subfigure}\\
	\vskip 2mm
	\begin{subfigure}[b]{0.32\textwidth}
		\includegraphics[scale=0.17]{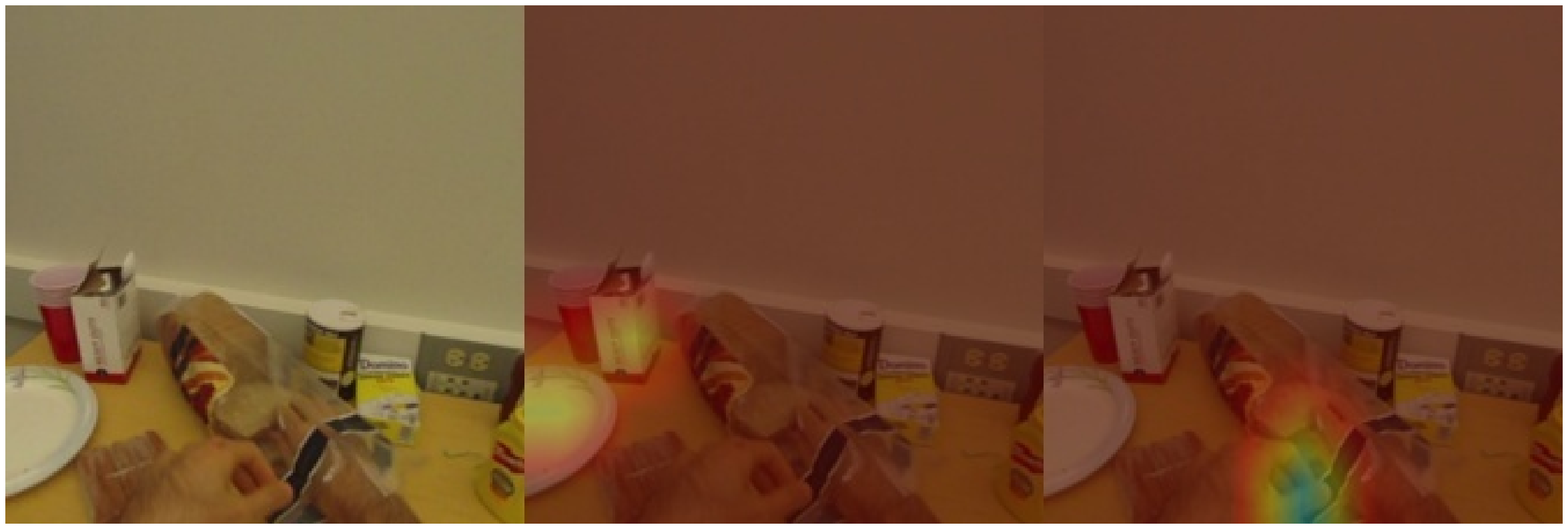}
	\end{subfigure}
	\begin{subfigure}[b]{0.32\textwidth}
		\includegraphics[scale=0.17]{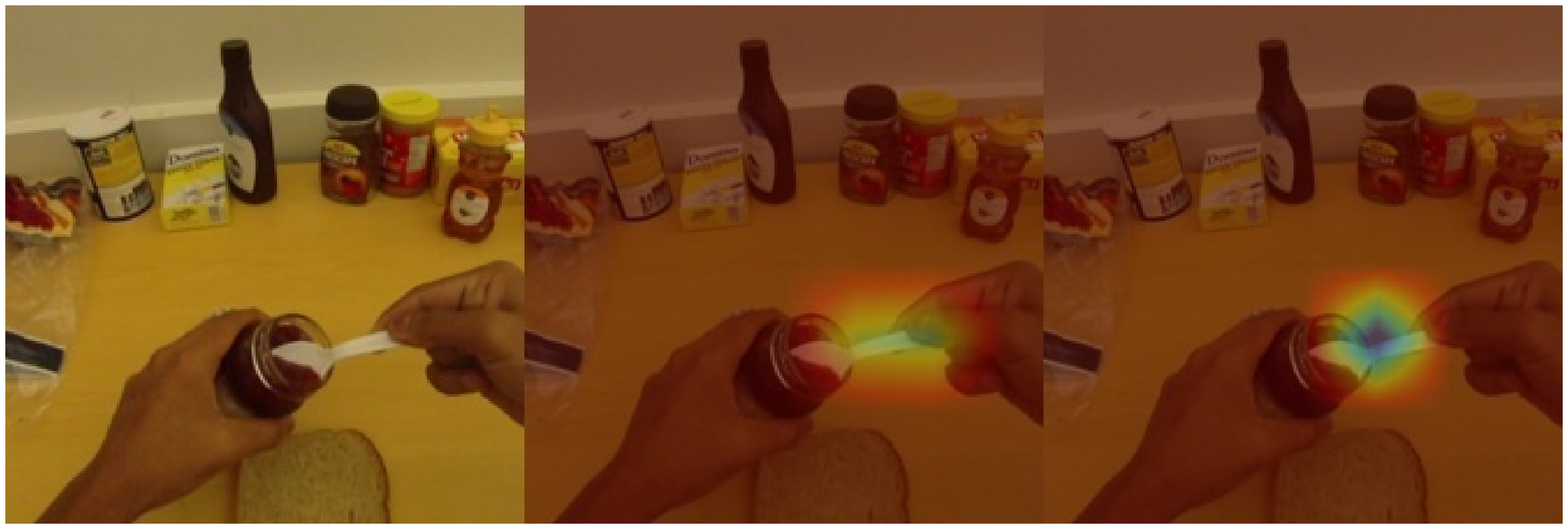}
	\end{subfigure}
	\begin{subfigure}[b]{0.32\textwidth}
		\includegraphics[scale=0.17]{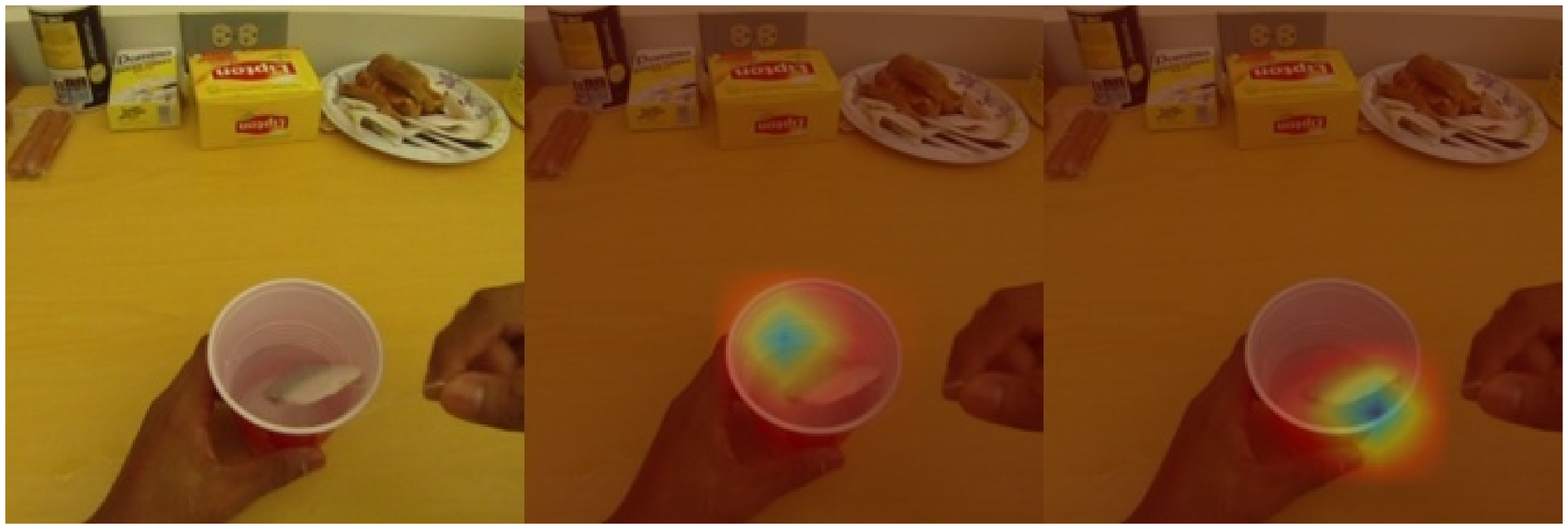}
	\end{subfigure}\\
	\vskip 2mm
	\begin{subfigure}[b]{0.32\textwidth}
		\includegraphics[scale=0.17]{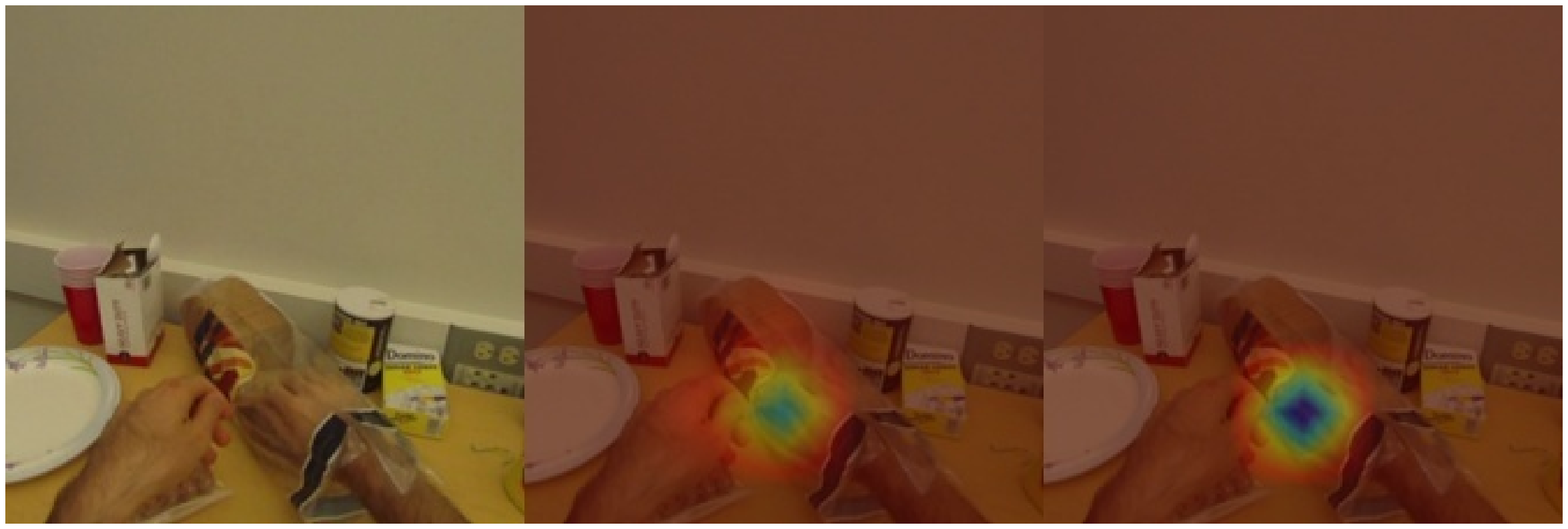}
	\end{subfigure}
	\begin{subfigure}[b]{0.32\textwidth}
		\includegraphics[scale=0.17]{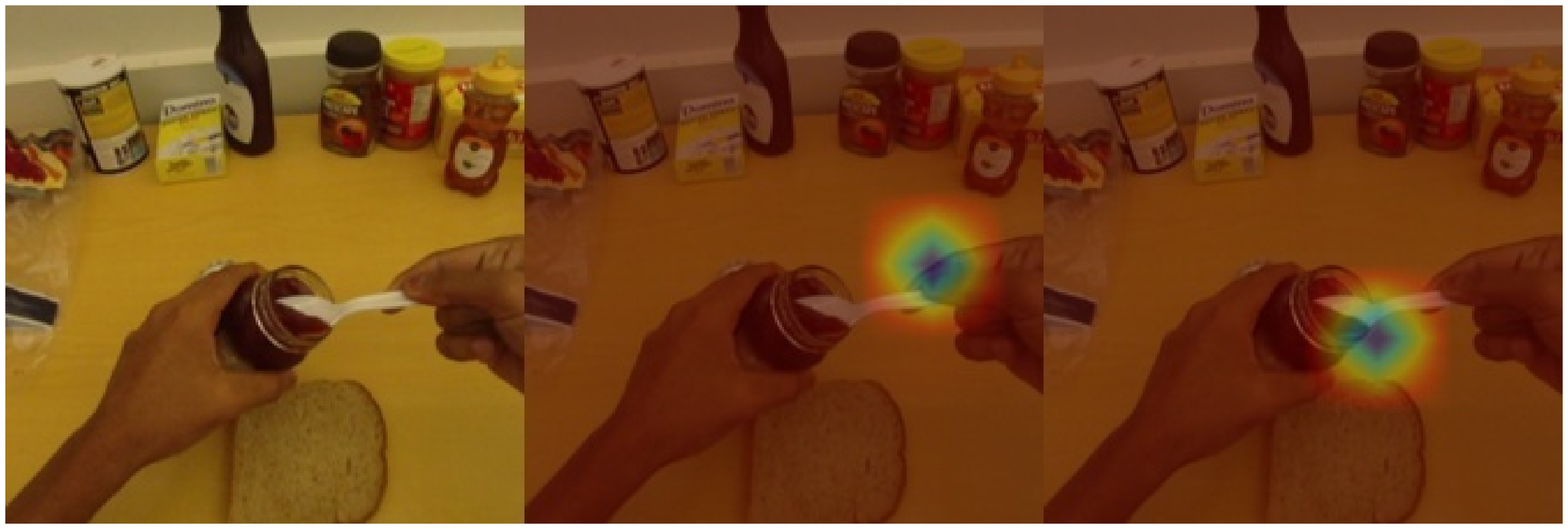}
	\end{subfigure}
	\begin{subfigure}[b]{0.32\textwidth}
		\includegraphics[scale=0.17]{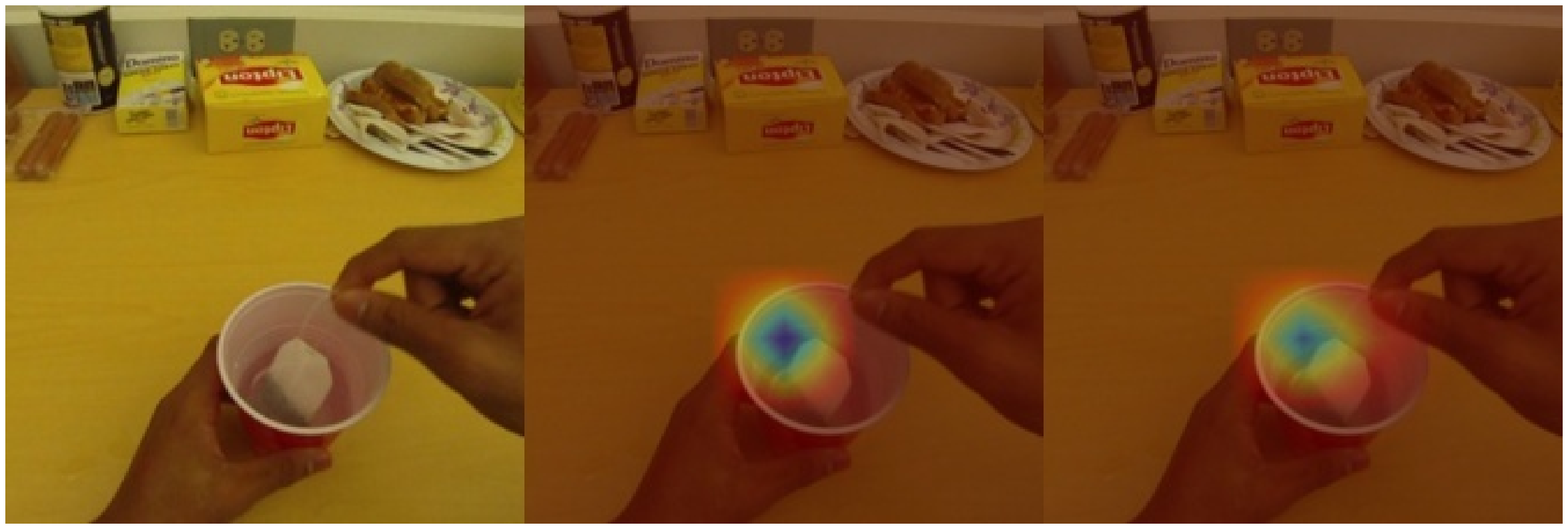}
	\end{subfigure}\\
	\vskip 2mm
	\begin{subfigure}[b]{0.32\textwidth}
		\includegraphics[scale=0.17]{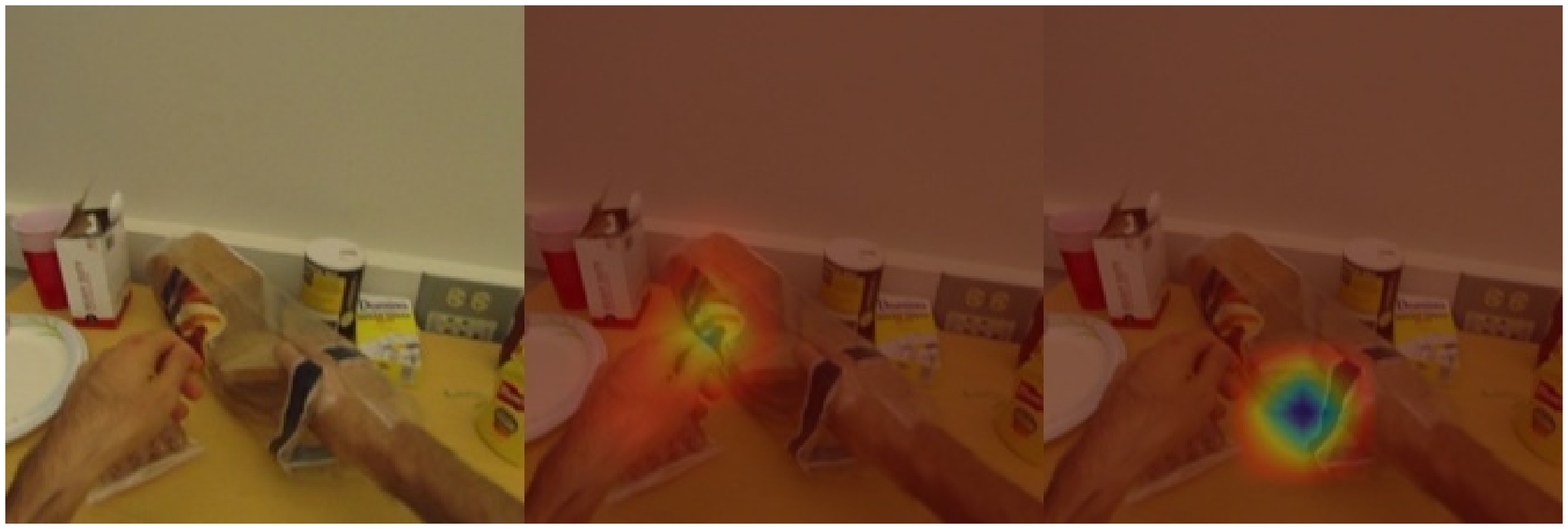}
	\end{subfigure}
	\begin{subfigure}[b]{0.32\textwidth}
		\includegraphics[scale=0.17]{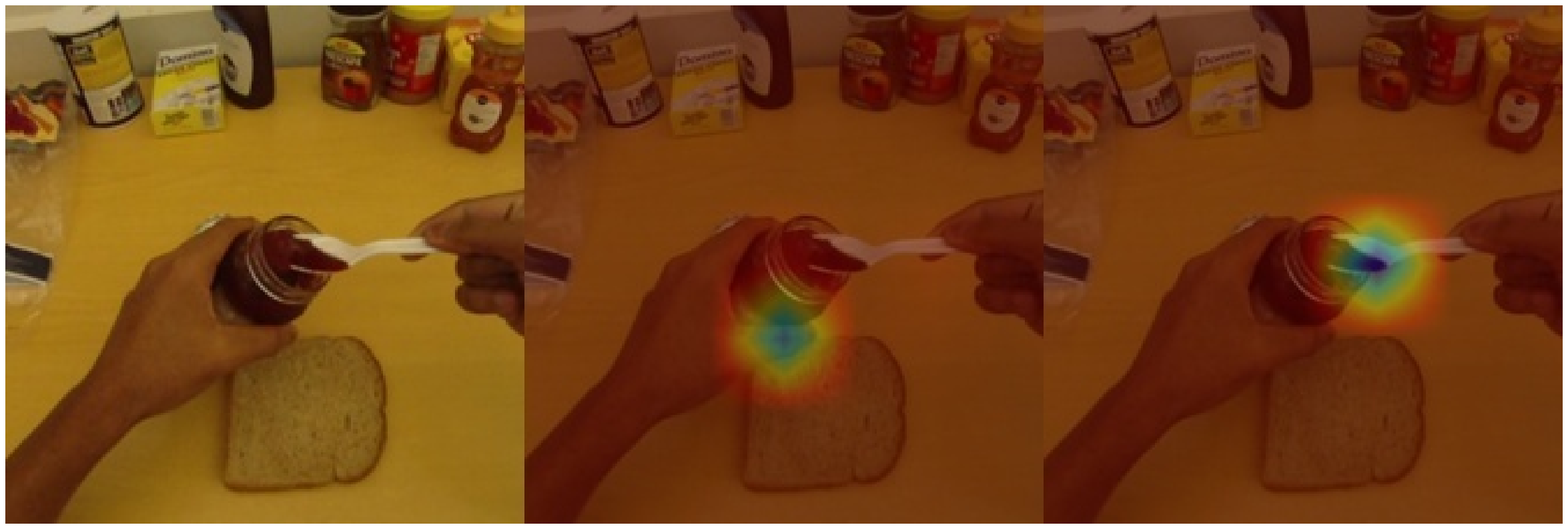}
	\end{subfigure}
	\begin{subfigure}[b]{0.32\textwidth}
		\includegraphics[scale=0.17]{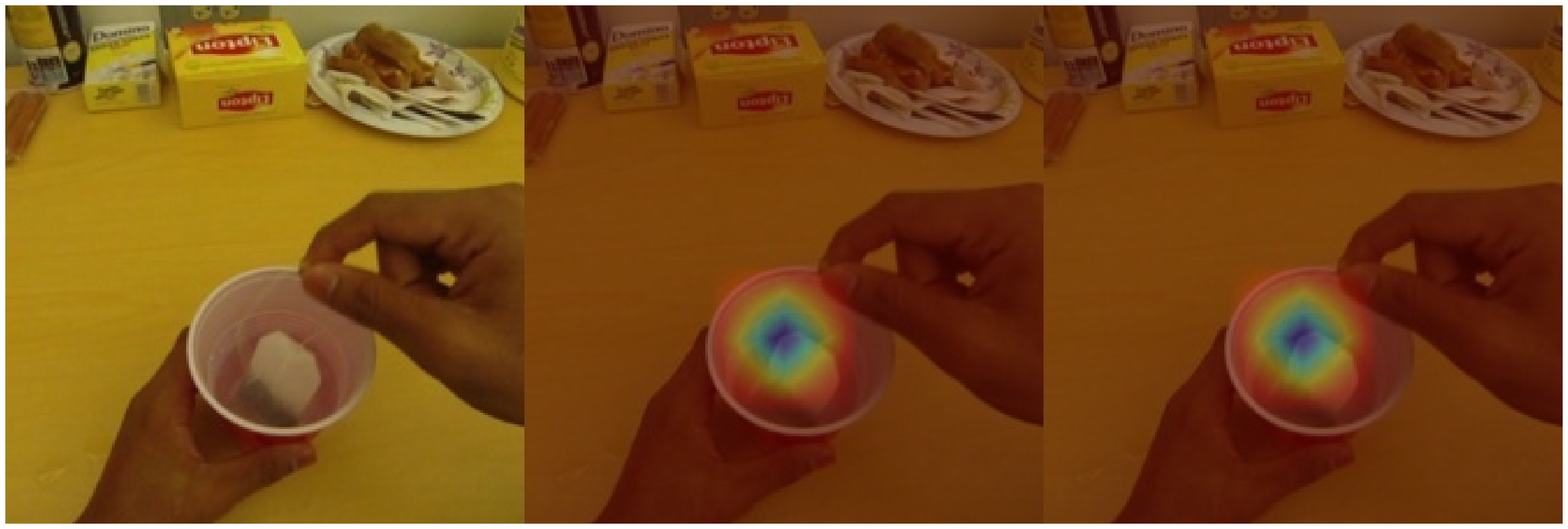}
	\end{subfigure}\\
	\vskip 2mm
	\begin{subfigure}[b]{0.32\textwidth}
		\includegraphics[scale=0.17]{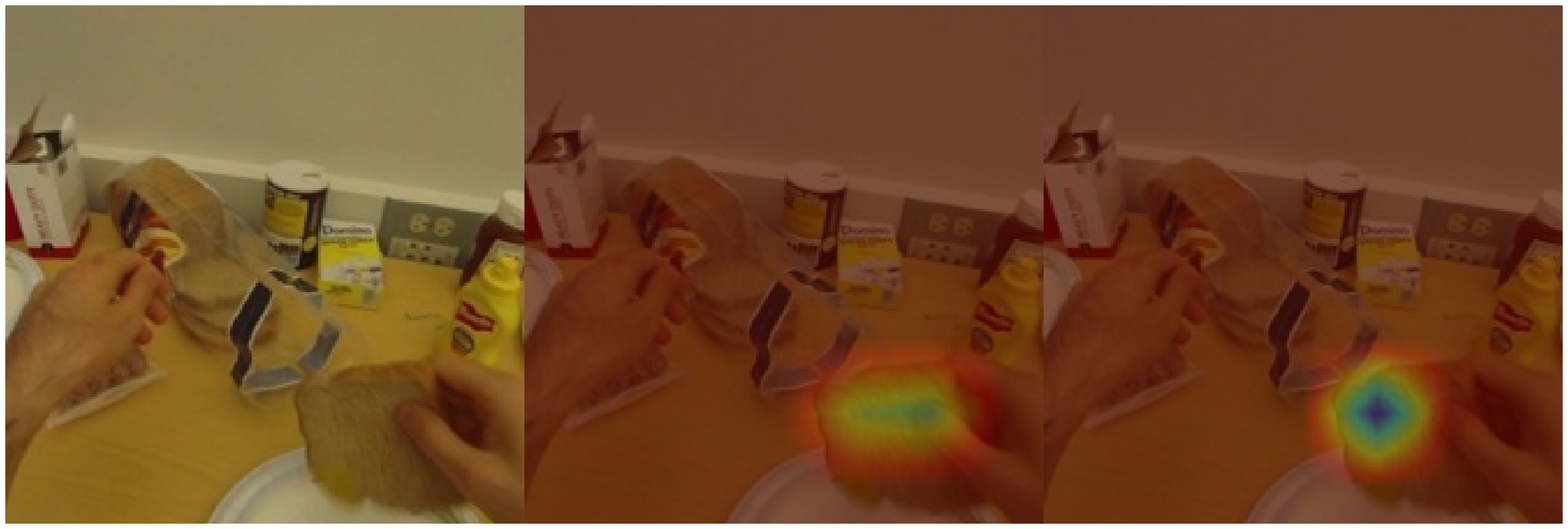}
		\caption{Take bread}
	\end{subfigure}
	\begin{subfigure}[b]{0.32\textwidth}
		\includegraphics[scale=0.17]{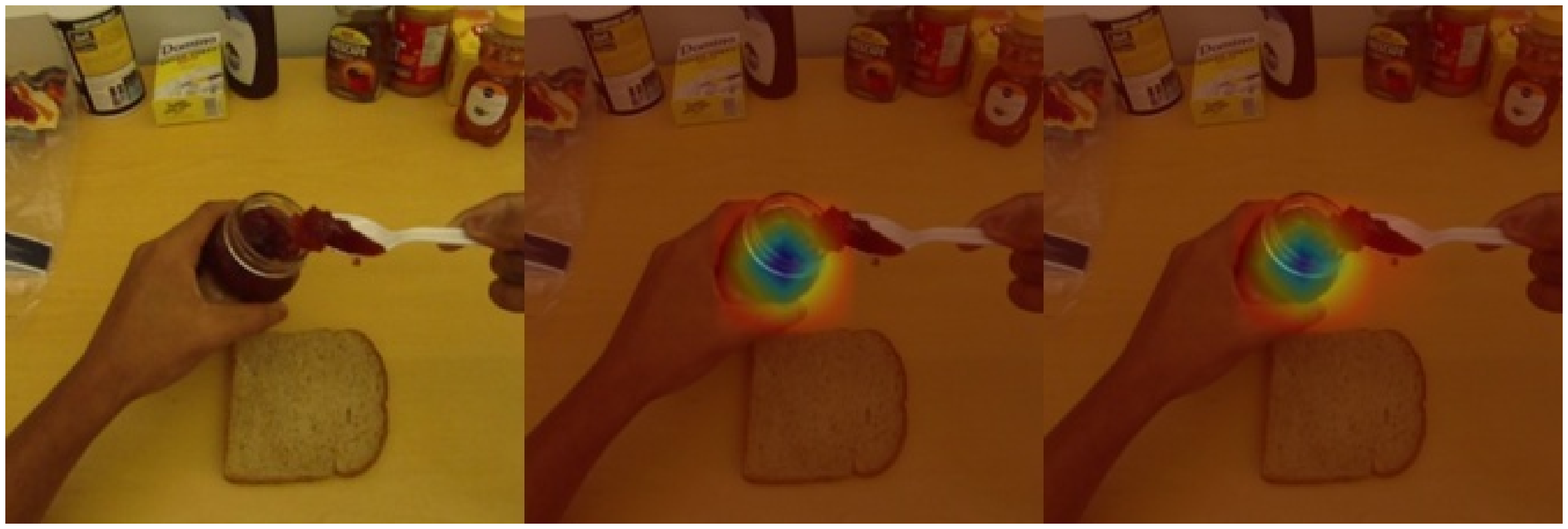}
		\caption{Scoop jam}
	\end{subfigure}
	\begin{subfigure}[b]{0.32\textwidth}
		\includegraphics[scale=0.17]{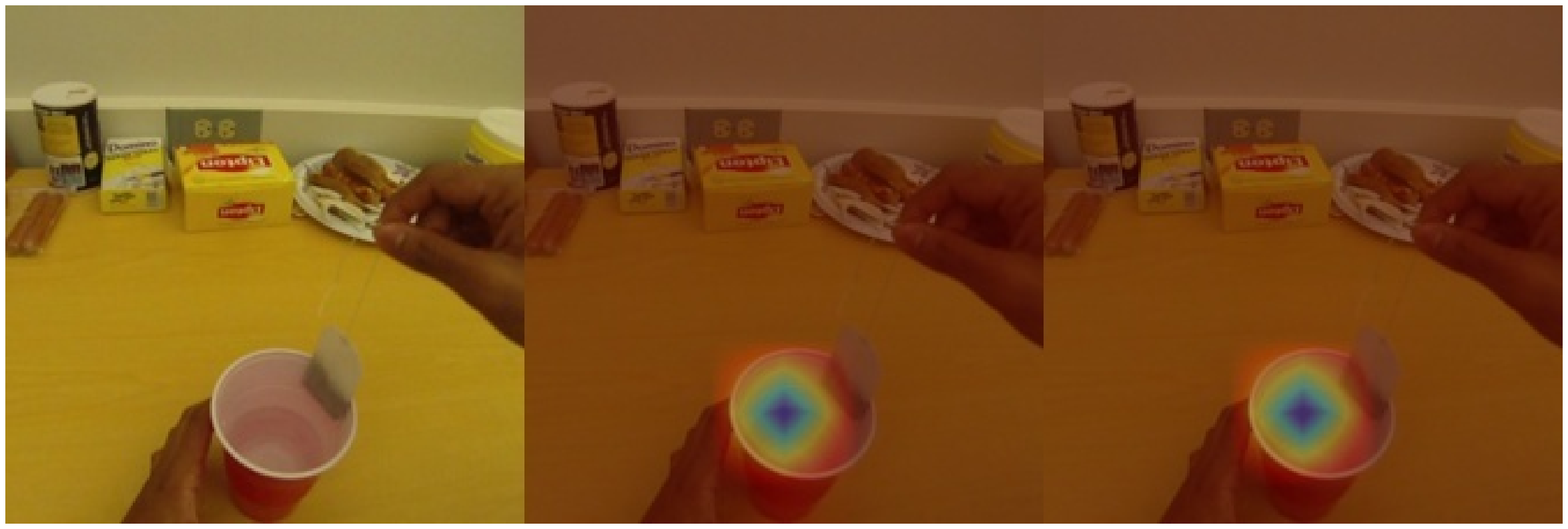}
		\caption{Shake tea}
	\end{subfigure}
			\vspace*{2mm}
	\caption{Spatial attention maps obtained for frames from GTEA(61) dataset. The clip identifiers are: (a) S1\_Hotdog\_C1 (b) S2\_Pealate\_C1 (c) S2\_Tea\_C1}
	\label{fig:fig_ex3}
\end{figure}
\FloatBarrier

\section{Confusion matrix of GTEA Gaze+ dataset}
\label{sec:3}
As discussed in section 4.3 of the paper, we show the
confusion matrices obtained when P1 and P3 are used as the test split of the GTEA Gaze+ dataset. These two splits resulted in the least recognition accuracy out of all the splits. P1 gave an accuracy of 50.2\% while P3 resulted in 48.84\%. We can see that the accuracy of classes containing the meta-object labels (`spoonForkKnife' and `cupPlateBowl') is low compared to the other classes. This verifies our hypothesis explained in section 4.3 regarding the reason for the lower performance of the proposed method on GTEA Gaze+ dataset compared to the methods that use strong supervision for training.
\begin{figure}[h]
	\centering
	
	\includegraphics[trim={.6cm 0 0 0},clip,scale=.95]{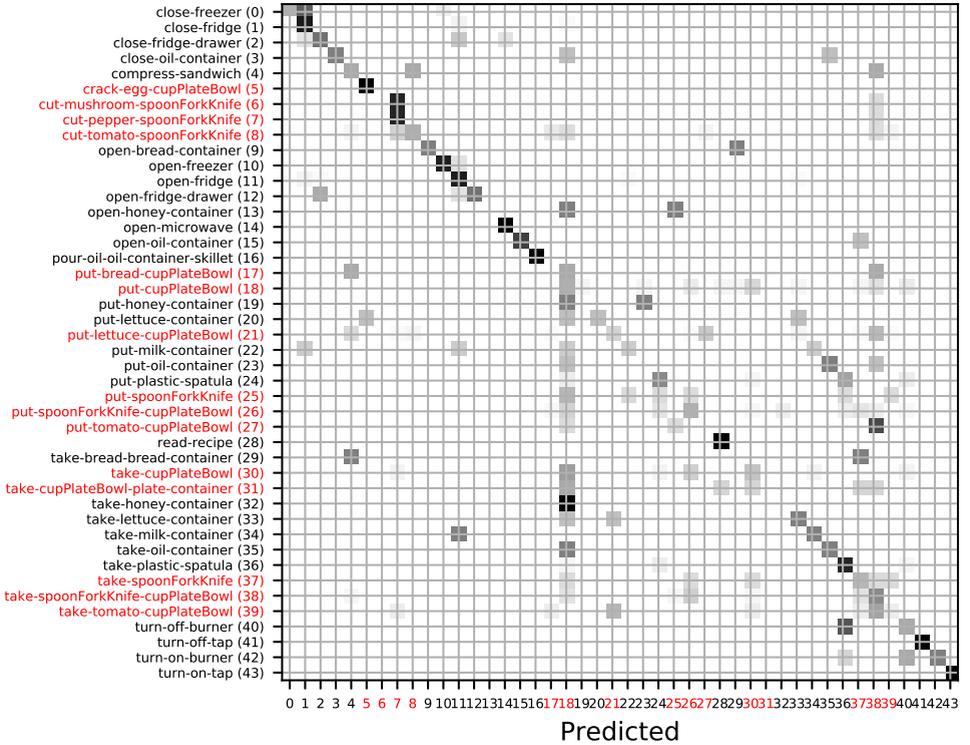}
	
	\caption{Confusion matrix of split P1 from GTEA Gaze+ dataset}
\end{figure}

\begin{figure}[h]
	\centering
	
	\includegraphics[trim={.6cm 0 0 0},clip,scale=.97]{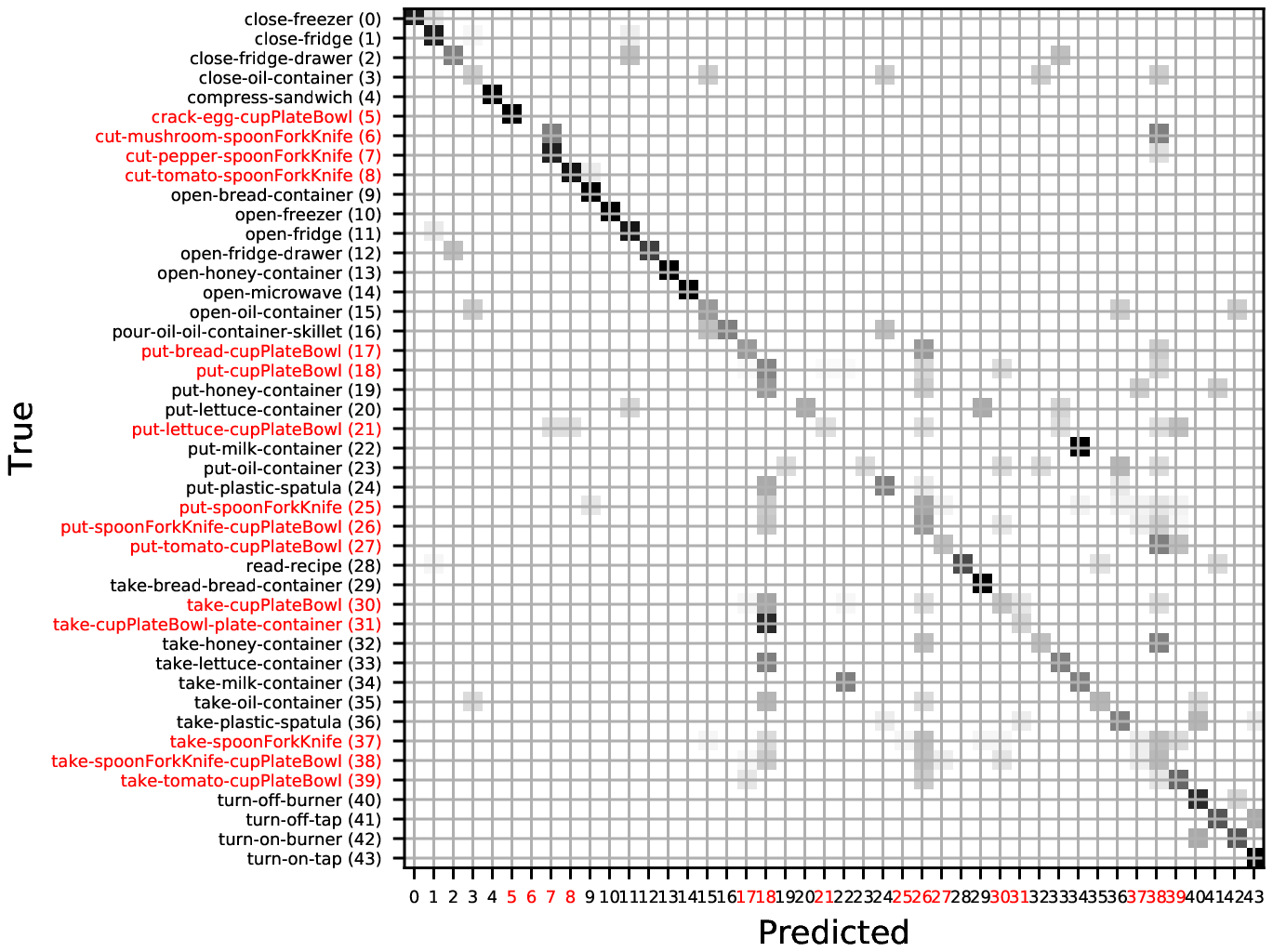}
	
	\caption{Confusion matrix of split P3 from GTEA Gaze+ dataset}
\end{figure}

\end{document}